# PI — TCAN

## Programme interdisciplinaire CNRS

## « Traitement des connaissances, apprentissage et NTIC »

# Projet OGRE

*Ordres de Grandeur et REpétition*

Rapport technique

Octobre 2005

*G. Bécher, P. Enjalbert, E. Fievé, L. Gosselin, F. Lévy, G. Ligozat*

# Introduction

Il y a trois ans, nous avons entrepris de comparer les approches linguistiques et IA de l'itération – celle d'un événement, telle qu'elle apparaît au plus simple par exemple dans "Tous les lundis, il mange au restaurant". Nous avions l'idée que la question est riche en problèmes scientifiques compliqués. Notre attente n'a pas été déçue, et si nous avons le sentiment d'avoir quelque peu contribué à défricher le terrain, nous ne l'avons certes pas épuisé. A commencer par sa définition, moins simple qu'il n'y paraît.

Premier constat, dans "Jean mange au restaurant", le verbe dénote un événement, dans "Tous les lundis, Jean mange au restaurant" le même verbe à la même personne, au même temps, au même mode, en dénote plusieurs. A notre connaissance, le premier à avoir entrepris une analyse approfondie du phénomène est Georges Kleiber [Kleiber 87]. Il discute, à propos des phrases qu'il appelle *habituelles,* la question de la référence verbale multiple : son trait caractéristique est qu' « il faut une pluralité d'occurrences réalisées pour satisfaire les conditions de vérité de ces phrases ». A cause de cette particularité, elles donnent lieu à des inférences sur des événements potentiels ou contrefactuels, qui signalent leur spécificité sémantique (ex: "Tous les lundis, Jean mange au restaurant ; lundi dernier il n'a pas mangé"). Ces propriétés communes recouvrent une multiplicité de significations qu'il n'est pas simple d'analyser. G. Kleiber classifie les phrases à référence verbale multiple en itératives ("Paul s'est rasé 3 fois la semaine dernière"), fréquentatives ("Paul est allé à l'école à pied le mois dernier") et habituelles ("Paul va à l'école à pied").

La distinction est intuitive au vu des exemples, mais son fondement n'est pas évident à analyser. Ce chapitre est essentiellement consacré à l'extraction automatique d'exemples et à comprendre les variations d'interprétation possibles. Nous citerons d'abord la discussion que mène G. Kleiber des termes logiques dans lesquels interpréter les phrases habituelles. Dans son analyse, celles-ci sont proches des fréquentatives, mais comportent en plus une dimension nomique. Il compare deux sémantiques. La première essaye de transposer au verbe la notion de généricité. Dans le cas nominal, celle-ci est polymorphe, entre les énoncés qui s'appliquent au genre en tant que tel (le chat est un mammifère) et ceux qui s'appliquent à la [quasi] totalité des individus en tant qu'ils appartiennent à l'espèce (les chats retombent [généralement] sur leurs pieds). Ce deuxième usage relève de formes plus ou moins étendues de quantification. Dans le cas verbal, la notion de genre n'a pas d'équivalent évident et la généricité verbale se traduit par une quantification sur les occurrences de l'événement.

L'approche quantificationnelle des habituelles soulève beaucoup de difficultés. Nous en décrivons rapidement trois, renvoyant le lecteur intéressé à [Kleiber 87]. Tout d'abord, sur quoi quantifie-t-on ? Dire que Paul va [quasi constamment] à l'école à pied ne signifie pas que cette activité occupe la totalité de son temps. On a donc besoin d'un ensemble d'*occasions restreintes* où l'événement en question pourrait se produire, et par rapport à quoi on quantifie les occasions référées (dans l'exemple, les jours de classe à l'heure de son ouverture). Mais la restriction dépend du contexte et introduit dans l'interprétation une forte dose de pragmatique. Ensuite, une fois choisies ces occasions restreintes, la quantification reste insatisfaisante : les phrases habituelles ne relèvent ni d'un quantificateur universel (qui interdirait les exceptions), ni d'un quantificateur générique. Dans cette seconde hypothèse en effet,, il faudrait multiplier les quantificateurs pour rendre compte des nuances entre 'généralement', 'habituellement', 'normalement',… . Enfin, des phrases traduites de façon identique par l'approche quantifica-tionnelle, par ex.  "Paul fume" et "généralement, Paul fume", n'acceptent pas les mêmes modificateurs (déjà, depuis 3 ans, …).

L'approche alternative repose sur une analyse en termes d'aspect et d'intervalles temporels. D'autres faits de langue militent en faveur de cette solution, et d'abord la sensibilité de l'habituel au temps verbal : "Paul [alla / est en train d'aller] à l'école à pied" exclut la lecture habituelle. L'importance du caractère non statif du prédicat (i.e. il ne peut pas représenter un état, donc s'étendre dans la durée sans impliquer des bornes) va dans le même sens : "Depuis un an, Paul écrit un roman" n'implique pas des occurrences multiples au même titre que "Depuis un an, Paul fume".
Une fois reconnues les occurrences multiples, [Kleiber 87] peut reformuler la différence entre itératives, habituelles et fréquentatives, Les fréquentatives réfèrent à un intervalle global et à la façon dont les différentes occurrences l'occupent. Les habituelles indiquent une forme de fréquence et lui ajoutent un caractère nomique. Il s'agit au sens étymologique d'une régularité, i.e. la répétition a quelque chose d'une loi. Par rapport à ces deux pôles, dans l'itération il y a multiplicité, mais ni ensemble, ni loi. En termes de conditions de vérité, on aboutit ainsi à une distinction entre vérité pour un intervalle (i.e. sur l'ensemble de l'intervalle – cas d'une phrase habituelle et d'une fréquentative) et vérité dans un intervalle (en certains points de cet intervalle – cas des itératives).

Nous avons cherché à développer cette analyse, dans trois directions. Tout d'abord en l'intégrant à la description de l'aspect de [Gosseelin 96], qui différentie les intervalles d'énonciation, de procès et de référence, i.e; qui distingue le moment dont on parle (l'intervalle de référence), et par rapport auquel on situe/met en valeur le procès, du support de ce procès lui-même (cf. l'article de L. Gosselin dans ce rapport). Cette distinction entre intervalle de référence et intervalle de procès rend compte des nuances de l'aspect et le rend calculable dans le cas non itératif non modalisé ([Person 04]). Ensuite en cherchant à traduire formellement cette notion d'ensembles d'intervalles sur lesquels le prédicat itéré est vrai ou pas. Si l'on veut pousser l'analyse de G. Kleiber jusqu'au point où les relations temporelles sont calculables, on est amené à représenter à la fois les intervalles des occurrences et un intervalle commun qui les contient toutes. Dans les fréquentatives on doit raisonner sur le rapport de *l'ensemble* des intervalles d'occurrence à l'intervalle commun, que celui-ci soit de quasi-recouvrement ('toujours', 'souvent', …), à l'opposé de quasi-absence ('rarement'), ou une autre nuance. La notion de régularité présente dans les habituelles se prête plus facilement à la manipulation formelle, dès lors du moins que les règles en sont assez apparentes. Les articles de G. Bécher et P. Enjalbert d'une part, de E. Fiévé et G. Ligozat de l'autre, explorent deux voies pour ces raisonnements. Enfin, la définition même de l'itération fait appel à la référence, et son interprétation fait largement appel à la pragmatique  - G. Kleiber l'a souligné pour la reconnaissance des occasions restreintes ou pour l'identification du focus. Néanmoins, la plupart des exemples sont des phrases artificielles. Il nous a paru utile de faire une étude expérimentale pour avoir une idée de la variabilité du phénomène. C'est l'objet du texte de F. Lévy qui ouvre ce rapport.

# Itérations

F. Lévy

Explication : le dénombrement des occurrences est fréquemment considéré comme l'archétype et même le seul représentant de l'itération. Ce critère du comptage nous semble inessentiel : entre "il s'est rasé tous les jours cette semaine" et "il s'est rasé trois fois cette semaine", nous voyons des nuances sur la description des intervalles d'occurrence, mais pas une rupture qui justifie une catégorisation distincte. Le corpus illustrera ce point de vue. Nous conservons donc le terme d'itération dans son sens habituel pour les logiciens de *répétition à l'identique*, i.e. comme un terme général pour l'ensemble des références verbales multiples

Nous avons utilisé comme corpus 3 ans du journal Le Monde, soit un document de 57 millions de mots, 2,7 millions de phrases. Un pré-traitement découpe le corpus en phrases, puis divers motifs sont extraits. Il fallait pour extraire automatiquement des phrases qui soient presque toutes des exemples d'itération se limiter à des structures syntaxiquement reconnaissables. Ce sont les mêmes qui sont représentées dans l'article de G. Bécher et P. Enjalbert. Dans la liste qui suit, *label* est un jour de la semaine, Noël, une saison ou une indication de période (heure, matin, mois, trimestre) :

- Tous les *label*, Toutes les *label* (ex. tous les mardis)
- Tous les *n label* ou Toutes les *n label*, où *n* est un nombre (ex: toutes les trois semaines)
- Chaque *label* (chaque dimanche, chaque été, …)
- fois par *label*. (ce schème incomplet inclut *n* fois par *label*, mais aussi plusieurs fois… , plusieurs centaines de fois…, quelques fois …, sans introduire de bruit)
- par *label*, où par n'est pas précédé de fois.
- *Label* par *label* (ex. jour par semaine) ; c'est un sous-ensemble du précédent, mais avec très peu de bruit.
- Les *nièmes label* de, où *nième* est un ordinal (les premiers jours de septembre)
- La plupart des *label*
- *n label* sur *n'*, où n et n' sont des nombres (un lundi sur trois)
- certains *labels*, quelques *labels*
- *n label* par *label*. Ce motif, inclus dans 'par *label*', comporte moins de bruit
- les tournures génériques : le *jour* [non suivi d'une date pour exclure "le lundi 19 janvier"], les *jours* de *période* (ex : les lundis d'été), souvent le *jour ou saison* (souvent le mardi, l'été)

L'ensemble pré-traitement + extraction consomme environ 1h30 de temps de calcul – avec un programme écrit en Perl, ce qui n'est pas très performant. Il en résulte environ 18000 exemples (bruit compris : on verra que certains patrons sont médiocrement sélectifs). La fréquence des différentes tournures n'est pas la même :

| Tournure | Nombre de phrases |
|---|---|
| Chaque | 7384 |
| Par (hors fois par) | 4381 |
| Tous les | 3063 |
| Le *générique* | 1868 |
| Tous les *n* | 699 |
| Fois par | 578 |
| *durée* par *période* | 370 |
| *n label* sur *n'* | 345 |
| Certains | 106 |
| La plupart des | 7 |

Nous proposons ici une analyse de ces exemples. La difficulté principale de la tâche tient à la notion même d'itération : son critère essentiel de reconnaissance est référentiel ; mais la notion de référence pertinente pour cela est constituée par *les éléments de la situation que l'énonciateur voulait décrire* (vs la description détaillée qu'un auditeur informé peut déduire ou imaginer). Il y a là une marge d'interprétation et des difficultés théoriques bien connues (cf. [Formigari], [Eco] dernier ch). Les traiter nous entraînerait loin de notre objectif, et loin de nos compétences. Ce travail ne doit être considéré que comme une première ébauche..

## Répétition et fréquence

### Intervalle de répétition et intervalle de procès

Toutes les formes considérées impliquent une répétition, indiquent l'intervalle de temps entre deux occurrences et parfois, de façon directe ou indirecte, précise ou vague, la durée d'une occurrence :

```
(1a) Chaque matin, je commence à composer à 9 heures, après
avoir emmené mes  enfants à l'école, et je travaille ainsi
jusqu'à 17 heures environ.
(1b) [Quel avenir réserve le conseil aux trois chaînes
actuelles _] F1 (fédérale), qui diffuse dix-huit heures par
jour (de 6 heures à minuit) …
(1c) Les députés socialistes ont d'ailleurs pris la décision
de se retrouver tous les mardis afin d'être régulièrement
informés de l'évolution du conflit par les membres du
gouvernement
(1d) Pour une belle histoire, c'est une belle histoire :
celle du padre Gaetano, prêtre dévoué à la défense des
enfants orphelins d'une  région déshéritée du Mexique le
jour, combattant de catch le soir.
(1e) Dépendante du Centre national du cinéma, renouvelée tous
les deux ans, cette commission, qui permet en principe à un
film de trouver son  financement, se compose de deux collèges
(1f) Dans les clubs de haïkus, où l'on se réunit une fois par
mois, les poèmes de chacun des membres font l'objet de
débats, de discussions,  de commentaires et d'un vote,
témoignant d'une authentique démocratie littéraire.
(1g) Résultat :  les premiers temps, TV 5 émet trois heures
par jour, de 19 heures à 22 heures.
```

(1h) Un pasteur très attendu et grâce auquel la petite communauté établie autour de Stellenbosch-Drakenstein poursuivra, dans sa langue, un dimanche sur deux, la pratique de sa religion.

Il est intéressant de reprendre au niveau de l'événement répété la distinction établie par L. Gosselin entre l'intervalle de référence, l'intervalle de procès et l'intervalle circonstanciel. Les tournures de langage étudiées ici ont l'avantage de marquer clairement l'intervalle circonstanciel. Ceci confère à l'événement répété une forme d'*aspectualité relative* par rapport à un intervalle de référence lui-même itéré. On comparera par exemple :

(2a) Chaque année le Cinéma du réel permet de découvrir quelques-uns des meilleurs documentaires qui se fabriquent dans le monde.
(2a') Les promotions devraient doubler d'ici à 1993, pour aboutir aux quelque trente mille nouveaux ingénieurs dont le pays a besoin chaque année.
(2b) Deux jeudis par mois, à 12 h 40, Marie-Christine Vallet présente Europe 93.
(2b') J'écris seize, dix-sept et même dix-huit heures par jour ; aussi longtemps que je peux.
(2c) Solidarité rurale oblige, des voisins passent tous les jours pour le ravitaillement et les médicaments
(2c') Ils sont livrés, tous les jours, par une noria de camions, qui ont juste le temps d'arriver pour que les robots montent les pièces sur les chaînes.
(2d) Je sortais de chez moi le matin et je voyais l'enseigne Hollywood accrochée sur la colline.
(2d') Dans ma chambre colmatée contre les gaz, en attendant les missiles, le soir, j'ai peur, peur de l'inconnu, a déclaré Yoram Kaniuk.
(2e=1e) renouvelée tous les deux ans, cette commission, qui permet en principe à un film de trouver son financement, se compose de deux collèges
(2e') La présidence de l'association est assurée alternativement, tous les deux ans, par Alain Richard, maire (PS) de Saint-Ouen-l'Aumône, et André Santini, maire (UDF) d'Issy-les-Moulineaux.
(2f) C'est ainsi, par exemple, que les radars et les systèmes de télécommunications modernes doivent changer de fréquence de manière aléatoire plusieurs centaines de fois par seconde pour tenter d'échapper au brouillage, et compliquer la tâche des missiles.
(2f') Deux réseaux permettent aujourd'hui de recueillir les informations nécessaires : l'un est constitué par les satellites américains NOAA placés en orbite polaire à 850 km d'altitude, qui dressent deux fois par jour une photographie complète du globe ; l'autre…
(2g) La SACD, Société des auteurs-compositeurs dramatiques, organise deux mercredis par mois jusqu'en juin 1992 des projections de fictions réalisées pour la télévision.

(2g') Prises  dans les glaces neuf mois par an, ces deux redoutes avancées se sont surveillées sans relâche, usant à cette fin de tous les perfectionnements de l'électronique.
(2h) Attribué une année sur deux, le prix Sonning est destiné à couronner une personnalité qui, au cours de sa vie, a oeuvré en faveur de la  culture européenne.
(2h') Les Pygmées Akas, dont le royaume vert, noyé sous les pluies neuf mois sur douze, s'étend sur la frontière de la République centrafricaine et du Congo.

Dans (2a), l'intervalle circonstanciel est beaucoup plus étendu que l'événement couvert (un festival de quelque jours) qui est localisé dans une partie restreinte de cet intervalle, ce que marque le choix d'un verbe ponctuel. Quelle que soit l'analyse que l'on adopte – procès restreint ou partie d'un meta-procès qui est le déroulement annuel – l'année sera temporellement structurée par l'événement en référence. Dans (2a') au contraire, le besoin n'est aucunement localisé et concerne l'ensemble de l'année.
De même dans (2b) l'événement (présenter une émission) est intrinsèquement borné (parce qu'il s'agit ici d'une *occurrence* d'émission). Dans (2b'), `écrire` est atélique et `seize heures` ne délimite pas une partie de la journée (contrairement à `jeudi`, spécifiant une partie du mois – les seize heures peuvent être discontinues). La tournure n *label* par *label* ne se prête pas à indiquer explicitement une régularité totale, mais la pragmatique y pourvoit facilement : tant les émissions de radio que les heures de sommeil ont sans doute une certaine régularité.
Dans (2c) encore, chacune des occurrences répétées est limitée dans le temps, alors que (2c') marque plutôt la régularité du flux que la répétition. L'exemple est d'autant plus intéressant que la différence d'aspectualité interne n'est pas liée au verbe : *passer* et *livrer* sont tous deux des  accomplissements[1], et c'est l'emploi du mot "noria" qui fait la différence – il suffit de considérer :

(2c") Ils sont livrés, tous les jours, par un camion, qui a juste le temps d'arriver pour que les robots montent les pièces sur les chaînes.

L'opposition des types de procès, ponctuel dans un cas, non borné dans l'autre, et en même temps de l'aspectualité relative, se retrouve dans trois autres exemples : `sortir de` versus `avoir peur` (2d); `renouveler une commission` versus `assurer la présidence` (2e), `attribuer un prix` versus `être noyé sous les pluies` (2h). Par contre, dans (2f'), `dresser une photographie` est lui aussi un procès borné (compatible avec "à midi", "en 2heures", pas avec "pendant 2 heures"). Pourtant on peut comprendre en l'espèce que les satellites mènent cette activité de façon continue. Dans (2g), organiser est un procès non borné (compatible avec "pendant + durée"). Pourtant, l'intervalle de procès ne couvre qu'un fragment de l'intervalle circonstanciel (les mercredis pertinents) ; le bornage est en fait amené par le groupe nominal `des projections de fictions`, qui spécifie l'activité.

## Contiguïté des intervalles de répétition

Nous avons considéré jusqu'à présent que les phrases analysées croisaient deux phénomènes : l'indication d'un intervalle circonstanciel itéré, et l'indication d'un processus qui occupe cet

[1] Au moins quand livrer désigne la phase finale de la livraison.

intervalle de façon différente suivant les cas. Il manque à ces critères au moins un élément essentiel : la façon dont la succession des procès élémentaires se constitue plus ou moins en un tout qui a ses propres propriétés sémantiques. Quand par exemple la série des intervalles de répétition est discontinue (les matins dans (1a), les lundis, mercredis ou vendredi dans (3))

(3a) cette ancienne productrice de Canal J décortique chaque mercredi, devant des enfants des centres aérés de la Mairie de Paris, les images du grand et du petit écran
(3b) On embarque à Amsterdam le lundi à partir de 19 heures et on débarque à Strasbourg le vendredi vers 15 heures.

on comprendra bien plus naturellement que le focus est mis sur la réitération d'épisodes semblables. Quand les intervalles de répétition sont jointifs (les heures, les semaines, les années…), le caractère véritablement itératif dépend de cette occupation de l'intervalle de répétition par le processus. Par exemple (4a) suppose une série des discours créés, alors que dans (4b) la série des métiers créés n'aurait pas de sens.

(4a) Chaque année, lors de son discours d'accueil des nouveaux élèves, le directeur de Sciences-Po, M. Alain Lancelot, fustige ces formations privées.
(4b) Chaque jour, un nouveau métier se crée, explique Bernard Lemée

On peut remarquer dans ce second exemple la quasi-impossibilité sémantique d'ajouter le même sorte de complément de temps que comporte le premier (?? chaque jour, dans l'après-midi, un nouveau métier se crée …). En restreignant le domaine du processus *à un fragment* de l'intervalle de répétition, on rompt la continuité du support temporel du processus. Nous avons aussi calculé dans le corpus la part des deux types d'intervalles dans les expressions en `chaque`, `tous les` et `fois par`, qui sont les plus fréquentes (les résultats de la recherche automatique du *le* + *label* générique sont trop bruités pour donner des statistiques fiables). Les expressions de la forme tous les deux jours, toutes les trois semaines, sont comptées dans les périodes discontinues. Ce choix est justifié dans (5a) et plus incertain dans (5b) où la périodicité tient plutôt de la figure de style :

(5a) Cette révision du DTS, qui intervient en principe tous les cinq ans, …
(5b) Pourquoi diable certains écologistes persistent-ils à nous annoncer la fin du monde toutes les cinq minutes, …

| Périodes continues | | Périodes discontinues | |
|---|---|---|---|
| année | 3389 | dimanche | 131 |
| ans | 257 | jeudi | 55 |
| heure | 49 | lundi | 51 |
| jour | 3426 | mardi | 55 |
| minute | 19 | matin | 415 |
| mois | 554 | mercredi | 79 |
| saison | 7 | nuit | 167 |
| seconde | 35 | samedi | 102 |
| semaine | 751 | soir | 549 |
| semestre | 9 | vendredi | 50 |
| trimestre | 51 | ans | 393 |
| | | heures | 25 |
| | | jours | 32 |
| | | minutes | 52 |
| | | mois | 144 |
| | | secondes | 11 |
| | | semaines | 27 |
| Total : 8547 | | Total : 2318 | |

Les unités continues sont nettement plus fréquentes : le jour et l'année figurent chacun dans plus d'exemples que l'ensemble des périodes discontinues. Ce phénomène traduit leur rôle comme unité de mesure de flux, et est lié à l'influence du pluriel sur l'interprétation de la phrase, que nous considérons maintenant.

# Procès et multiplicité référentielle

## Multiplicité des acteurs

La complexité des situations qui combinent itération et pluriel s'explique largement par la grande souplesse sémantique de ce dernier et par la place laissée à la pragmatique pour exprimer et, à la réception, pour résoudre des questions de signification essentielles. En effet, dès que le prédicat a plusieurs sujets ou plusieurs compléments, le nombre de procès référencés peut varier, en dehors du mécanisme de l'itération, selon que la prédication est collective, totalement ou partiellement distributive ([Kayser, Gayral, Lévy]). (6) :

(6a) Aki a trente-trois ans, Mika trente-cinq …, ils animent chaque année en juin dans leur pays le Festival du soleil de minuit, et ne tiennent pas en place.

(6b) En accord avec le président de la Comex, il passa à mi-temps et trouva cinq PME de secteurs d'activités différents qui acceptèrent de l'employer une demi-journée par semaine.

(6c) l'institution prud'homale qui statue chaque année sur quelque 200 000 litiges individuels entre salariés et employeurs …

(6d) Cent soixante aller-retour, pas moins, trois fois par semaine pour se rendre de Metz, où elle habite, à Verdun, où elle enseigne

(6a) correspond bien à un intervalle discontinu pendant lequel les deux acteurs participent collectivement à un seul procès (l'animation du festival), et ce de façon répétitive. Dans (6b), chaque acteur (chaque PME) correspond à un procès particulier qui se déroule dans un intervalle différent (le sujet est employé cinq demi-journées par semaine), et chacun de ces procès est itératif. Dans (6c), le pluriel est en complément d'objet ; le verbe `statuer` est un accomplissement et son échelle de réalisation (dans le domaine du droit) est bien inférieure à l'année (comparer à (2c) et (2c')). Bien que sa signification première soit clairement distributive (on ne peut statuer sur des litiges que un par un), la multiplication d'occurrences indifférenciées en fait une activité continue qui occupe l'intervalle de répétition (l'année). Enfin dans (6d), alors que syntaxiquement, on pourrait comprendre que cent soixante aller-retour ont lieu chaque fois que l'héroïne se rend de Metz à Verdun, la multiplicité déborde l'intervalle de répétition – elle donne en fait les bornes de cette répétition : l'année scolaire accomplie par cette enseignante débutante (même si le calcul met en évidence une simplification rhétorique).

Nous avons utilisé jusqu'à présent la notion de procès en donnant un grand poids à l'exactitude référentielle. Cette notion reprend celle d'événement de Reichenbach en soulignant la possibilité d'avoir dans ce rôle des processus ou des états aussi bien que des événements. Lorsque la référence verbale est unique, les problèmes de détermination du procès semblent limités à celui du sens lexical du verbe. La référence verbale multiple introduit une nouvelle question : quelle structure sémantique / intentionnelle reçoit cette pluralité ? Qu'il y ait factuellement[2] plusieurs événements n'entraîne pas nécessairement que le focus soit mis sur la singularité de chacun. Comme le soulignait l'étude de G. Kleiber, il y a de nombreux exemples où le point de vue est plus macroscopique. (6c) est un exemple typique d'énoncé où l'on est plus tenté d'interpréter la multiplicité, pourtant distributive et indiquant donc logiquement une pluralité d'événements factuels, comme un procès global. On trouve de très nombreux exemples dans lesquels une forme linguistique correspondant formellement à une pluralité comptable de procès élémentaires doit plutôt être lue comme une mesure de densité :

(7a) Dès sa sortie, P1 est un triomphe : 1 million de dollars de recettes et 350 000 spectateurs chaque jour.
(7b) Pour l'instant on n'en est pas là, et aucun signe de détérioration de la situation n'a été constaté par les contrôleurs qui, neuf fois par jour, font des prélèvements au large.
(7c) D'ici à l'an prochain, les usines de Poissy et de Mulhouse devraient être capables de sortir 1 500 voitures par jour.
(7d) Parmi les centaines de propositions dues à des particuliers ou à des associations qui lui parviennent tous les ans, la commission municipale chargée du dossier (2) doit donc effectuer un tri draconien.
(7e) Le porte-parole du département d'État, Mme Margaret Tutwiler, a déclaré, lundi 15 avril, que, selon les estimations des secouristes internationaux sur place, entre 400 et 1 000 réfugiés irakiens meurent toutes les vingt-quatre heures, aux abords de la frontière irako-turque.
(7f) Quant au groupe Fiat, il ne s'est pas contenté … de mettre au chômage technique, à tour de rôle, les 287 000 salariés de sa branche automobile (à raison d'une ou deux semaines en moyenne par mois).

Le repérage du caractère massif de l'ensemble de références élémentaires est très lié aux connaissances encyclopédiques. On sait dans (7a) que le succès se mesure à la fréquentation, dans (7b, c, d, e, f) que seul le nombre compte et que les événements ne sont ni identifiés, ni localisés dans l'intervalle de répétition – ils l'occupent en quelque sorte en continu. Nous en donnons encore deux autres exemples qui montrent que cette massivité est indépendante du caractère continu de l'intervalle de répétition lui-même. Dans (8a, b), on a à faire avec une multitude d'événements regroupés dans des intervalles de temps non consécutifs :

(8a) L'armée irakienne est au bord de l'effondrement, vient de confier au Washington Post le général Norman Schwarzkopf, le commandant en chef de l'opération Tempête du désert, précisant qu'en ce moment les forces de la coalition détruisaient, chaque nuit, une centaine de chars ennemis.
(8b) Et un objectif : atteindre la barre des 700 000 visiteurs par an en poursuivant une animation et une fréquentation actuellement étalée sur quarante-quatre dimanches par an.

---

[2] Dans la mesure où on peut assimiler la référence à un morceau de réalité – ce qui n'est ici qu'une approximation provisoire.

(8b) est particulièrement intéressant en ceci qu'il illustre la co-existence pour le même processus d'une vision continue (un an) et d'une vision discrète (44 dimanches).

## Procès discontinus

Les exemples de densité du paragraphe précédent regroupent une multiplicité d'événements en un procès se déroulant sur un intervalle continu. La même incertitude sur ce qui doit être tenu pour un procès affecte les occurrences qui se correspondent dans des intervalles successifs. La question se pose même pour un événement unique itéré sur des intervalles disjoints : la discontinuité suffit-elle à justifier la pluralité des procès ? Il semble dans certains cas beaucoup plus justifié de considérer qu'un procès unique s'étend sur plusieurs intervalles disjoints. Une telle interprétation aurait pu être considérée pour (8a), et elle semble fondée pour (9):

(9a) Le dernier représentant de la catégorie[note : des dessins animés], Flash, produit par Danny Bilson et Paul de Meo, est depuis quelques semaines à l'affiche de TF 1, le mercredi après-midi.
(9b) On a beau nous désigner chaque soir les méchants, les agresseurs, on sent bien, en dépit du martelage, que l'obsédante référence au nazisme ne suffit pas.
(9c) M. Nallet s'est également entretenu avec le bâtonnier et des représentants du barreau de Lyon qui mènent une opération pilote de consultation juridique gratuite, chaque mercredi, et de défense spécialisée pour les mineurs.
(9d) Cette avant-première audacieuse _ le long-métrage, qui dure treize heures, sortira en salles fin mars _ s'inscrit dans le cadre d'une politique de programmes spéciaux diffusés sur le réseau câblé parisien chaque nuit de pleine lune.
(9e) Par la suite, les blessés continuent à recevoir de l'atropine par voie intramusculaire ou intraveineuse à raison de 2 mg toutes les dix minutes.

Dans (9a), la position du complément 'le mercredi' en incidente inessentielle (on peut le supprimer sans altération majeure du sens de la phrase) joue en ce sens. La sémantique du domaine aussi : le sujet de l'article est plutôt la grille des émissions, qui est relativement stable – ce qui est cohérent avec le choix d'une formulation générique "le mercredi". Le marqueur d'itération semble bien plutôt référer à un rythme interne au procès "être à l'affiche" qu'à une répétition de ce procès. Dans (9b), l'argument est très différent : c'est la reprise anaphorique de l'ensemble des itérations par un procès nominalisé (marteler) qui justifie de représenter cet ensemble comme un unique procès. Dans (9c), il serait difficilement acceptable de considérer qu'il y a une nouvelle opération pilote chaque mercredi. Dans (9d), c'est le mot 'politique' qui peut justifier de constituer l'ensemble des programmes spéciaux en une entité unique avec son créneau horaire particulier. Dans (9e), une série régulière d'actes médicaux constitue un traitement.

## Le double niveau

(9a) relève en fait d'une forme, les expressions en le + date incomplète (lundi, soir, etc.), que nous appellerons l'usage générique. Nous ne connaissons pas d'indice que les mécanismes d'extraction automatique par expressions régulières dont nous disposons pourraient utiliser pour différentier à coup sûr l'usage générique de l'usage singulier. On peut facilement éliminer certaines tournure : "le + *jour de semaine*+ *chiffre*", "le matin du", "le soir même" ; Il reste 1400 exemples relevés dans le corpus, dans lesquels l'usage générique est majoritaire mais pas le seul. Un traitement plus fin de l'aspect (cf par ex. [Reppert-Maire 90], [Gosselin 96], [Person 04]) permettrait d'éliminer certaines phrases comme (10a) dont le verbe est au passé composé, alors que (10b) par exemple ne peut l'être que par le contexte du récit :

```
(10a) Dès le soir de la mort de Djemel Chettou, dont le
meurtrier a été arrêté et sera jugé, des voyous ont manifesté
et cassé en toute impunité.
(10b) Seul signe compatible avec l'existence d'un processus
infectieux, le pouls des cyclistes était  le matin de leur
abandon assez accéléré.
```

Nous voudrions plutôt insister sur un autre point en rapport avec la référence. La sémantique de ces tournures hésite plus entre la série concrète dont le cadre est un intervalle englobant bien repéré, et une signification plus abstraite (évoquant une grille ou un emploi du temps par exemple) dont l'application sur l'axe du temps réel est sous-déterminée. La forme générique marque une place dans une matrice temporelle type[3], et la régularité est une conséquence seconde de la régularité avec laquelle cette matrice est incarnée. Cette distinction explique la possibilité pour le procès de ne pas occuper toutes les occurrences de l'intervalle de répétition. On opposera de ce point de vue (11a, b) à (11c, d) : l'incarnation est dans les deux premiers cas épisodique et seuls certains soirs sont pertinents ; elle est générale dans (11c) et universelle dans (11d) dont le présent a une valeur déontique marquée.

```
(11a) Discrètement maquillée de beige et de rose, elle me
dira s'ensanglanter la bouche et se noircir les cils pour
sortir le soir avec son boy friend.
(11b) Votre mère n'ayant pas les moyens de s'offrir une TSF,
vous allez, le soir, écouter " Ici Londres " chez un voisin
(11c) Je reviens chez moi le soir vers 20 heures.
(11d) Le Parc, …, est ouvert de 10 heures à 18 heures (21
heures le samedi, 20 heures du 25 décembre au 3 janvier, 1
heure le 31 décembre).
```

Si la séparation de la matrice et de son incarnation est particulièrement évidente dans une partie des tournures génériques, cette hypothèse explique facilement certaines interprétations dans les autres tournures. Celles que nous citons ici n'ont pas d'incarnation dans une réalité particulièrement référencée (par exemple, on ne pourrait pas dater les références multiples) :

[3] La distinction entre une matrice temporelle structurée et ce qui est dit de ses instanciations a été introduite  dans les travaux du groupe par Yann Mathé sous la forme d'un procès-modèle.

(12a) Ce minimum vital, versé au début de chaque mois, permet de payer la redevance à la Sonacotra
(12b) Un bon marcheur parcourt 35 kilomètres par jour en savates avec des dénivellés effrayants et des passages très dangereux, explique Philippe Dupuich,…
(12c) Malgré les hésitations, les mots que l'on cherche en français, ils inventent : l'histoire de Ti Paul, ouvrier d'usine, qui rencontre Merchat, personnage un peu trouble, moitié clochard - moitié sorcier, qui fait dentelle devant la boutique ", autrement dit qui boit trop, qui bat sa femme et qui se sauve, tous les samedis, avec l'argent des allocations ".
(12d=11d)) Le Parc, …, est ouvert de 10 heures à 18 heures (21 heures le samedi, 20 heures du 25 décembre au 3 janvier, 1 heure le 31 décembre).
(12e) Deux décrets seront publiés avant la fin du mois de mars pour organiser ce contrôle, qui obligera tous les véhicules de plus de cinq ans à se présenter tous les trois ans dans l'un des quinze cents centres qui seront agréés.
(12f) pour les militaires, les satellites offrent, outre leur discrétion, une série d'avantages qui tiennent à leur grande autonomie et à leur capacité de survoler à intervalles réguliers (environ une fois par jour) tout point du globe et d'observer notamment ce qui a changé entre deux passages.
(12g) Parmi les productions en cours : … la mégasérie Riviera …, 260 épisodes de vingt-six minutes, dont une soixantaine déjà tournés (un épisode par jour, six jours sur sept !)

Le cadre dans lequel se situe l'itération n'est ni situé, ni borné, par exemple parce que l'énoncé a une signification modale déontique (12 d,e) ou aléthique (12f), une signification typique (12a, b), qu'il décrit une histoire dans l'histoire et réfère un temps virtuel (12c), ou qu'il décrit une structure temporelle attachée à l'objet 'série télévisée' avant même toute diffusion (12g). Cette potentialité du cadre dans lequel se situe l'itération à être posé dans un temps parallèle est, nous semble-t-il, très proche de la capacité de la série à se constituer en procès illustrée dans (9). Nous en donnons un autre exemple, où l'apparente contradiction des références temporelles se résout parce que le circonstant temporel *dans la matrice* réfère en fait l'ensemble des instanciations de cette matrice *dans le cadre* (ici non borné).

(12h) De 6 heures du matin à minuit, dans Milan et trente-quatre communes environnantes, les véhicules immatriculés en Lombardie ne seront autorisés à rouler qu'un jour sur deux, selon que leur numéro d'immatriculation se termine par un chiffre pair ou impair.

## Structures complexes et portée

Dans certains cas, la matrice de l'itération est elle-même une structure complexe : un ensemble ou une succession de procès, ou un procès lui-même itéré. Un premier exemple mettra en évidence le phénomène :

(13a) A raison de trois heures hebdomadaires pendant trente-cinq semaines par an durant sept ans, un élève d'une classe de vingt-cinq élèves aura pu pratiquer effectivement une langue étrangère pendant une douzaine d'heures, si l'on tient compte du temps pris par l'enseignant pour prodiguer explications et conseils.

L'adjectif hebdomadaire ne fait pas partie des marqueurs qui nous ont servi à sélectionner nos exemples, mais l'enchâssement des modèles itérés est limpide. Dans beaucoup d'exemples, les relations entre les différentes parties de la phrase ou du texte sont moins explicites et résultent d'inférences. Les quelques exemples qui suivent illustrent plusieurs mécanismes inférentiels.

(14a) Les hommes qui, chaque matin, font l'embauche, arborent, glissées dans le ruban de leur chapeau, des allumettes qui indiquent la somme qu'ils prélèvent sur la paie

(14b) Comme chaque mercredi, à l'aube, quelques centaines de personnes se retrouvent au pied du Transsibérien et s'installent tant bien que mal pour six jours de rail qui les conduiront à Moscou.

(14c) On embarque à Amsterdam le lundi à partir de 19 heures et on débarque à Strasbourg le vendredi vers 15 heures. Il en coûtait en 1992, par personne, 4 056 F (pont inférieur) et 4 753 F (pont principal), de la mi-avril à fin mai ; 5 070 F et 5 941 F de début juin à la mi-octobre

(14d) A l'initiative d'une vingtaine de bénévoles qui avaient déjà travaillé à la distribution des vivres après les inondations, 3 500 paniers-repas vont ainsi être distribués chaque semaine. Le centre ne sera ouvert que le mercredi après-midi et le samedi matin ; les sinistrés pourront venir chercher, une fois par semaine, l'équivalent de sept repas (riz, pâtes, lait, fruits...) à consommer à domicile.

(14e) A Paranoa, …, 3 000 enfants seulement fréquentent le CIAC ou les baraquements de la traditionnelle école publique, où l'on accueille les enfants trois ou quatre heures par jour seulement.

(14f)Le nombre de lecteurs s'accroît, mais la quantité moyenne de livres lus et achetés chaque année fléchit

(14g) Le Comité international de la Croix-Rouge (CICR), qui, deux fois par jour, distribue des repas à 1 500 000 Somaliens, a dû réduire ses rations de deux tiers, car il est en rupture de stocks depuis neuf jours.

On trouve dans (14a) un exemple très net de ce qu'on pourrait appeler la *propagation* de l'itération : la portée syntaxique du marqueur 'chaque matin' est limitée à la subordonnée 'font l'embauche', alors que sa portée sémantique inclut la principale 'arborent des allumettes'. Le procès de la subordonnée 'indiquent la somme' est moins évidemment itératif, alors que 'prélèvent sur la paye', qui lui est syntaxiquement subordonné, récupère ce caractère. On peut faire deux remarques. Tout d'abord l'hypothèse de la matrice d'itération explique (sans pour autant suffire à en rendre compte exactement) la propagation de l'itération. On retrouve en effet les mêmes problèmes de portée de marqueurs temporels en dehors de l'itération, ex : "au moment où il franchit le seuil, il fait le signe de croix qui protège du mauvais esprit" (la protection est une propriété permanente). Ensuite, il est intéressant de faire varier la place du complément temporel :
(14a') Les hommes qui font l'embauche arborent chaque matin …
(14a") Chaque matin, les hommes qui font l'embauche arborent …
La propagation semble se faire uniquement vers l'avant, au moins dans ce cas, et être peu influencée par la structure syntaxique.
Dans l'exemple (14b), l'intervalle itéré 'chaque mercredi' se propage à travers la coordination ('et s'installent…'), mais aussi au procès prospectif qui déborde de l'intervalle de répétition 'six jours de rail qui les conduiront à Moscou'. N'importe quel système automatique devra résoudre cette apparente contradiction.
(14c) présente au contraire une apparente double itération qui en réalité signale deux moments du même procès : embarquer et débarquer sont le début et la fin d'un voyage ; leurs occurrences sont appariées et évoquent chacune le procès dans son entier – celui-la même qui est repris anaphoriquement à la phrase suivante ('il **en** coûtait …').
(14d) et (14e) présentent deux versions différentes de l'influence de la référence sur l'itération : c'est le *nombre* de paniers repas qui se répète chaque semaine, mais bien entendu chacun n'est distribué qu'une fois ; alors que le verbe 'fréquentent' conduit à lire la suite 'chaque enfant est accueilli chaque jour'. Le caractère collectif (tous aux mêmes horaires) ou individuel (étalés sur la journée) de la prédication est sous déterminé. Dans (14d), l'itération se propage aux deux phrases suivantes qui détaillent la matrice de la semaine. L'apparente contradiction entre l'ouverture du centre le mercredi et le samedi, et la distribution un jour par semaine, se résout par l'interprétation de 'pouvoir' : la potentialité existe deux fois par semaine, mais il faut comprendre ici que c'est l'autorisation qui est donnée couvre une réalisation par semaine.
Alors que la syntaxe de (14f) fait porter le marqueur d'itération sur les deux adjectifs qui précèdent (les livres lus et achetés chaque année), il porte en fait sur la taille des périodes de relevé statistique des quantités lues et achetées. L'exemple comporte aussi un phénomène de portée rétrograde du marqueur d'itération à travers une coordination : le nombre de lecteurs s'accroît *chaque année*, ce qui n'est pas évident à la lecture de la première proposition seule.
Dans (14g), l'interprétation référentielle atteint sa limite. Cet extrait décrit une situation collective et l'intention de l'auteur n'est sans doute pas de la détailler. Ce qui est distribué deux fois par jour est sous-déterminé : on peut lire 'deux distributions organisées par jour' ou bien 'reçues par chaque bénéficiaire', mais le nombre d'acteurs et la complexité du déroulement réel ne sont guère compatibles avec ces interprétations, et l'itération vise sans doute un principe d'organisation de la tâche. L'essentiel du message est dans le caractère dramatique de la situation et la difficulté d'y faire face.

# Conclusion

Les tournures itératives que nous avons rassemblées, bien qu'elles ne constituent que le fragment le plus accessible à une extraction automatique, soulèvent plusieurs questions passionnantes pour le TAL. C'est l'objet des articles qui suivent dans ce rapport d'en explorer au moins en partie les solutions. Du coté de la linguistique, il s'agit de compléter l'inventaire partiel fait ici et de reconnaître les éléments qui déterminent l'interprétation de ces formes : que peut-on dire de la répartition temporelle des instances, de leur structure aspectuelle ? Comment constituer ce que nous avons appelé ici la matrice de répétition ? Son caractère virtuel relève-t-il de structures de signification et d'interprétation semblable à celles qui rendent compte des modalités ? Ces mécanismes aspectuo-temporels éclairent-ils la détermination du procès focal et la propagation de l'itération ? Du coté du calcul, il y a besoin d'une représentation qui exprime les relations entre intervalles et entre les procès qu'ils supportent sans imposer un détail référentiel inadéquat ; nous avons considéré plusieurs formalismes.
Au delà de ce qui est fait, les germes sont aussi semés d'un travail à poursuivre, et qui se poursuivra. Tant sur le plan linguistique que sur celui de la représentation symbolique, beaucoup de questions restent ouvertes, et cela apparît dans notre bilan. Sur le plan expérimental, le travail peut être mieux outillé, pour étudier plus systématiquement les phrases extraites et y chercher des régularités. Et l'extraction systématique d'exemples mérite d'être poursuivie pour d'autres indices plus difficiles : il y a bien sûr les verbes (répéter), et nous avons rencontré dans les textes cités ici même des exemples d'adjectifs (hebdomadaire) ou de nom (les repas, un feuilleton) dont la sémantique porte en elle-même des éléments de répétition.

**Bibliographie**

# L'itération dans le modèle SdT

Laurent Gosselin

**Sommaire :**

# 1. Présentation du modèle

## 1.1. Le format de représentation

### *1.1.1. Des structures d'intervalles*

Les structures aspectuo-temporelles utilisées dans le modèle SdT mettent en œuvre quatre types d'intervalles disposés sur l'axe temporel : l'intervalle d'énonciation [01,02], l'intervalle du procès [B1,B2], l'intervalle de référence (ou de monstration) [I,II], et l'intervalle circonstanciel [ct1,ct2]. L'intervalle d'énonciation indique les limites temporelles de l'acte physique d'énonciation, l'intervalle du procès correspond à une opération de catégorisation (*i.e.* la subsomption d'une série de changements et/ou de situations sous la détermination d'un procès). L'intervalle de référence [I,II] est lié à une opération de monstration (il correspond à ce qui est perçu/montré du procès, par exemple ce qui est asserté lorsque l'énoncé est assertif[4]). L'intervalle circonstanciel est marqué par les compléments de localisation temporelle (ex. *mardi dernier*) et les compléments de durée (ex. *pendant trois heures*). À chaque énoncé est associé un et un seul intervalle d'énonciation [01,02]; à chaque proposition (principale, subordonnée ou indépendante) sont associés au moins un intervalle de procès [B1,B2] et au moins un intervalle de référence [I,II]; à chaque circonstanciel temporel correspond au moins un intervalle circonstanciel [ct1,ct2]. Exemple :

(1) *Luc avait terminé son travail depuis deux heures*

[4] Il constitue l'équivalent du *Topic Time* de W. Klein (1994).

fig.1

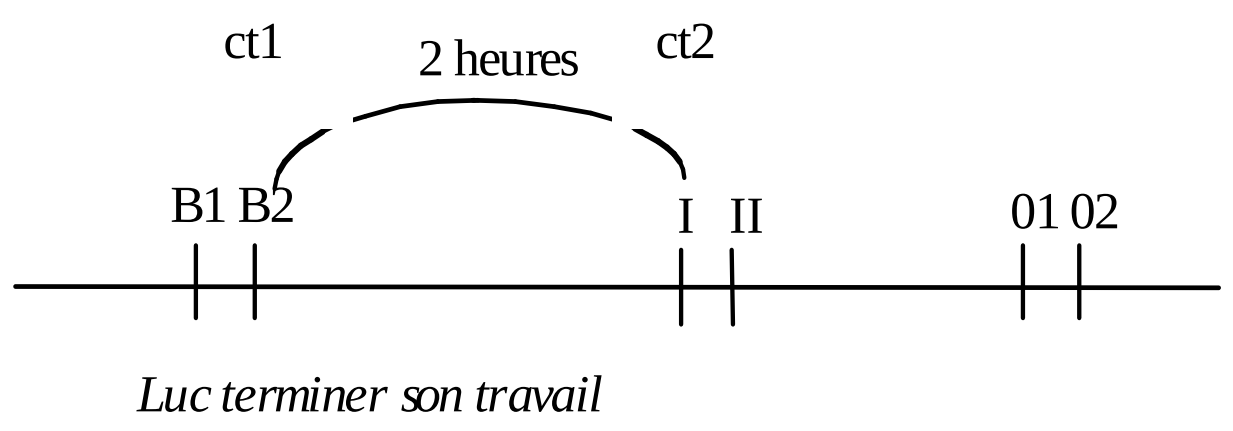

En énonçant (1), le locuteur parle d'un certain moment (le moment de référence, noté [I,II]) et situe le procès « *Luc terminer son travail* » deux heures avant ce moment de référence, qui est lui même situé dans le passé (il est antérieur au moment de l'énonciation).

Pour représenter les phrases complexes, on duplique l'axe temporel pour chaque proposition subordonnée. Ainsi à l'énoncé :

(2) *Hier, Pierre m'a raconté qu'il était allé à la pêche samedi dernier*

on associe la structure[5] :

fig.2

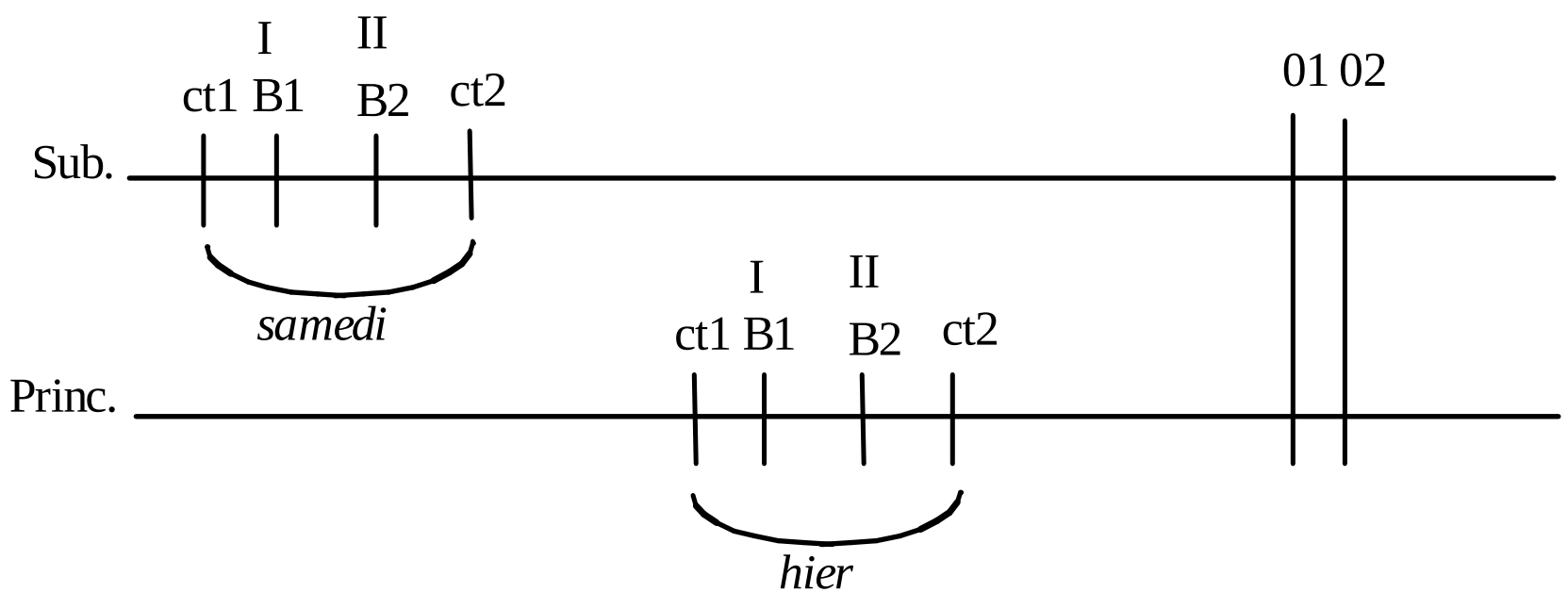

### *1.1.2. Définitions*

Nous adoptons les définitions suivantes :

*1.2.2.1. Relations entre bornes et entre intervalles*

Soit i, j, k, des bornes quelconques d'intervalles quelconques (éventuellement du même intervalle),

i = j [coïncidence]

[5] Précisons que les représentations iconiques constituent, au mieux, des approximations des structures aspectuo-temporelles, qui ne peuvent être exactement appréhendées que par les représentations symboliques.

$i \propto j$ [i précède j, mais en est infiniment proche; la précédence est immédiate]
$i \prec j$ [i précède j, mais ne se trouve pas dans son voisinage immédiat]
$i < j =_{df} (i \propto j) \vee (i \prec j)$ [i précède j de façon immédiate ou non]
$i \leq j =_{df} (i < j) \vee (i = j)$ [i précède (de façon immédiate ou non) ou coïncide avec j].

Par définition, pour tout intervalle [i,j], $i < j$, ce qui veut dire que tout intervalle a une durée, aussi infime soit-elle.

Les relations entre intervalles se laissent exprimer à partir des relations entre bornes. Soit deux intervalles [i,j] et [k,l], il nous a paru utile de définir les relations suivantes pour le traitement des phénomènes linguistiques :

[i,j] ANT [k,l] $=_{df}$ $j < k$ **antériorité**

[i,j] POST [k,l] $=_{df}$ $l < i$ **postériorité**

[i,j] SIMUL [k,l] $=_{df}$ $(i \leq l)$ & $(k \leq j)$ **simultanéité**

A ces trois relations mutuellement exclusives; on adjoint cinq relations appelées à jouer un rôle important pour l'analyse linguistique :

[i,j] RE [k,l] $=_{df}$ $(i < k)$ & $(j > l)$ **recouvrement**[6]

[i,j] CO [k,l] $=_{df}$ $(i = k)$ & $(j = l)$ **coïncidence**

[i,j] ACCESS [k,l] $=_{df}$ $(i \leq k)$ & $(j \geq l)$ **accessibilité** (on peut accéder à [k,l] à partir de [i,j])[7]

[i,j] SUCC [k,l] $=_{df}$ $k < i$ **succession**

[i,j] PREC [k,l] $=_{df}$ $i < k$ **précédence**

*1.1.2.2. Aspect lexical*

L’aspect lexical correspond au « type de procès » marqué par le verbe et son environnement actanciel. Il s’agit du procès tel qu’il est « conçu », alors que l’aspect grammatical définit la façon dont il est « montré/perçu ».

A la suite de [VEN 67] et de la vaste littérature qu’il a suscitée, nous mettons en œuvre trois critères pour classer les procès : la dynamicité, le bornage et la ponctualité. Leur attribution repose – schématiquement  sur les tests suivants :

Un prédicat exprime un procès dynamique s’il est compatible (sans changement de sens) avec *« être en train de* Vinf. » : *être en train de parler*/*manger une pomme*/ ?* *habiter une maison*/?* *être petit*. (Remarque : ce test ne s’applique pas aux procès ponctuels, qui sont dynamiques par définition).

Un prédicat désigne un procès borné de façon extrinsèque (atélique) [JAC 92] si, conjugué au passé composé, il est compatible (sans changement de sens) avec « *pendant* + durée » et non avec « *en* + durée » : *Il a marché pendant deux heures*/ ?**en deux heures* ; *il a été malade*

---

6  Cette relation correspond à la relation «*during*» inversée (di) dans le modèle de J.F. Allen (1984).
7 Il s'agit en fait de la relation d'inclusion inversée. Le terme d'accessibilité correspond cependant mieux au processus cognitif que cette relation va servir  à circonscrire.

*pendant dix mois*/ ?* *en dix mois*. Dans le cas contraire, le procès sera tenu pour intrinsèquement borné (télique) : *Il a terminé son article en deux heures*/ ?* *pendant deux heures*.

Un prédicat dénote un procès ponctuel si la construction « *mettre* n temps *à* Vinf. » , conjuguée au passé composé, équivaut à « *mettre* n temps *avant de* Vinf » : *il a mis deux heures à trouver la solution ≈ avant de trouver la solution* ; *il a mis une heure à recopier sa rédaction ≠ avant de recopier sa rédaction*. (Remarque : ce test ne s'applique pas aux procès atéliques, qui sont non ponctuels par définition).

Même s'il existe des cas intermédiaires, et si l'application des tests demande certaines précautions (cf. Gosselin 1996, pp. 41-72), leur mise en œuvre conduit à classer les différents procès selon les quatre grandes catégories de la typologie de Vendler (1967) :

les **états** sont non dynamiques, atéliques et non ponctuels (*habiter une maison*, *être en vacances*) ;

les **activités** sont dynamiques, atéliques et non ponctuelles (*marcher*, *manger des frites*, *dormir*[8]) ;

les **accomplissements** sont dynamiques, téliques et non ponctuels (*manger une pomme*, *terminer un travail*) ;

les **achèvements** sont dynamiques, téliques et ponctuels (*apercevoir un avion*, *atteindre un sommet*, *trouver une solution*).

Mentionnons que dans le modèle proposé, la télicité (le fait que le procès soit orienté vers une fin) est indiquée par un typage des bornes du procès : bornes « extrinsèques » pour les procès atéliques (états et activités), bornes « intrinsèques » pour les procès téliques (accomplissements et achèvements), tandis que la ponctualité est marquée par le choix de la relation entre les bornes du procès :

Procès ponctuel (achèvement): B1 ∝ B2

Procès non ponctuel (état, activité ou accomplissement) : B1 ≺ B2

***1.1.2.3. Aspect grammatical*** *(ou « point de vue aspectuel*[9] *»)*

C'est la relation entre l'intervalle de référence (de monstration) et celui du procès qui définit l'**aspect grammatical**. On distingue quatre aspects de base en français :

> Avec l'aspect **aoristique** (perfectif), justement qualifié dans la perspective guillaumienne d'"aspect global"[10], le procès est montré dans son intégralité (les deux intervalles coïncident) : I = B1, II = B2; ex. : *Il traversa le carrefour*[11].

---

[8] Le fait que *dorm*ir ou *attendre* soient à classer parmi les activités montre bien qu'il s'agit d'une classification purement linguistique et non référentielle.

[9] Cf. C. Smith (1991).

[10] Cf. M. Wilmet (1980) et (1991).

[11] On fait pas figurer ici le moment de l'énonciation, car on veut ne représenter que l'aspect grammatical.

fig. 3

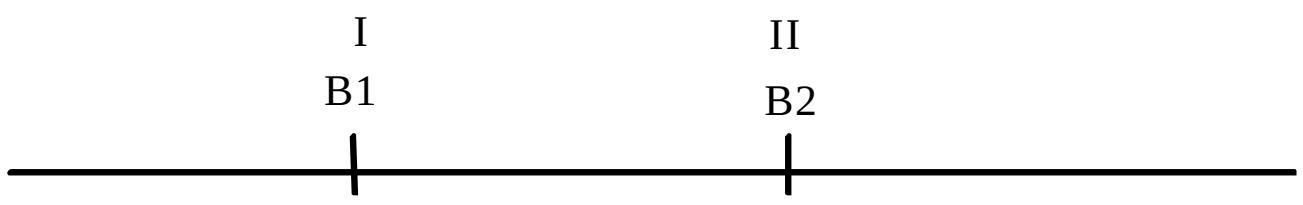

L'aspect **inaccompli** (imperfectif) ne présente qu'une partie du procès (« aspect sécant ») : l'intervalle de référence est inclus dans celui du procès, les bornes initiale et finale ne sont pas prises en compte : B1 < I, II < B2 ; ex. : *Il traversait le carrefour* (l'imparfait doit être interprété ici au sens de *était en train de traverser*).

fig.4

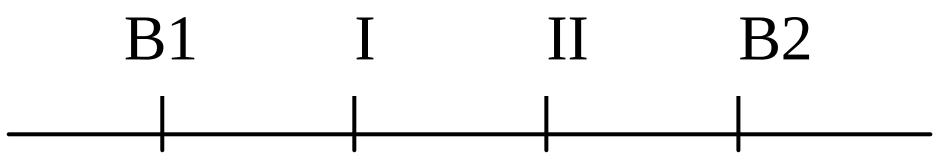

L'aspect **accompli** montre l'état résultant du procès : B2 < I; ex. : *Il a terminé son travail depuis dix minutes.*

fig.5

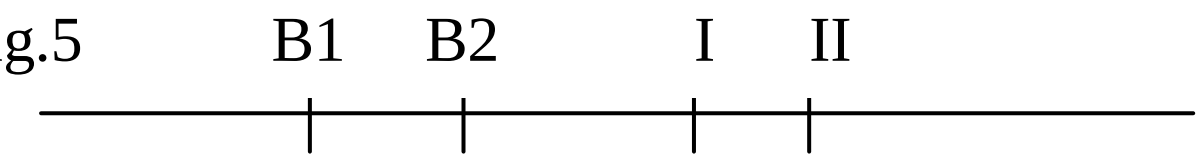

L'aspect **prospectif**[12] en présente la phase préparatoire : II < B1; ex. : *Il allait traverser le carrefour.*

fig .6

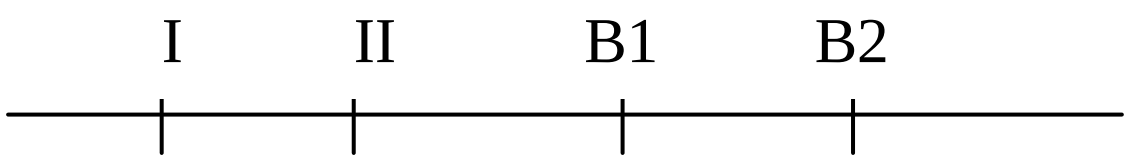

*1.1.2.4. Temps absolu*

Le **temps** lui-même reçoit une nouvelle définition : il s'agit de la relation entre l'intervalle de référence et le moment de l'énonciation (considéré lui-même comme un intervalle : [01,02]).

Temps **présent** : les deux intervalles coïncident ou se chevauchent : I ≤ 02, 01 ≤ II

Temps **passé** : II < 01

Temps **futur** : 02 < I.

Cette définition du temps, qui s'oppose à la conception traditionnelle selon laquelle c'est la position du procès par rapport au moment de l'énonciation qui constitue le temps absolu, permet d'éviter l'indécidabilité qui devrait logiquement affecter un grand nombre de relations temporelles dès lors que l'on substitue des intervalles aux points. Par exemple, en énonçant : *Il y a dix minutes (quand je suis sorti), il pleuvait,* le locuteur n'indique en rien si le procès (la pluie) a cessé ou non au moment de l'énonciation. Le procès a certes une partie passée, mais il

[12] Cf. E. Benveniste (1966), p. 239.

se peut très bien qu'il se poursuive dans le présent et même dans le futur. La seule information sûre, c'est qu'au moment de référence, situé dans le passé, et localisé grâce au circonstanciel, le procès était en cours (aspect inaccompli). D'où la représentation iconique :

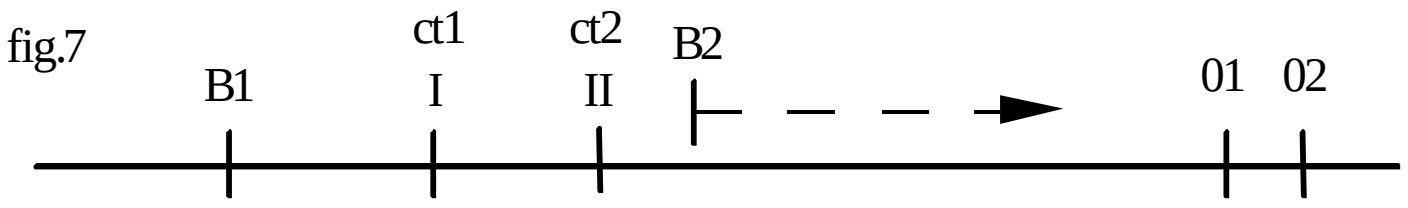

De même avec l'énoncé « *Quand j'ai regardé par la fenêtre, il allait pleuvoir* », on sait que l'intervalle de référence est antérieur à celui de l'énonciation (temps passé) et à celui du procès (aspect prospectif), mais aucune contrainte linguistique ne porte sur les relations chronologiques entre le procès et le moment de l'énonciation.

*1.1.2.5. Temps relatif*

Le temps relatif se trouve désormais défini non plus comme la relation entre deux procès, mais comme la relation entre deux intervalles de référence (notés respectivement [I,II] et [I’,II’]) :

**Simultanéité** : I ≤ II’, I’ ≤ II

**Antériorité** : II < I’

**Postériorité** : II’ < I.

Exemple :

(3) *Il a dit qu’il était venu la veille*

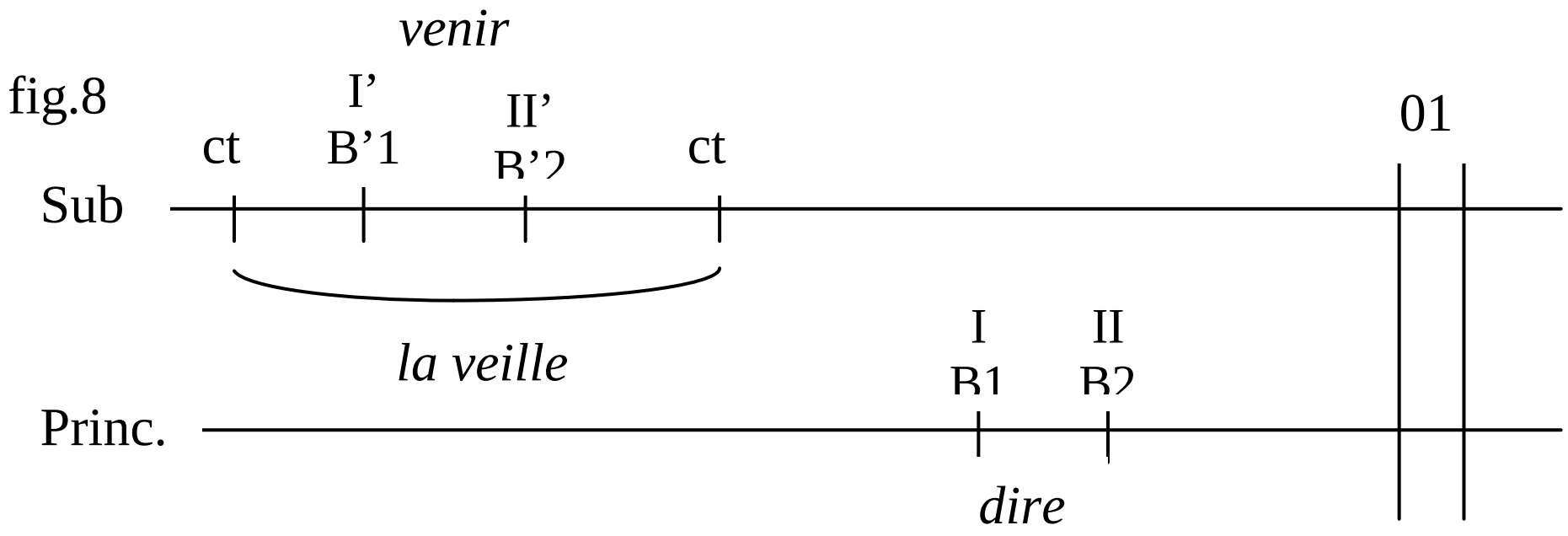

Les relations entre le procès et le moment de l’énonciation, ainsi qu’entre procès différents, sont obtenues indirectement, en tenant compte de l'aspect. Exemples :

(4) *Il sortit son portefeuille*
[I,II] coïncide avec [B1,B2] (aspect aoristique)
[I,II] est antérieur à [01,02] (temps passé)
donc [B1,B2] est antérieur à [01,02].

(5) *(Quand j'ai été le voir) le petit dormait*
[I,II] est inclus dans [B1,B2] (aspect inaccompli)
[I,II] est antérieur à [01,02] (temps passé)

donc B1 est antérieure à 01, mais la relation entre B2 et [01,02] n'est pas linguistiquement contrainte (le procès peut très bien se poursuivre dans le présent et même dans le futur)[13].

(6) *Marie croyait que Paul viendrait*
[I,II] est inclus dans [B1,B2] (aspect inaccompli dans la principale)
[I,II] est antérieur à [01,02] (tps passé)
[I',II'] coïncide avec [B'1,B'2] (aspect aoristique dans la complétive)
[I',II'] est postérieur à [I,II] (temps relatif : ultérieur)
donc la position de B2 par rapport à [01,02] reste linguistiquement indéterminée (le croit-elle encore ?); la position de [I',II'] et par conséquent celle de [B'1,B'2] par rapport à [01,02] le sont aussi[14].

Au plan du texte, on cherche à calculer les relations entre bornes appartenant à des propositions et à des phrases différentes. Soit pour exemple :

fig.9

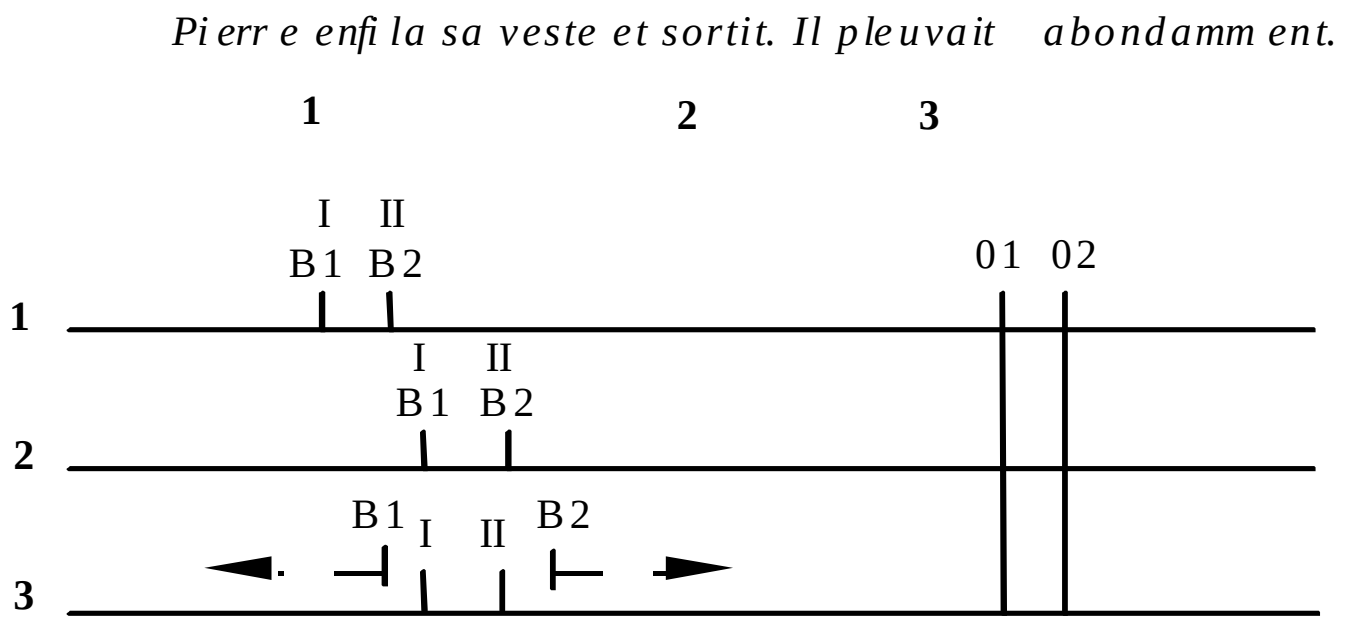

Les flèches associées au procès « *il pleuvait* » indiquent que la localisation de ses bornes relativement aux autres éléments de la structure reste partiellement indéterminée. On ne sait pas, par exemple, si la pluie avait déjà commencé quand Pierre a enfilé sa veste[15].

Remarquons qu'il est très important, lorsqu'on analyse les phénomènes temporels dans les textes, de préciser quelles relations sont contraintes et quelles relations restent indéterminées, l'indétermination relative étant une propriété essentielle de la sémantique des textes.

Ces structures sont obtenues à partir des instructions codées par les marqueurs de temps et d'aspect (morphèmes lexicaux et grammaticaux, structures syntaxiques) ainsi que par des contraintes pragmatico-référentielles.

## 1.2. Les principes de calcul : la compositionnalité holiste

On distingue, pour le calcul sémantique, deux principes, qui ne sont pas exclusifs l'un de l'autre : le principe de compositionnalité, selon lequel la signification du tout est déterminée par celles de ses parties, et le principe de contextualité, qui pose, à l'inverse, que la signification des parties et déterminée par celle du tout dans lequel elles se trouvent intégrées.

---

[13] Cf. la représentation iconique ci-dessus (fig.7).
[14] C'est sur la base de tels exemples qu'il nous a paru nécessaire d'abandonner la définition du temps absolu comme relation directe entre le procès et le moment de l'énonciation, ainsi que la définition du temps relatif comme relation directe entre les deux procès.
[15] Nous n'avons, pour simplifier, assigné qu'un seul intervalle d'énonciation pour l'ensemble de la séquence. En toute rigueur, il faudrait distinguer trois intervalles contigus.

Une théorie est dite « compositionnelle atomiste » si elle ne retient que le principe de compositionnalité (à l'exclusion du principe de contextualité). La nôtre sera dite « compositionnelle holiste », dans la mesure où elle articule simultanément les deux principes (cf. Gosselin 2005).

On admet que chacun des marqueurs aspectuo-temporels code une ou plusieurs instruction(s) pour la construction d'intervalles ou de relations entre bornes sur l'axe temporel. Ces instructions constituent la part aspectuo-temporelle de la valeur en langue (hors contexte) du marqueur. La composition sémantique réside alors dans l'assemblage de ces divers éléments des structures sémantiques (intervalles, relations entre intervalles, relations entre bornes). Le but de cet assemblage est d'abord d'obtenir une structure globale qui soit cohérente) ; or il arrive très souvent que différentes instructions codées par un même énoncé soient contradictoires (au sens où les éléments de structure à construire sont incompatibles). Nous avons montré (dans Gosselin 1996) que ces cas de conflit sont résolus par la mise en œuvre de modes de résolution de conflit, qui consistent à déformer – le moins possible – les structures globales selon des procédures régulières et prédictibles, de façon à satisfaire à toutes les exigences (correspondant aux instructions codées par l'énoncé ou dépendant de principes généraux sur la bonne formation des structures). Exemples :

– conflit insoluble (pour lequel il n'existe pas de mode de résolution) :

> (7) * Il marcha depuis deux heures
>
> le passé simple marque l'aspect aoristique, c'est-à-dire la coïncidence de [I,II] avec [B1,B2] ; [*depuis* + durée] indique au contraire que B1 précède I et mesure cet écart (comme dans « *Il marchait depuis deux heures quand je l'ai rencontré* »).

– conflit résolu par la contraction du procès sur sa borne initiale :

> (8) Il dormit à 10h40
>
> dormir : activité : B1 ⎨ B2
>
> passé simple : aspect aoristique : I = B1, II = B2
>
> à 10h40 : circonstanciel ponctuel : ct1 ∝ ct2
>
> construction syntaxique du circonstanciel : intégré au SV : ct1 ≤ B1, B2 ≤ ct2
>
> d'où ct1 = B1 ∝ B2 = ct2, ce qui est incompatible avec l'aspect non ponctuel marqué par dormir (B1 ⎨ B2)
>
> résolution du conflit par déformation de la structure : le procès se contracte sur sa phase initiale, ponctuelle, et dormir équivaut à s'endormir.

Résumons-nous : à chaque marqueur est(sont) associée(s) une ou plusieurs instruction(s) pour la construction d'éléments constitutifs de la structure sémantique globale ; ces instructions sont considérées comme représentant les valeurs en langue des marqueurs, et constituent des entrées stables pour un système de calcul. Plongées dans un contexte, soit elles ne rencontrent aucun conflit et donnent alors lieu aux effets de sens typiques, soit elles entrent en conflit avec d'autres instructions ou avec des contraintes pragmatico-référentielles, et les conflits sont résolus au moyen de procédures régulières de déformation des structures, qui donnent lieu à des effets de sens dérivés. C'est parce que, dans ce modèle, les conflits apparaissent très fréquents et sont résolus de façon régulière qu'un calcul sémantique de la PCG est possible et qu'il peut être considéré comme holiste (ainsi l'itération dans (8) n'est pas prise pour une valeur propre à l'imparfait, mais pour le résultat d'une résolution de conflit – au même titre que l'itération dans « *Longtemps, je me suis couché de bonne heure* »).

# 2. Représentations de l’itération

## 2.1.Une représentation au moyen d’intervalles

La représentation de l’itération combine deux intervalles de procès, l’un qui correspond à la série de procès réitérés ([Bs1,Bs2], l’autre au « procès modèle » ([B1,B2]), i.e. au modèle d’occurrence de procès. A ces deux intervalles de procès sont associés deux intervalles de référence, l’un qui porte sur la série tout entière ([Is,IIs]), l’autre qui porte sur le procès modèle ([I,II]). Des rapports entre ces quatre intervalles résultent les relations de temps et d’aspect.

De façon générale, dans le modèle SdT le temps (absolu) est défini par la relation entre l’intervalle de référence et l’intervalle d’énonciation ; tandis que l’aspect correspond à la relation entre l’intervalle du procès et l’intervalle de référence. Concernant l’itération, le temps correspond à la relation entre [Is,IIs] et [01,02]. En revanche, l’aspect de l’itération présente un caractère spécifique, dans la mesure où il se calcule à la fois relativement au procès modèle et à la série prise globalement. D’où la possible coexistence de plusieurs aspects simultanément. Exemple :

(9) *Depuis deux mois, il mangeait en dix minutes*

fig.10

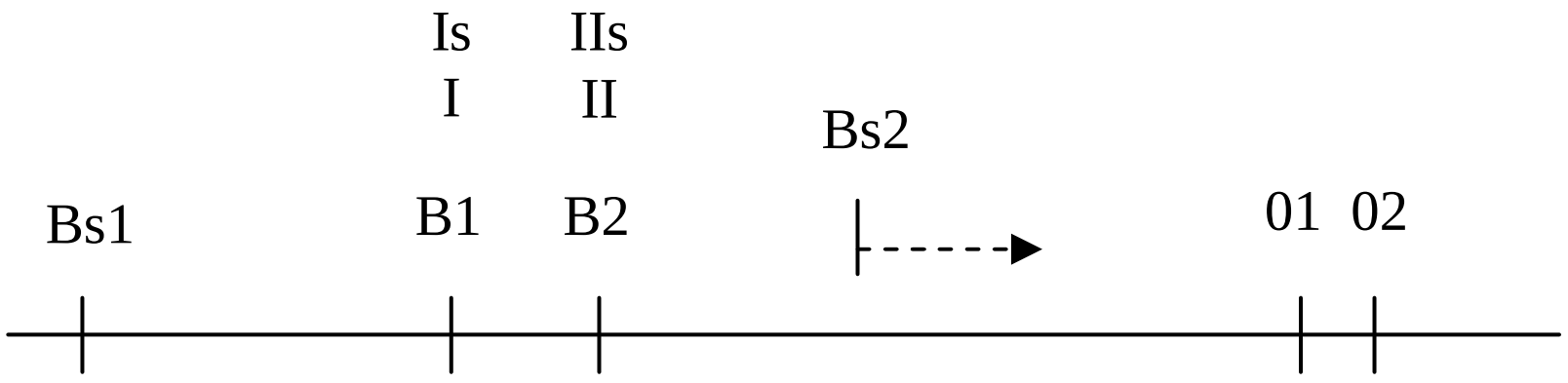

L’aspect est aoristique sur les occurrences de procès (sur le procès modèle), et inaccompli sur la série ; d’où la compatibilité de circonstanciels apparemment contradictoires : *en dix minutes* évalue la distance entre B1 et B2 (la durée de chaque occurrence de procès), tandis que *depuis deux mois*, porte sur l'intervalle qui sépare Bs1 (le début de la série) de I (le moment considéré).

## 2.2. La série itérative comme type de procès complexe

Dans ce modèle, la série itérative ([Bs1,Bs2]) est conçue comme un type de procès complexe, lui-même composé d'une **série de procès** semblables (qui a le statut d'une série de changements). Les bornes de la série itérative peuvent être soit extrinsèques (elle a alors un statut d'activité), soit intrinsèques (elle a valeur d'accomplissement). Un procès itératif est borné de façon extrinsèque lorsque le nombre d'occurrences de procès n'est pas déterminé (on parle de **fréquence**); il est intrinsèquement borné dans le cas contraire (on parle de **répétition**) :

(10a) *Pendant une heure/?* en une heure, Marie a sautillé*

(10b) *Pendant deux jours/?* en deux jours, Marie a souvent éternué*

(10c) *Pendant deux heures/?* en deux heures, Marie a mangé des gâteaux*

(11a) *En une heure/?* pendant une heure, Marie a éternué vingt-deux fois*

(11b) *En deux heures/?* pendant deux heures, Marie a mangé cinq gâteaux*

Remarquons que lorsque le circonstanciel est placé en fin de phrase, il peut être interprété comme portant sur chacune des occurrences de procès ou bien sur l'ensemble de la série itérative. Comparons les énoncés :

(12a) *Marie a fait trois fois le tour du parc en cinq minutes*

(12b) *?* Marie a fait trois fois le tour du parc pendant cinq minutes*

(13a) *Marie a couru trois fois en cinq minutes*

(13b) *Marie a couru trois fois pendant cinq minutes*

(14a) *Marie a fait des tours de parc en un quart d'heure*

(14b) *Marie a fait des tours de parc pendant un quart d'heure*

L'énoncé (12a) est virtuellement ambigu; le circonstanciel, compatible uniquement avec les bornes intrinsèques, peut servir aussi bien à indiquer la taille de l'intervalle associé à la série tout entière (qui est intrinsèquement bornée), qu'à marquer la taille des intervalles associés à chacun des procès qui composent la série itérative (ces procès étant des accomplissements). En revanche, (12b) paraît ininterprétable, car [*pendant* + durée] exclut les bornes intrinsèques. Dans les exemples (13a), (13b), la série itérative reste intrinsèquement bornée, mais les procès qui la constituent ne le sont plus (*courir* désigne une activité); aussi le circonstanciel *en cinq minutes* ne peut-il porter que sur la série itérative, tandis que *pendant cinq minutes* indique nécessairement la durée de chacune des occurrences de procès. Le phénomène s'inverse avec (14a) et (14b), puisque, cette fois, la série n'est pas intrinsèquement bornée, quoique chacun des procès qui la composent le soit : en (14a), le circonstanciel ne peut porter que sur les occurrences de procès (≈ *Il est arrivé à Marie de faire des tours de parc en un quart d'heure*); en (14b), il porte nécessairement sur la série tout entière.

## 2.3. Séries itératives de procès différents

L'itération peut aussi porter non plus sur des occurrences individuelles de procès identiques, mais aussi sur des séries de procès différents et successifs. On obtient une itération de séries de procès :

(15a) *A huit heures du matin, il* ***descendait*** *des hauteurs de Montmartre, pour prendre le vin blanc dans la rue Notre-Dame-des-Victoires. Son déjeuner (...) le* ***conduisait*** *jusqu'à trois heures. Il se* ***dirigeait*** *alors vers le passage des Panoramas, pour prendre l'absinthe. Après la séance chez Arnoux, il* ***entrait*** *à l'estaminet Bordelais, pour prendre le vermouth* (...) (Flaubert, *L'éducation sentimentale*, Folio:57)

(15b) *Félicité tous les jours s'y rendait.*

*A quatre heures précises, elle* ***passait*** *au bord des maisons,* ***montait*** *la côte,* ***ouvrait*** *la barrière, et* ***arrivait*** *devant la tombe de Virginie.* (Flaubert, *Un cœur simple*, G-F:64-65).

Deux types de séries sont à l'œuvre dans ces structures : une série d'entités identiques (série M : [BsM1,BsM2]), une série d'entités différentes (série D : [BsD1, BsD2]) :

fig.11

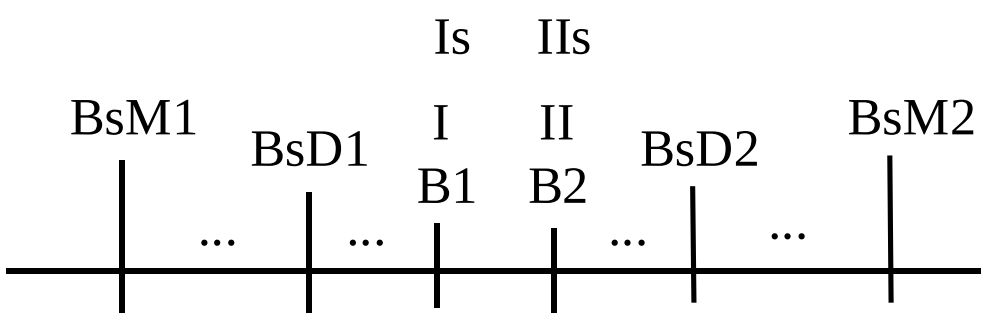

Dans cette structure, le procès est vu de façon aoristique (si bien qu'il peut être pris dans une succession), tandis que la série D (parce que le procès considéré n'en constitue qu'une partie, une étape) et la série M (itérative) sont présentées sous l'aspect inaccompli.

# 3. Le calcul de l'itération

## 3.1. Les sources de l'itération

L'itération peut être explicite ou implicite. Elle est explicite lorsqu'elle provient de marqueurs itératifs ; elle est implicite lorsqu'elle résulte de la résolution de conflits.

Parmi les marqueurs explicites d'itération, on retiendra :

**a)** les verbes intrinsèquement itératifs (ex. *sautiller*),

b) les déterminants du SN objet dans certains types de groupes verbaux (ex. : *manger deux fraises*),

c) certains circonstanciels de localisation temporelle (ex. *chaque mardi*),

d) les adverbes itératifs fréquentatifs (ex. *parfois*),

e) les adverbiaux itératifs répétitifs (ex. *trois fois*),

f) les adverbes présuppositionnels (ex. *encore*).

fig.12 **sources de l'itération**

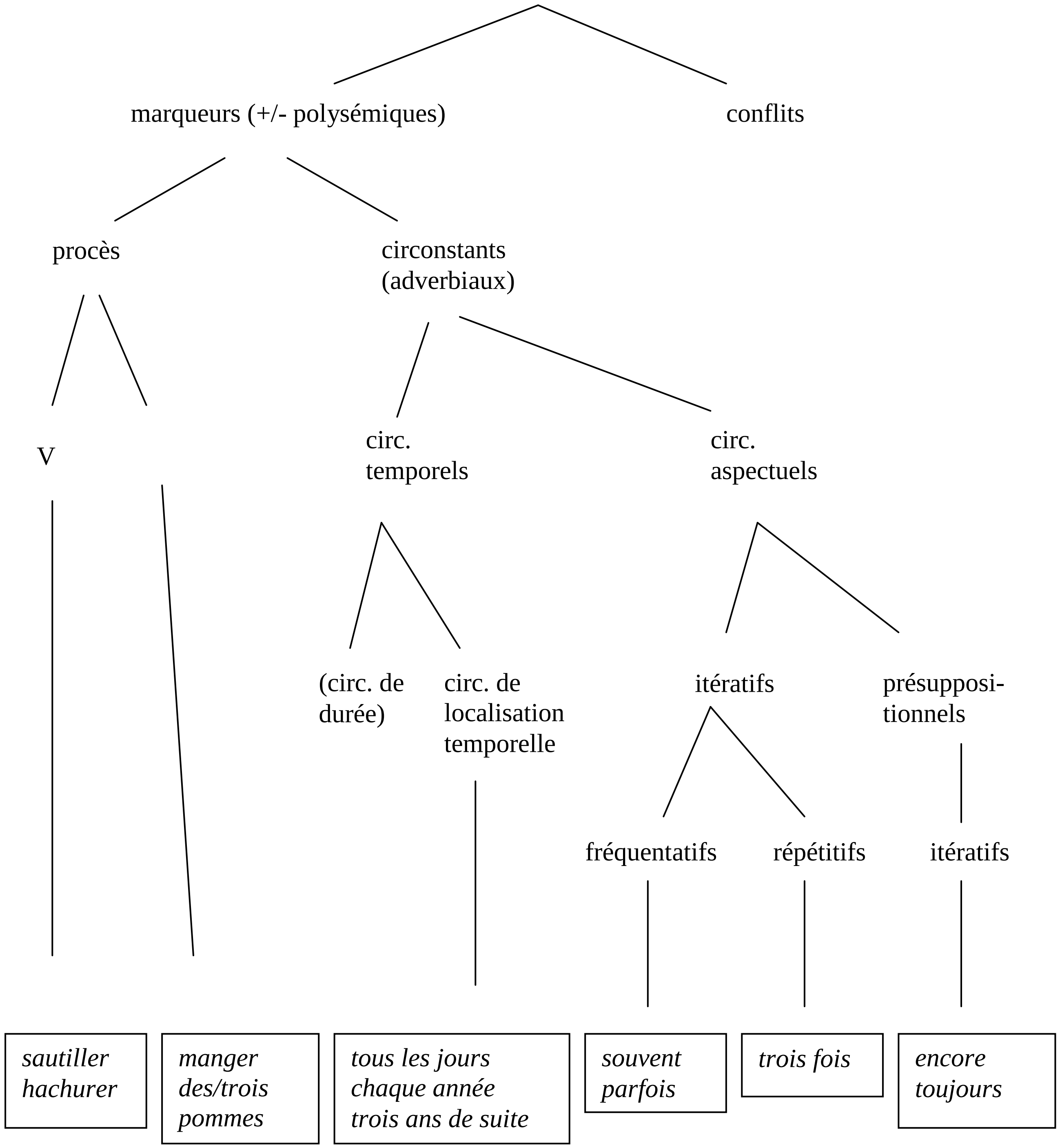

Encore faut-il préciser que comme ces éléments sont tous plus ou moins polysémiques, ils ne marquent l'itération que dans des contextes particuliers, qu'il convient de définir. Afin de ne pas trop alourdir l'exposé, nous n'évoquons ici que les itérations provenant de circonstants ou de conflits.

## 3.2. Les marqueurs explicites

### *3.2.1. Les circonstanciels de localisation temporelle*

La notion de circonstant, définie de façon essentiellement négative (comme subsumant tout complément qui ne désigne pas un actant) recouvre une réalité hétérogène. Dans le domaine aspectuo-temporel, on opposera les **circonstanciels de temps** qui construisent un intervalle circonstanciel ([ct1,ct2]) sur l'axe temporel, aux **adverbes d'aspect**, qui modifient les relations entre les intervalles construits à partir des autres marqueurs de l'énoncé (en particulier entre l'intervalle de référence et celui du procès).

Parmi les circonstanciels temporels, on distingue à nouveau :
a) les circonstanciels de **durée**, qui définissent la taille de l'intervalle circonstanciel sans le localiser autrement que par rapport au procès et/ou à l'intervalle de référence;
b) les circonstanciels de **localisation**, qui situent l'intervalle circonstanciel de façon plus ou moins précise et plus ou moins déterminée par rapport au calendrier (localisation absolue), à l'intervalle de l'énonciation (localisation déictique), ou à un autre procès (localisation relative); ce dernier type de localisation est caractéristique des subordonnées circonstancielles, dans lesquelles l'intervalle circonstanciel est situé par rapport au procès exprimé par la subordonnée.

Les circonstanciels de localisation temporelle peuvent être intrinsèquement itératifs (ex. *: tous les dimanches*, *chaque mercredi*). Il en va de même des subordonnées temporelles, qui opèrent, elle aussi, un forme de localisation temporelle (*chaque fois que*, *toutes les fois que,* etc.). Ils créent directement une série d'intervalles circonstanciels, et indirectement une série de procès (série itérative).

### *3.2.2. Les circonstants aspectuels*

Les circonstants (ou « adverbiaux ») aspectuels se laissent diviser en **adverbes itératifs**, qui se répartissent à leur tour en adverbes de **fréquence** et de **répétition**, et en **adverbes présuppositionnels** (*déjà, encore*). On examine très rapidement le fonctionnement de ces adverbes.

#### *3.2.2.1. Les adverbes itératifs*

Les adverbes itératifs déclenchent, comme leur nom l'indique, l'itération du procès et donc la constitution d'une série itérative bornée de façon intrinsèque (si le nombre d'occurrences du procès est déterminé) ou extrinsèque (si le nombre d'occurrences du procès reste indéterminé). Les adverbes et locutions adverbiales de répétition (*trois fois, à cinq reprises*) indiquent le nombre d'occurrences du procès, et déclenchent la constitution d'une série intrinsèquement bornée ([Bsi1,Bsi2]) :

(16a) *En dix ans, il a gagné la coupe quatre fois*

(16b) ?* *Pendant dix ans, il a gagné la coupe quatre fois.*

Les adverbes de fréquence marquent une évaluation le plus souvent subjective et relative à une norme du nombre d'occurrences de procès . Ils se distribuent sur un continuum :

fig. 13

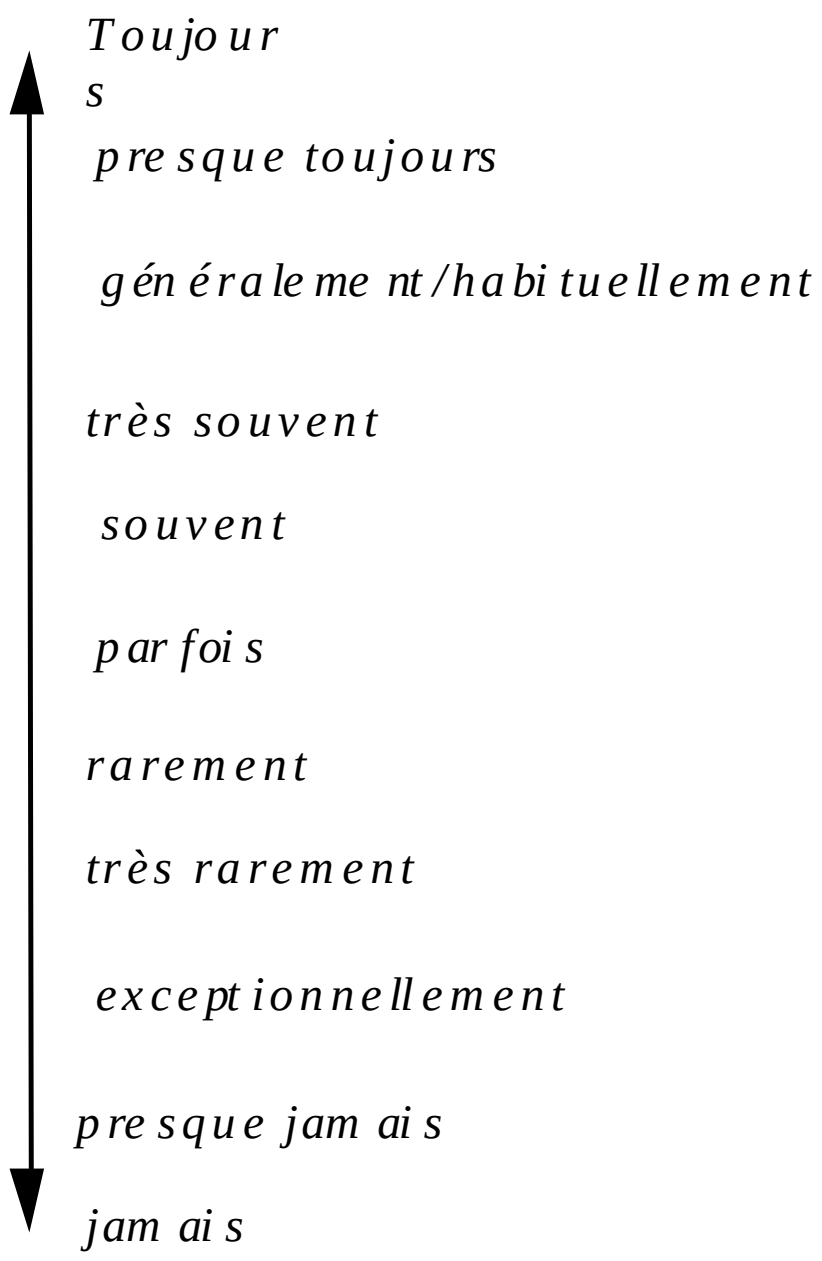

et servent à construire une série de procès bornée de façon extrinsèque.

*3.2.2.2. Les adverbes présuppositionnels*

Les adverbes présuppositionnels *déjà* et *encore* donnent des informations sur ce qui n'est pas directement montré du déroulement du procès. *Encore* indique que le procès avait déjà lieu avant la vue qui en est donnée (présupposition) et qu'il va vraisemblablement cesser ensuite (implication); *déjà* marque qu'il y a eu au moins un moment où le procès n'avait pas lieu, avant le début de la vue (présupposition), et qu'il va vraisemblablement avoir encore lieu après la vue (implication). On peut représenter ce fonctionnement, de façon très schématique, par les structures :

fig. 14

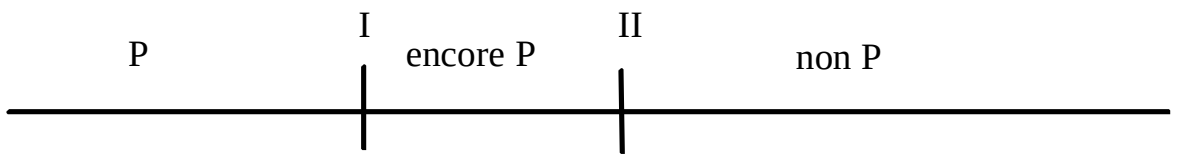

fig. 15

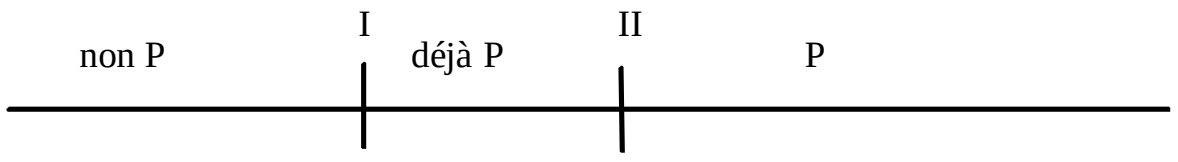

Or on sait que ces adverbes peuvent prendre, dans le domaine aspectuel, deux valeurs bien distinctes, que l'on a pu qualifier d'itérative et de durative, et qui sont identifiables au moyen de paraphrases; ainsi, *encore* duratif équivaut à [*continuer de* Vinf] ou à [*être toujours* Adj], itératif, il est substituable par *à nouveau*, *une fois de plus*. L'effet de sens itératif s'impose quand l'aspect est aoristique et/ou quand le procès est intrinsèquement borné; tandis qu'un procès borné de façon extrinsèque et présenté sous un aspect inaccompli autorise les deux interprétations :

(17a) *Il dormait encore* (itératif ou duratif)

(17b) *Il a encore dormi* (itératif à cause de l'aspect aoristique)

(17c) *Il mange encore du pain* (itératif ou duratif)

(17d) *Il mange encore une pomme* (itératif à cause de la valeur d'accomplissement du procès)

(18a) *Il dort déjà* (itératif : on s'attend à ce qu'il dorme à nouveau; ou duratif : on s'attend à ce qu'il continue de dormir)

(18b) *Il a déjà dormi* (itératif)

(18c) *Il mange déjà du pain* (itératif ou duratif)

(18d) *Il mange déjà une pomme* (itératif).

## 3.3. Les conflits

### *3.3.1. Le processus d'interprétation et la résolution des conflit*

Le processus d'interprétation consiste à mettre en commun des contraintes linguistiques (instructions et principes de bonne formation) et des contraintes pragmatico-référentielles (arrière-plan encyclopédique et conversationnel).

Conformément aux principes de la psychologie cognitive, on admet que le processus d'interprétation vise à obtenir la **cohérence** (qui suppose la compatibilité entre contraintes linguistiques) et la **plausibilité** (qui implique la compatibilité entre la représentation linguistique et l'arrière-plan pragmatico-référentiel).

Un **conflit** résulte d'une incompatibilité entre contraintes (cf. § 2.1. ci-dessus). Il existe des modes de résolution de conflit. Une **résolution de conflit** consiste en une déformation minimale des représentations telle que toutes les contraintes soient satisfaites.

Le principe général qui guide ces résolutions de conflit se laisse formuler comme suit :

> Déformer le procès – le moins possible – de façon à le rendre compatible avec les instructions marquées par les éléments du contexte linguistique.

Cette déformation peut prendre deux aspects :
a) la dilatation, avec ou sans itération;
b) le glissement de la figure du procès sur l'axe temporel, vers la portion la plus facilement accessible (i.e. la plus proche et/ou la plus saillante) qui satisfait aux exigences du contexte.

**Exemple** : l'itération comme résolution de conflit

(19) *Pierre nageait pendant deux heures depuis très longtemps*

l'énoncé (19) présente un conflit entre l'imparfait qui marque fondamentalement l'aspect inaccompli, [*depuis* + durée] qui n'est compatible qu'avec les aspects inaccompli et accompli, et [*pendant* + durée] qui impose l'aspect aoristique :

imparfait : B1 < I, II < B2 (aspect inaccompli)
[*pendant* + durée] : ct1 = B1 = I, ct2 = B2 = II
[*depuis* + durée] : ct1' = B1 ou B2, ct2' = II
Il y a donc conflit entre instructions : B1 = I et B1 < I

Ce conflit est résolu par l'itération. L'itération consiste à créer une série de procès, notée [Bs1,Bs2]. De sorte que chacune des occurrences de procès peut être vue de façon aoristique (sa durée est mesurée au moyen du circonstanciel [*pendant* + durée]), tandis que la série itérative dans son ensemble est présentée sous l'aspect inaccompli (marqué par l'imparfait). [*depuis* + durée] porte donc sur la série globale, et marque le décalage entre Bs1 et Is (voir la fig.10 du § 2.1. ci-dessus).

Remarquons que le même mécanisme de construction d'itération peut servir à résoudre un conflit entre instructions (linguistiques) et contraintes pragmatico-référentielles, comme dans l'exemple :

(20) Pierre a nagé pendant dix ans

(car il n'est évidemment pas référentiellement plausible que le procès se soit déroulé sans discontinuer pendant cette durée).

Pour que ce mode de résolution soit déclenché, il faut que le procès soit réitérable et que cela ne crée pas d'autre conflit.

### *3.3.2. Les classement des conflits*

Les types de conflits se laissent classer sur à partir de 1) **origine du conflit** (la nature des contraintes incompatibles), et 2) la **nature du conflit** (le type de relation crucialement mise en cause) :

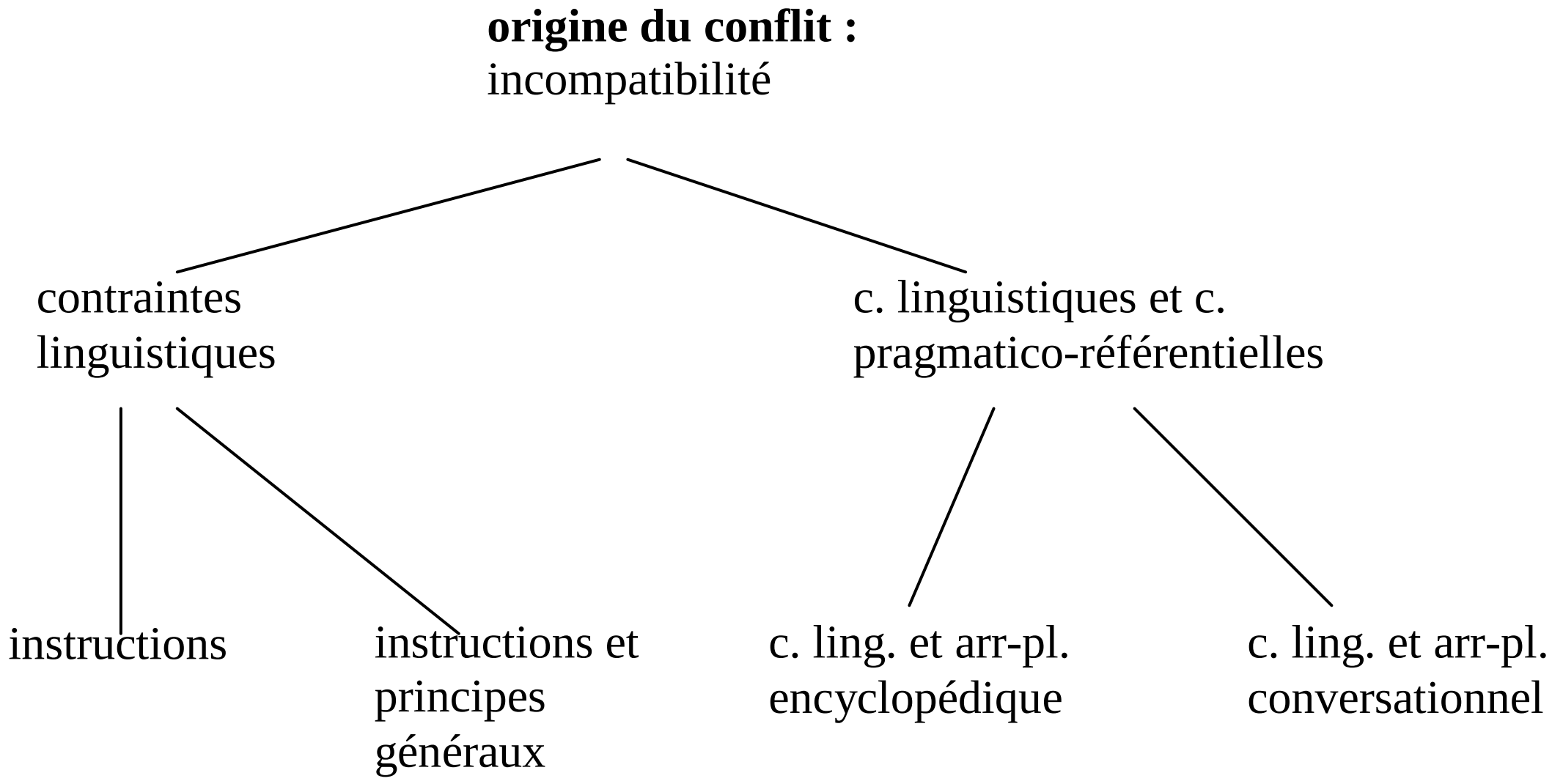

fig.16

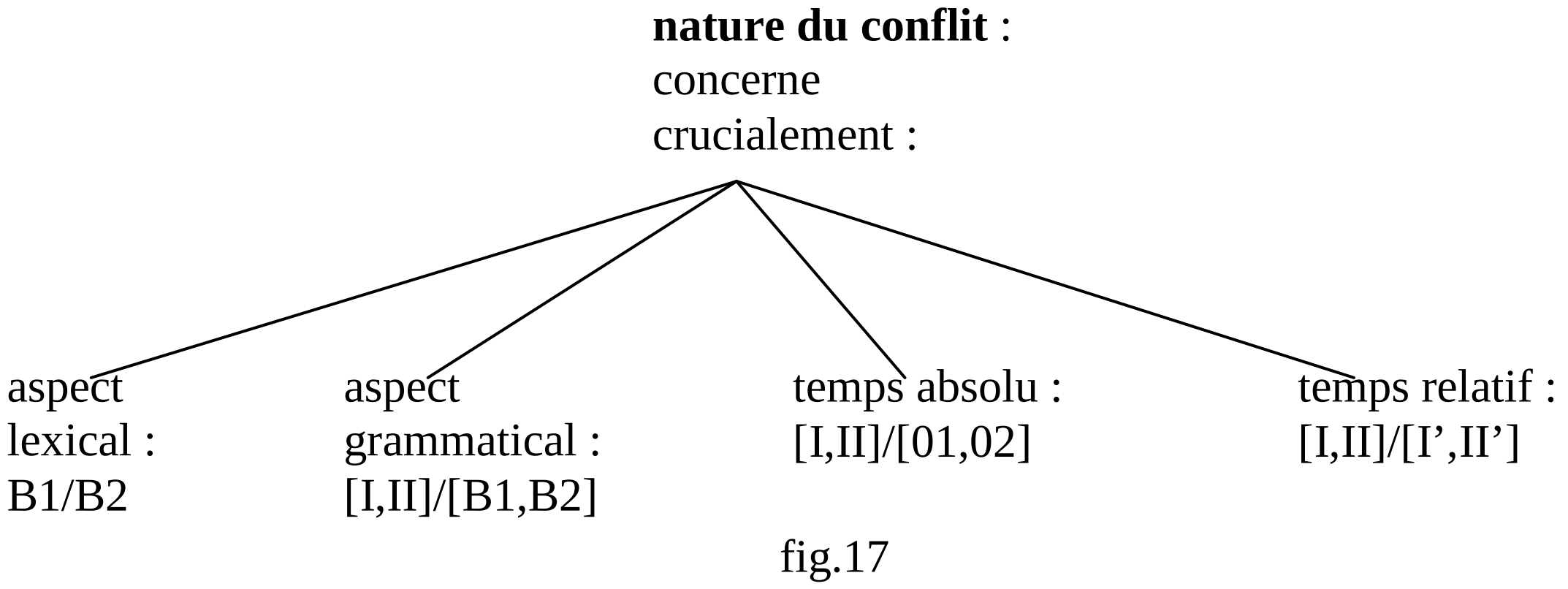

fig.17

On obtient ainsi le classement suivant, des **conflits résolus par l'itération** :

| | **instructions** | **instructions vs principes généraux** | **c. ling. vs arr-pl. encyclopédique** | **c. ling. vs arr-pl. conversationnel** |
|---|---|---|---|---|
| **asp. lexical** | proc. ponctuel + circ. de durée<br>*tousser pdt/depuis 5 min.*<br>*longtemps se coucher de b. heure* | | *faire du vélo pdt 10 ans* | |
| **asp. grammatical** | proc. ponctuel + asp. inaccompli<br>*Paul tousse/toussait* | Asp. inaccompli + circ de durée/date<br>*Il joue/jouait le lundi/dix minutes* | | |
| **temps absolu** | | | | prst + situation<br>*Je joue du piano* |
| **temps relatif** | | asp. inaccompli + marque de succession :<br>itération de séries de p. successifs<br>*Il marchait, puis il mangeait* | Asp. inacc + incompatibilité des procès<br>*Elle déclara qu'elle dormait sans somnifère*<br>itération de séries de p. successifs<br>*il mangeait, il dormait* | |

# Un modèle « objet[16] » de l'itération

Yann Mathet

## Sommaire

## 1. Introduction

La modélisation de la temporalité se fait souvent au moyen d'intervalles disposés sur un axe temporel, et liés entre eux par un certain nombre de relations. L'utilisation de différents types d'intervalles (énonciation, procès, référence, circonstanciel) et de relations (recouvrement, antériorité, etc.) permet de rendre compte avec finesse, comme il a été montré au chapitre précédent, de notions telles que le temps absolu, le temps relatif, ou encore l'aspect.
Le travail sur l'itération nous montre cependant la nécessité de disposer d'un modèle plus riche, celui-ci se prétant peu à la manipulation d'entités itérées. En effet, même dans une itération simple comme
(1) *elle est allée sept fois à la montagne,*
il y a bien sûr, en filigrane, 7 procès, les **itérés**, à créer et à relier entre eux (ou du moins leurs intervalles associés), ce qui n'est déjà pas aisé d'un point de vue « représentation schématique » ; mais il y a aussi, bien sûr, à coder le fait que ces différents procès résultent du même précidat « elle aller à la montagne », l'**itérant**. Le lien entre les différents itérés et leur existence même n'ont de sens qu'en présence de l'itérant dont ils résultent. Par ailleurs, la façon dont les itérés sont produits est en soi un processus varié, et dont il faut rendre compte. Dans l'exemple précédent, il faut coder le fait que 7 itérés sont produits, alors que l'information sera d'ordre fréquentiel si l'on remplace s*ept fois* par *souvent*.
Pour ces raisons, nous proposons dans cette partie un modèle « objets » de la temporalité qui permet de considérer l'itération à part entière. Ce modèle doit par ailleurs intégrer les différents intervalles et relations du modèle SdT de  Laurent Gosselin, afin de continuer à rendre compte de phénomènes tels que l'aspect.

---

[16]Nous utilisons dans ce chapitre les diagrammes de classes UML pour lesquels il est simplement nécessaire d'être capable de distinguer la notion d'héritage, représentée au moyen d'une flêche allant de la classe fille à la classe mère, de la notion d'association, réprésentée au moyen d'un trait simple (association pauvre) ou se finissant soit par un losange vide (agrégation), soit par un losange plein (composition).
L'héritage signifie que la classe fille « est un sous type » de la classe mère (un peu à la façon d'une relation d'hyponymie), comme par exemple une classe Bateau peut être une sous classe d'une classe Véhicule, tandis que l'association signifie simplement qu'une classe utilise une autre classe en tant que composant, comme la classe Bateau pourrait être en association avec la classe Moteur en tant qu'un objet bateau dispose, entre autres, d'un moteur.

# 2. Analyse et représentation des itérations : itérant et itérés

## 2.1. Exemple introductif

Le processus d'itération a ceci de particulier qu'il génère plusieurs procès (les itérés) à partir d'un seul (l'itérant). Dans l'exemple (1), il résulte du syntagme “sept fois” la création de sept procès distincts et disjoints qui s'étalent dans le temps. Mais ces différents procès, les **itérés**, ont un statut particulier, un air de famille, puisqu'ils résultent de la copie (du clonage) d'un même procès, *aller à la montagne*.
Nous pouvons schématiser ce processus, dans le cas de notre exemple, de façon simple : dans une itération, nous sommes en présence d'un itérant (*aller à la montagne*), qui donne lieu à la création d'itérés (sept procès étalés dans le temps) par le biais d'un itérateur (*sept fois*), selon le schéma suivant.

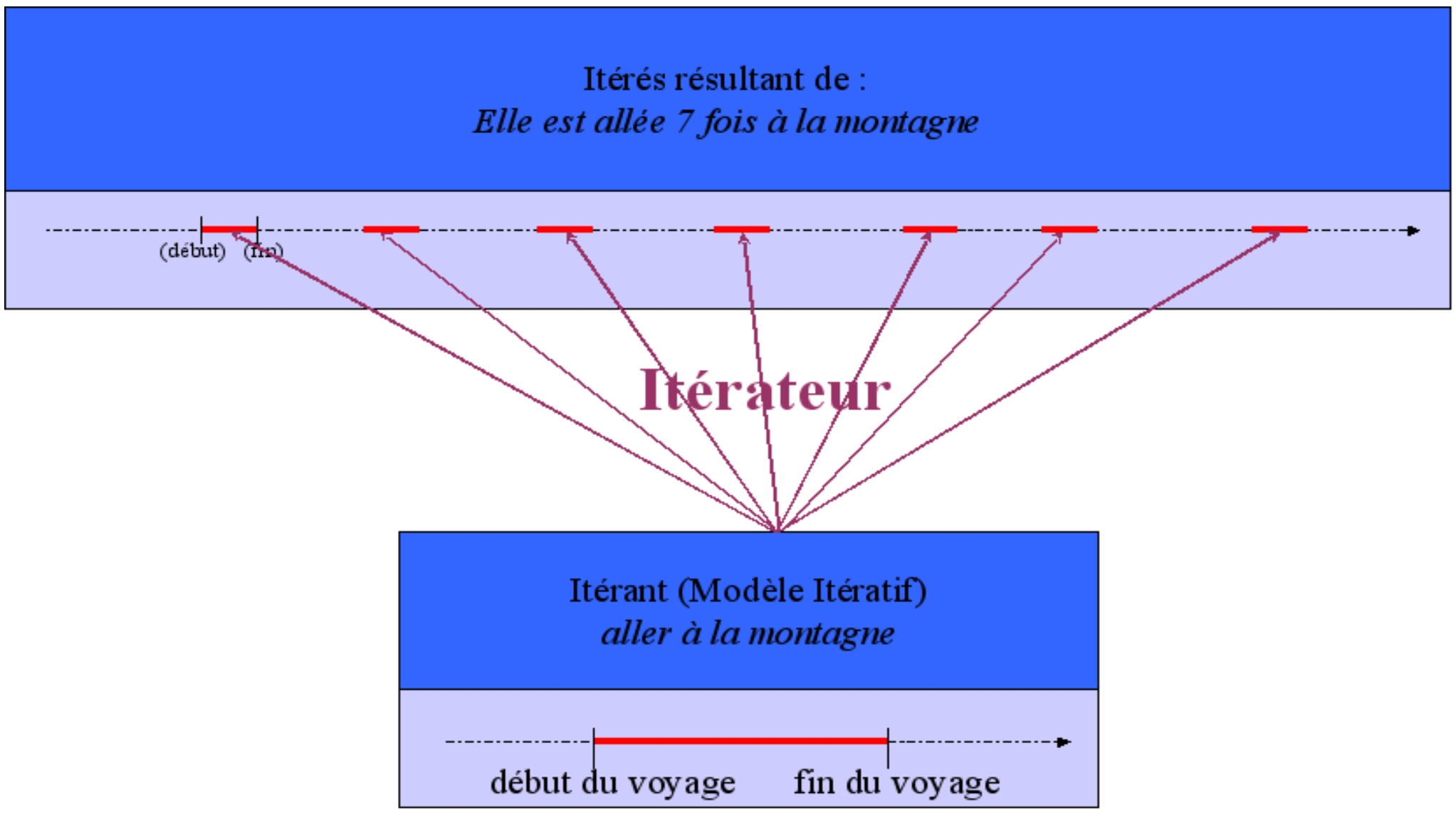

## 2.2 Temporalité conventionnelle, temporalité modèle

L'itérant et les itérés, en tant que procès, viennent naturellement se placer sur un axe temporel. Cependant, alors que les itérés viennent se placer sur l'axe temporel conventionnel (on peut très bien asserter “la deuxième fois qu'elle est allée à la montagne, c'était pour Noël 2001”, ce qui montre bien l'ancrage des itérés dans la temporalité conventionnelle), l'itérant vient se placer quant à lui sur une temporalité virtuelle, indépendante des procès classiques.
Nous appellerons **temporalité conventionnelle** l'axe dans lequel viennent s'inscrire les intervalles (de procès, de référence, etc.) issus de l'interprétation des formes conjuguées, itératives (plusieurs procès) ou non (un seul procès).
Nous appellerons **temporalité modèle** un axe temporel dans lequel viennent s'incrire des intervalles servant de modèle à d'autres intervalles, notamment dans le cas des itérations (nous n'avons travaillé que sur l'itération, mais cette temporalité modèle pourrait peut-être tenir un rôle comparable dans le traitement des modalités).
Dans un but de simplification, lors de l'interprétation d'un texte “informatif” nous postulons l'existence d'une temporalité conventionnelle unique, et d'un nombre indéterminé de temporalités modèles.

Concernant le phénomène itératif, un itérant est ancré dans une temporalité modèle, tandis que ses itérés se placent dans la temporalité conventionnelle, chacun de ces derniers étant contruit par clonage et projection de l'itérant. Plusieurs itérants d'une même itération se placent dans la même temporalité modèle, tandis que deux itérants de deux itérations distinctes appartiennent à des temporalités modèles distinctes.
La temporalité modèle possède les propriétés de la temporalité conventionnelle - disposant notamment d'un axe ordonné permettant différents types de relations entre intervalles - à ceci près qu'elle ne permet pas d'ancrage temporel absolu. En effet, le propre de cette temporalité est de proposer des modèles à dupliquer et à projeter dans la temporalité conventionnelle. Cette opération de duplication/projection serait en contradiction avec un ancrage absolu des itérants.
(2a) Chaque année, elle va à la montagne.
(2b) Chaque année, elle va à la montagne le 25 décembre.
(2c) ? Chaque année, elle va à la montagne le 25 décembre 2005.
Nous constatons dans l'exemple (2) que *aller à la montagne*, et *aller à la montagne le 25 décembre* sont de bons candidats à la temporalité modèle en vue d'une itération, tandis que *aller à la montagne le 25 décembre 2005*, en raison de son ancrage temporel absolu (et donc unique), n'est pas un modèle reproductible possible.

## 2.3. Lien entre itérant et itérés

Nous avons indiqué que les itérés sont construits par une opération de clonage/projection de l'itérant. Il est toutefois nécessaire de préciser la nature de cette opération, qui du fait de son essence linguistique déroge un peu à la rigueur mathématique.
(3) Chaque jeudi, de 20 heures à 22 heures, ils faisaient une partie de boston.
Considérons à présent l'itération issue de l'exemple (3). On peut considérer en première approche que l'itérateur, vu comme une fonction mathématique simple, duplique l'itérant de la temporalité modèle un certain nombre de fois vers la temporalité conventionnelle. Cet itérateur est régulier, de période une semaine.

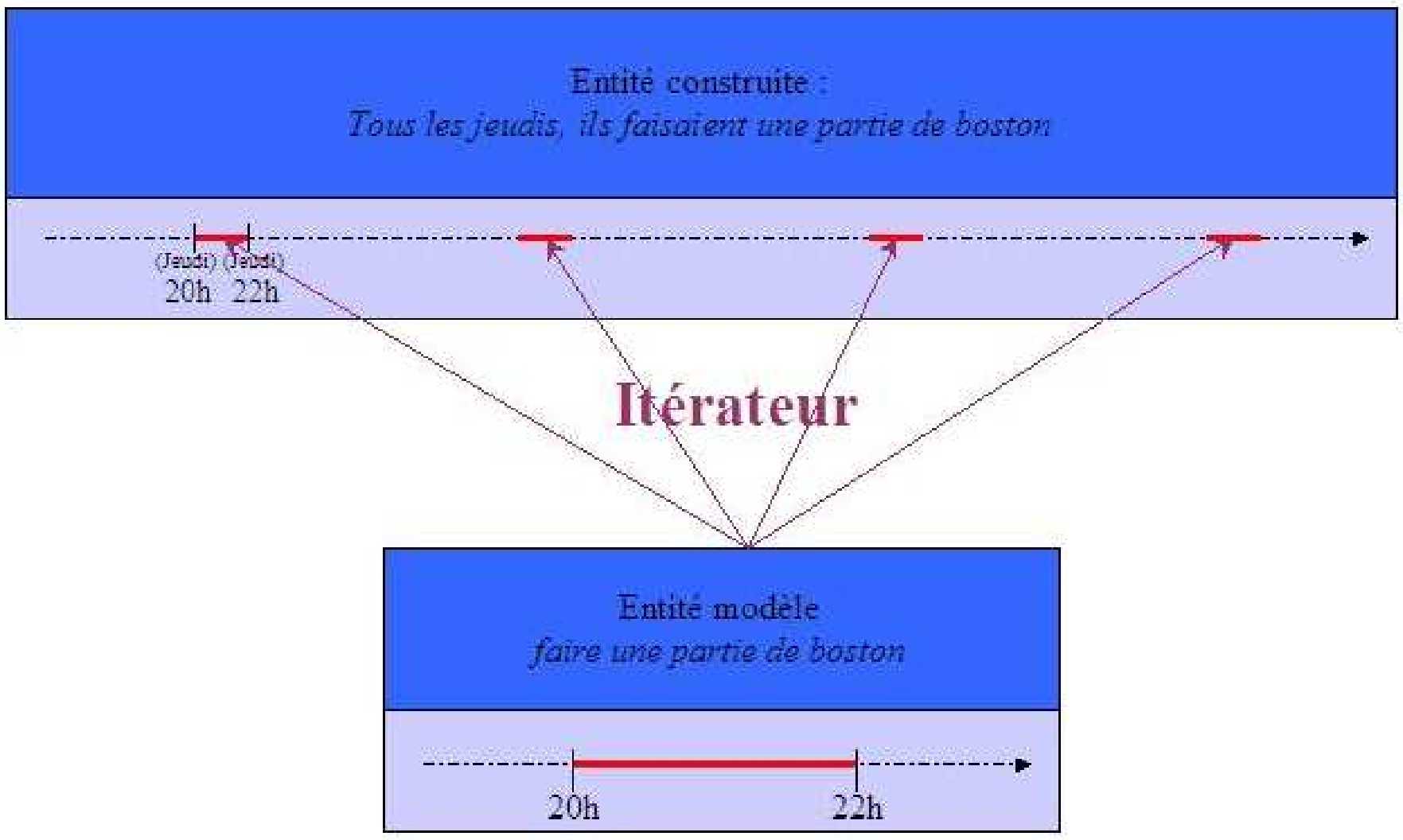

Si elle a le mérite d'être simple, cette construction sort assez franchement de la réalité linguistique dont elle veut rendre compte. En particulier, sa régularité mathématique rigoureuse, faisant démarrer toutes les parties à 20 heures et les faisant s'achever toutes à 22 heures, tend à surspécifier les itérés : en l'occurrence, le locuteur ne s'engage pas à ce que les parties soient toutes calibrées de façon aussi précise. Plus encore, l'entité itérante se décline en différents itérés qui peuvent différer non seulement quant à l'ancrage temporel, mais aussi quant à leur contenu :

(4) Parfois, ils invitaient le voisin ; la partie était alors plus longue

Dans l'exemple (4), le contenu de ce qui est porté par l'itérant de (3) varie *parfois*, intégrant un joueur de plus. Les itérés qui en résultent font pourtant partie intégrante de l'itération, *parfois* faisant référence à un sous ensemble des itérés. De plus, dans le même temps, ces itérés dérogent quelque peu à leur modèle aussi en ce qui concerne la durée. Dans cet exemple, on pourrait parler d'un itérant à (au moins) deux facettes linguistiquement distinctes (l'une modélisant les parties sans le voisin, l'autre avec ce dernier), distribuées avec une prédominance de la première (la deuxième, introduite par *parfois*, devant être plus rare).

Il est linguistiquement possible générer autant de dérogations à l'itérant que l'on veut :

(5) le premier jeudi de mars fut particulier : la partie débuta à 19h30.

Et l'on peut même envisager des dérogations au modèle initial plus profondes, comme dans l'exemple (6), où les parties de boston se voient parfois remplacées par un jeu d'une autre nature, mais le tout se fondant finalement dans un itérant hyperonymique « parties de cartes ».

(6) Une partie de poker venait parfois se substituer à l'habituel boston.

Ces différents exemples nous amènent à considérer l'opération de projection comme plus lâche qu'une projection mathématique classique, en ce sens qu'elle utilise l'itérant comme **modèle** d'entité à projeter, lequel modèle peut se voir décliner en différent projetés, les itérés, s'inspirant du modèle, certes, mais chacun ayant potentiellement ses particularités.

De ce fait, un itéré possède deux facettes. Sa facette « entité résultant d'un itérant », et est en ce sens une copie d'un modèle, semblable aux autres itérés (cette facette est en quelque sorte la part de temporalité modèle de l'itéré). Mais à l'opposé, sa facette « entité singulière », qui s'inscrit dans la temporalité conventionnelle, rend compte des spécificités propres de cet itéré (comme le fait que telle partie de boston à débuté à 20h18 et s'est achevée à 22h23).

En l'absence de toute précision linguistiquement formulée, un itéré fraîchement construit se voit attribuer des valeurs par défaut issues de l'itérant. Dans notre exemple, il s'agit du fait que le procès correspond à une partie de boston, que celle-ci débute à 20 heures, et qu'elle s'achève à 22 heures. Un itéré nait donc sous sa seule facette « entité résultant d'un itérant », et n'a de singulier que son ancrage temporel. Dès lors, des ajouts singuliers peuvent être faits, comme dans l'exemple (5), et comme illustré dans la figure ci-dessous dans le 9ème itéré.

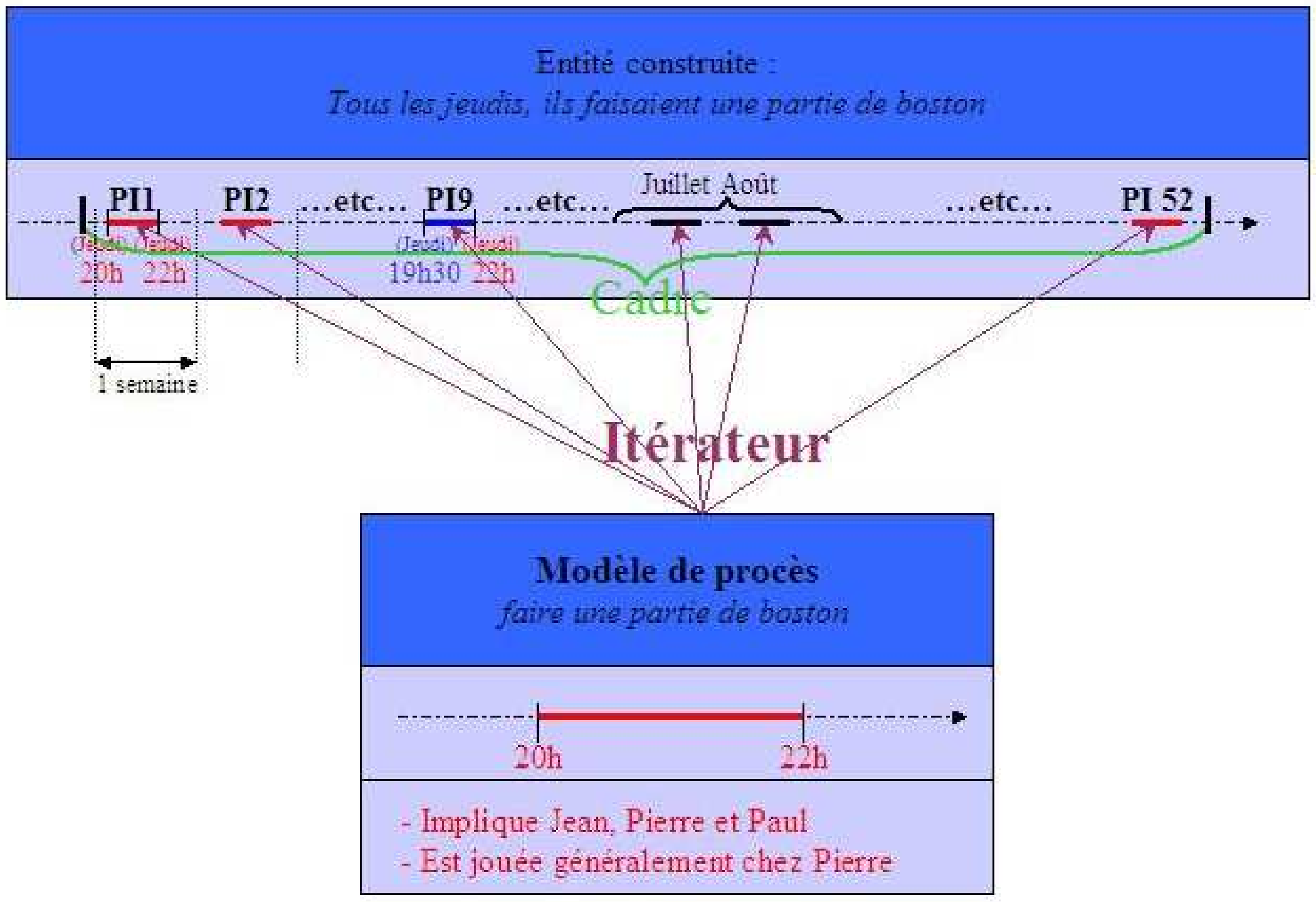

D’un point de vue formel, nous utiliserons le terme **modèle itératif** pour désigner la classe d’objets rendant compte des **itérants**, la notion de « modèle » rendant bien compte du lien qui existe entre itérant et itérés, et notamment du fait que l’itérant se comporte comme un modèle pouvant être décliné en des images itérées toutes un peu différentes.

### Les intervalles de la SdT dans le présent modèle

Nous pouvons à présent faire un premier lien entre le modèle de l’itération de la SdT présenté par Laurent Gosselin dans le chapitre précédent (partie 2, p. 9) et le présent modèle objet.
Les intervalles liés à une « série de procès itérés », respectivement nommés [Bs1, Bs2] pour ce qui est de l’ensemble des itérés, et [Is, IIs] pour ce qui est de l’intervalle de référence de la série, se placent dans la temporalité conventionnelle du présent modèle.
Les intervalles liés au modèle d’occurrence, respectivement nommés [B1, B2] pour ce qui est du procès, et [I, II] pour ce qui est de l’intervalle de référence, se placent quant à eux dans la temporalité modèle du présent modèle.
Ces intervalles qui étaient placés sur un même axe dans le chapitre précédent verront donc leurs équivalents respectifs prendre place sur des axes différents : un axe unique relatif à la temporalité conventionnelle, et un à plusieurs axes modèles relatifs à la temporalité modèle (généralement un axe par itération distincte).

## Présentation du modèle

Les deux types de temporalité ayant été présentés, et nos objectifs de modélisation ayant été posés, nous présentons à présent le modèle objets de l’itération.

### 3.1. Itération

Classe principale de ce modèle, c’est elle qui permet de représenter une itération, comme celle résultant de l'exemple (1).
Une itération est composée d'un **Itérateur** et d'un **modèle itératif**.

L'**itérateur** est un objet permettant d'itérer des événements dans le temps. Cela peut être en proposant un intervalle non convexe comme celui résultant de « tous les jeudis » (au sein duquel on place un itéré par composant convexe), mais aussi par d'autres moyens que par des intervalles, comme dans « 3 fois » (numéraire), « souvent » (fréquentiel), « à chaque fois qu'il vient » (événementiel). Concernant ce point particulier, le présent modèle déroge un peu à ce qui est présenté dans les parties [REF LAURENT] et [REF PATRICE&GERARD], sans s'y opposer, mais cette singularité nécessite quelques éléments de discussion que nous avons reportés en partie 7 de ce chapitre.
Le **modèle itératif** correspond à (un modèle de) ce qui est itéré (que nous avons désigné **itérant** précédemment). Il est constitué d'un ou de plusieurs procès modèles, qui peuvent être liés les uns aux autres par des relations temporelles ou d'un autre ordre (par exemple la causalité).

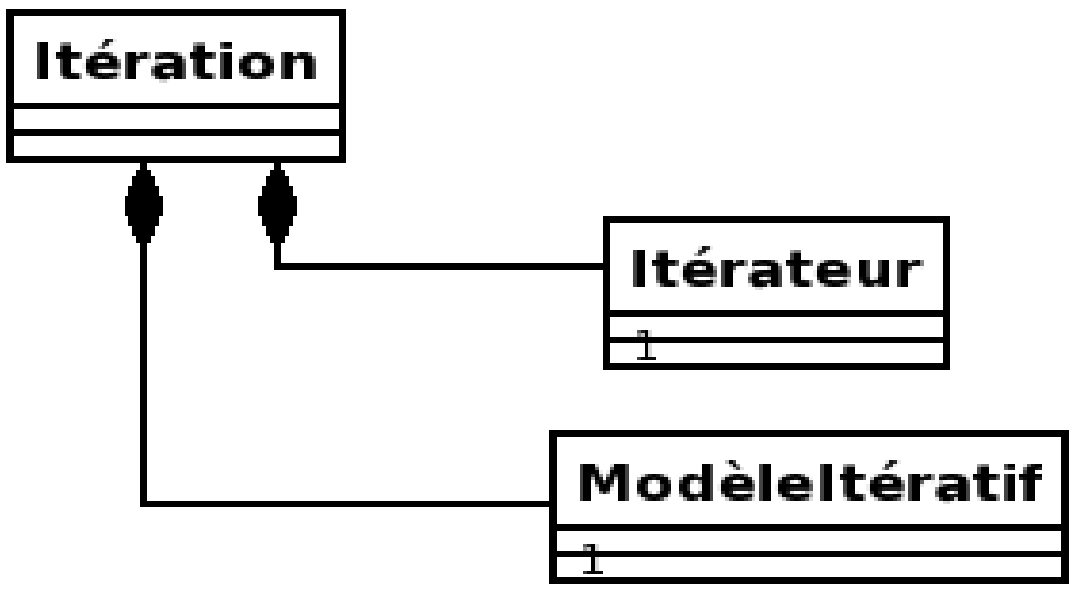

L'un et l'autre de ces objets peuvent correspondre à des éléments linguistiques distincts, mais pas forcément.
(7)Tous les jeudis, ils faisaient une partie de boston.
(8)Il va souvent à la montagne / dans le jardin.
(9)Chaque fois qu'ils viennent, nous faisons un billard.
En effet, dans l'exemple classique (7), l'itérateur semble directement provenir du complément circonstanciel « tous les jeudis », et le modèle de «faisaient une partie de boston», mais :
- dans (8), l'itérateur est construit à partir de l'adverbe *souvent* et de l'interprétation complète du procès, en tenant compte d'éléments pragmatico-référentiels, puisque *montagne* donnera une fréquence de par exemple 3 ou 4 fois par an, alors que *jardin* une fréquence de plusieurs fois par jour.
- dans (9), le procès «ils viennent» sert à créer l'itérateur, mais fait aussi partie du modèle itératif. C'est plus flagrant sans doute dans *Chaque fois qu'ils viennent, ils prennent la voiture,* où le procès « ils prendre la voiture » s'inscrit dans le procès « ils venir ».

## 3.2. Itérateur

### *3.2.1 Présentation*

Le rôle d'un itérateur est de proposer une suite (souvent ordonnée) de "positions" (connues ou non) de l'axe temporel à chacune desquelles on pourra faire correspondre temporellement une instanciation du modèle d'itération. D'un point de vue formel, il donne entre 1 et n (voire une infinité) "positions" (et d'un point de vue objet, une méthode getNextPosition()).
Ces "positions" ne sont pas obligatoirement renseignées temporellement. Par exemple, dans (10), tout au plus peut-on dire qu'elles sont circonscrites au cadre proposé par *sa vie.*
(10)De toute sa vie, il est allé trois fois à mer.
(11)Par trois fois, tu me renieras.
Par ailleurs, un itérateur possède un cadre, intervalle recouvrant l'intégralité de l'itération. Ce cadre est souvent anaphorique, comme en (11), mais peut être présent dans la phrase, comme dans (10).

Linguistiquement, ce cadre correspond à l'intervalle **[Bs1, Bs2]** du modèle SdT de Laurent Gosselin.
Enfin, conformément au modèle SdT, un itérateur possède un booléen (attribut ne pouvant prendre que deux valeurs, vrai ou faux) indiquant si l'itération résultante est intrinsèquement bornée ou non. Ce booléen sera notamment vrai dans le cas d'une itération induite par « trois fois », et faux dans le cas d'une itération induite par « quelque fois ».

### 3.2.2. Modélisation

Les différentes classes d'itérateurs sont les suivantes :

1) Intervalle non convexe (tous les jeudis, lundi et mercredi derniers)
2) Numéraire (3 fois, quelques fois, manger 3 pommes)
3) Evénementiel (quand ils viennent, dès qu'ils arrivent, lorsqu'il fait beau : état sujet à changements vrai/faux)
4) Fréquentiel (souvent, fréquemment, de temps en temps)
5) Et peut être un amalgame de plusieurs types d'itérateurs (3 jeudis et une autre fois), classe nommée Itérateur composé.

Une distinction principale est à faire entre la classe Intervalle non convexe et les autres. En effet, avec cette classe, la temporalité est déjà inscrite dans l'itérateur, puisque différents intervalles extérieurs à l'itération sont fournis d'emblée, tandis qu'il n'y a pas d'intervalle préexistant avec les autres classes.

On peut donc faire la séparation claire entre deux classes principales, ItérateurParIntervalles d'une part, et ItérateurDirect d'autre part.

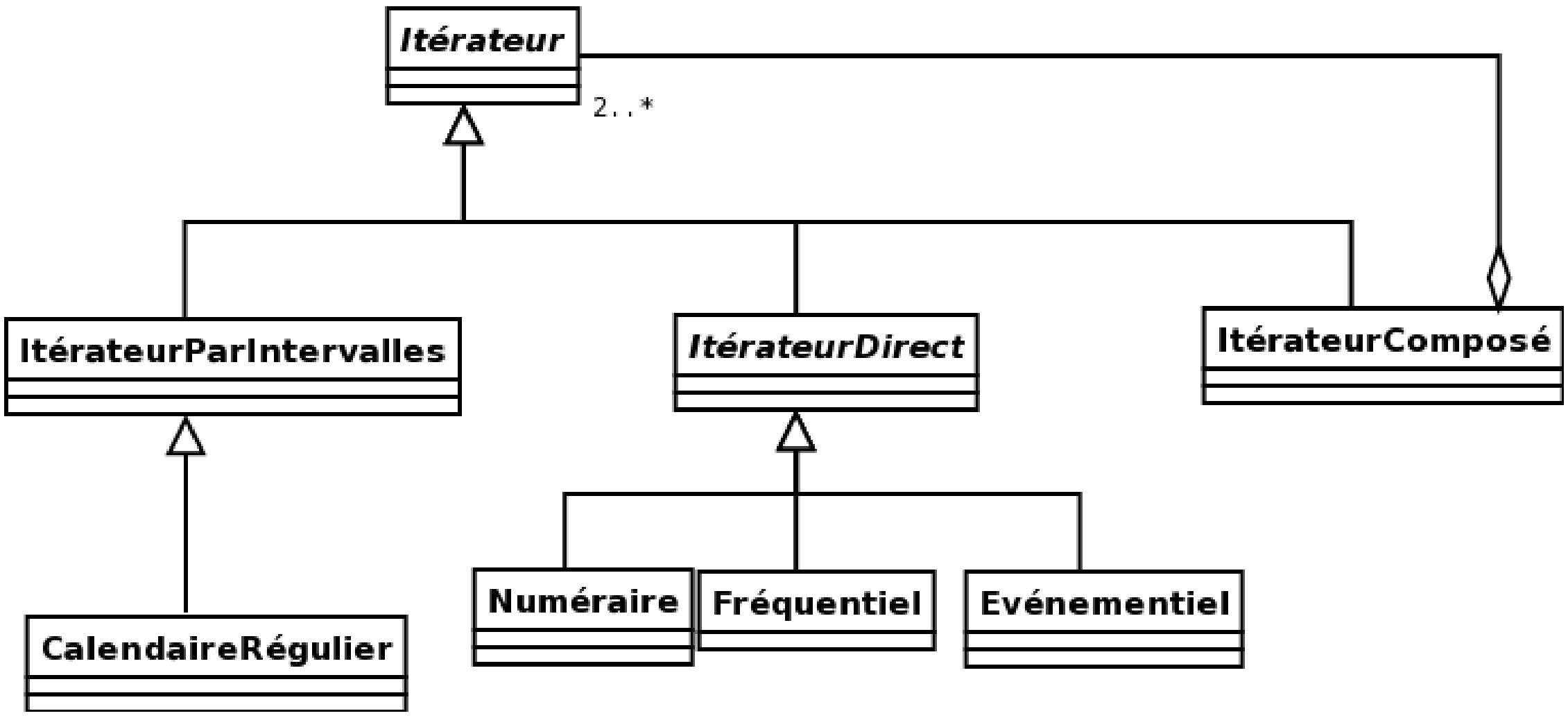

Une alternative possible consite à ne pas distinguer les ItérateursParIntervalles des ItérateursDirects, en considérant qu'un itérateur par intervalles délivre une suite de "positions" en nombre égal au nombre d'intervalles qu'il possède, et que chacune de ces "positions" s'assortit d'une contrainte d'appartenance à l'intervalle correspondant.
Ainsi, *les trois derniers jeudis* et *les trois dernières fois* seraient assez proches, chacun d'eux donnant trois "positions" successives, mais le premier assortissant ces trois positions de leur appartenance nécessaire au trois intervalles temporels correspondants (les 3 jeudis concernés).

Le diagramme est alors le suivant.

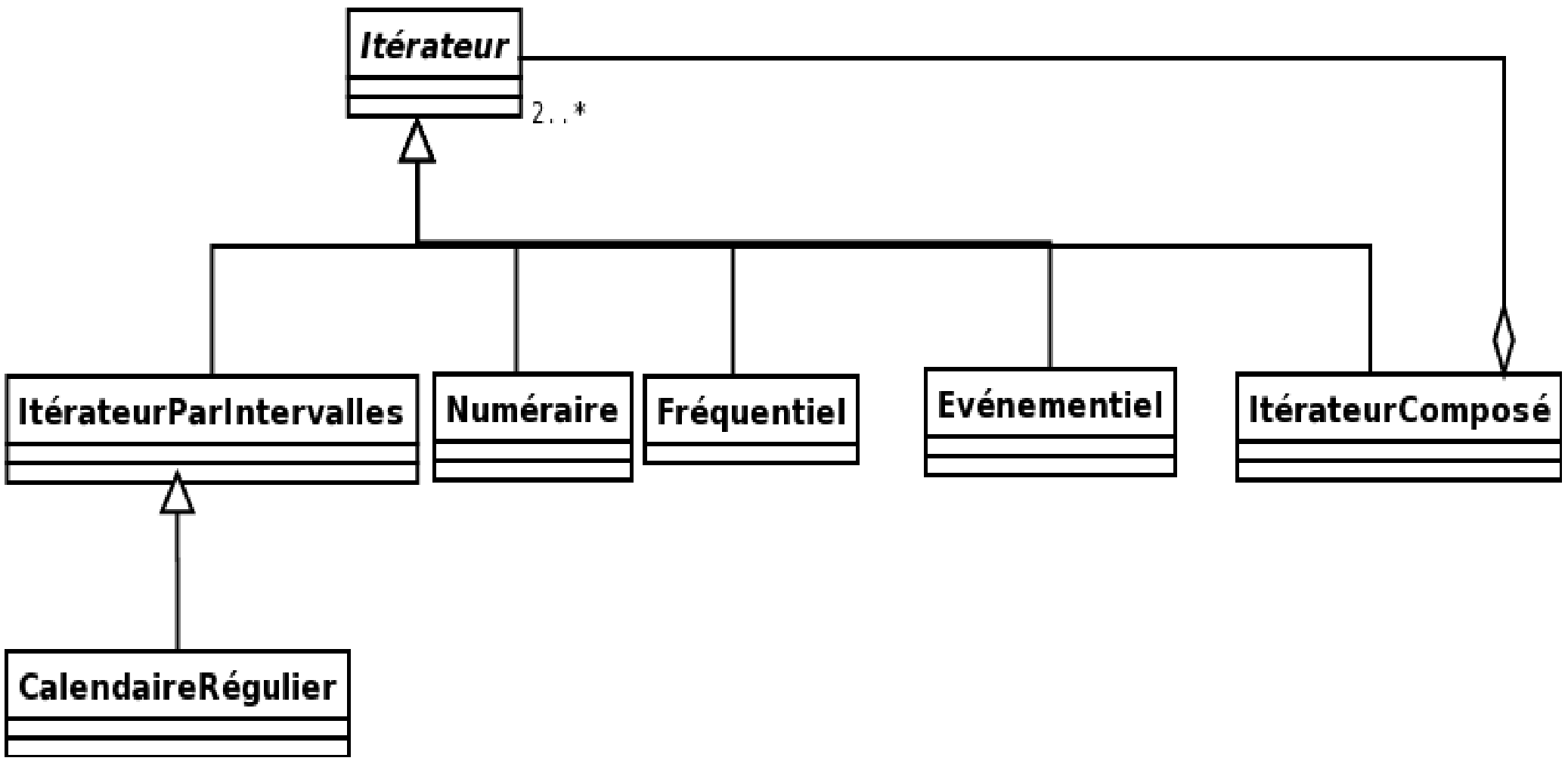

### *3.2.3. Lien avec les « sources de l'itération » de la SdT*

Même si ce chapitre s'intéresse essentiellement à la façon de formaliser les itérations, il est néanmoins intéressant d'essayer de coupler certaines classes à certaines données linguistiques. En reprenant les « sources de l'itération » proposées par Laurent Gosselin p.11 et fig.12 p.12 de la partie précédente, nous pouvons proposer les associations suivantes :

a) Verbes intrinsèquement itératifs : nous n'en rendons pas compte dans le présent modèle
b) Déterminants du SN objet tels que « manger deux pommes » : Numéraire
c) Certains circonstants de localisation temporelles tels que « chaque mardi » : CalendaireRégulier
d) Adverbes itératifs fréquentatifs comme « parfois » : Fréquentiel
e) Adverbes itératifs répétitifs comme « trois fois » : Numéraire
f) Adverbes présuppositionnels comme « encore » : item à rattacher à une itération déjà existante, et donc d'un type quelconque, ou, en l'absence d'itération préalable, création d'une itération de type Evénementiel.

## 3.3. Modèle Itératif

Le rôle d'un modèle itératif est de servir de modèle à ce qui va être itéré. Par exemple, dans (6), il s'agit du procès modèle *ils [faire] une partie de boston*. C'est le cas le plus simple, mais nous verrons de modèles itératifs plus riches, constitués de plusieurs procès modèles. Formellement, le modèle itératif est constitué de 1 à n procès modèles, et de relations entre ces différents procès modèles.

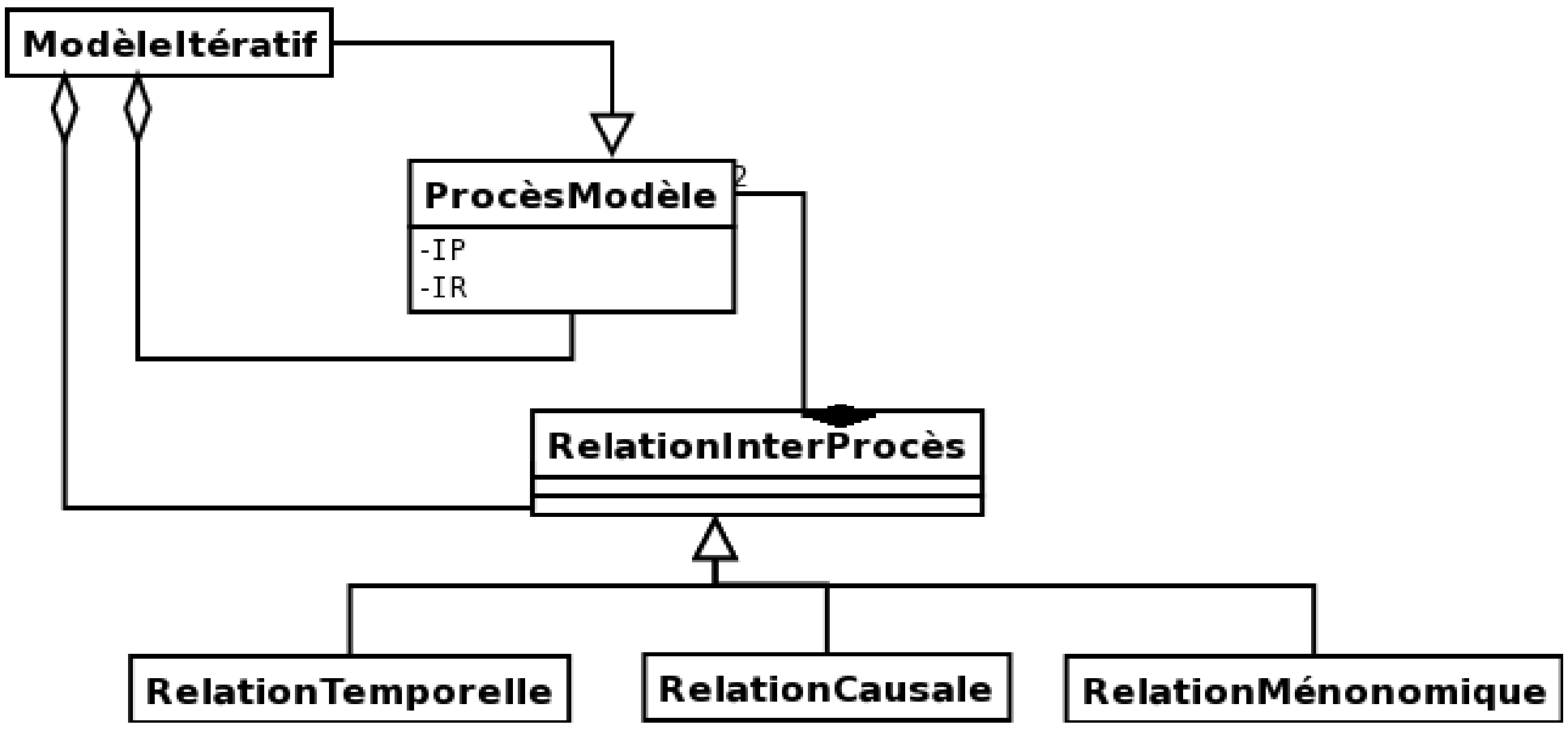

Ce schéma représente l'agrégation entre le modèle itératif et ses procès modèles d'une part, et entre le modèle itératif et les relations inter procès d'autre part. Par ailleurs, une relation inter procès met en jeu exactement deux procès modèles, ce que nous voyons dans la relation de composition. Enfin, une relation inter procès peut se décliner en relation temporelle, causale, méronomique ou autre, ce que nous voyons comme autant de relations d'héritage.

Le modèle Itératif hérite de la classe Procès Modèle, ce qui signifie qu'un modèle itératif peut lui-même servir d'ingrédient à un autre modèle itératif, de façon récursive.

### *3.3.1. Procès modèle*

Un procès modèle est un procès qui vient se placer sur un axe virtuel, que nous appelons temporalité modèle. La temporalité modèle a les même propriétés que la temporalité réelle (axe d'instants ordonnés), mais est déconnectée de toute référence temporelle absolue. Elle sert juste à placer des procès modèles les uns par rapport aux autres, sans qu'ils n'aient d'ancrage absolu (il n'est pas possible de les dater).
Par ailleurs, un procès modèle possède les caractéristiques des procès classiques. On lui attribue notamment les deux intervalles définis dans le modèle SdT, l'intervalle de procès [B1, B2] et l'intervalle de référence [I, II].

### *3.3.2. Relations entre procès modèles*

Plusieurs types de relations peuvent être établis entre plusieurs procès d'un même modèle itératif (et uniquement au sein d'un même modèle itératif). Nous n'avons pas encore travaillé en détail sur ces relations, aussi en prosons-nous un simple aperçu :

1. relation temporelle, exemple (12)
2. relation causale, exemples (13)
3. relation méronomique (élaboration : un procès est une partie constitutive d'un autre), exemple (14).

(12) Elle arrivait après Pierre
(13) Pierre tombait car Marie le poussait / Quand il se mettait à pleuvoir, nous sortions nos parapluies.
(14) Les parties de boston commençaient par la distribution des cartes

Une relation temporelle entre deux procès modèles fait intervenir l'une des relations définies par Laurent Gosselin dans le modèle SdT. Il s'agit généralement de relation de coïncidence entre intervalles de référence, par une opération de saturation.
Remarque : il sera sans doute nécessaire d'étoffer le modèle itératif avec des intervalles circonstanciels modèles, pour traiter par exemple (15) ou (16) :
(15) Chaque fois, les cartes avaient été préparées **la veille**.

(16) Chaque fois, les cartes étaient préparées en **10 minutes**.
Notons que ces circonstanciels doivent s'inscrire dans la temporalité modèle, et donc logiquement ne comporter aucun ancrage dans la temporalité réelle. Dans (17), on peut supposer que la temporalité modèle représente une « année modèle », puisque l'itération est introduite par «chaque année», et cette temporalité modèle comporte donc un (et un seul) 10 juillet. Par contre, ce 10 juillet ne peut être daté de façon absolue, comme en (18).
(17) Chaque année, ils préparaient le feu d'artifice dès le 10 juillet
(18)* Chaque année, ils préparaient le feu d'artifice dès le 10 juillet 1984

Une relation d'un type non temporel peut avoir des conséquences temporelles. Par exemple, sans doute, la succession pour ce qui est de la causalité, le recouvrement (overlaps) pour la méronomie, etc.

*3.3.3. Constructions récursives : l'itération comme procès modèle.*
Enfin, une itération peut elle-même constituer un procès modèle, et donc être inscrite dans une autre itération, récursivement.

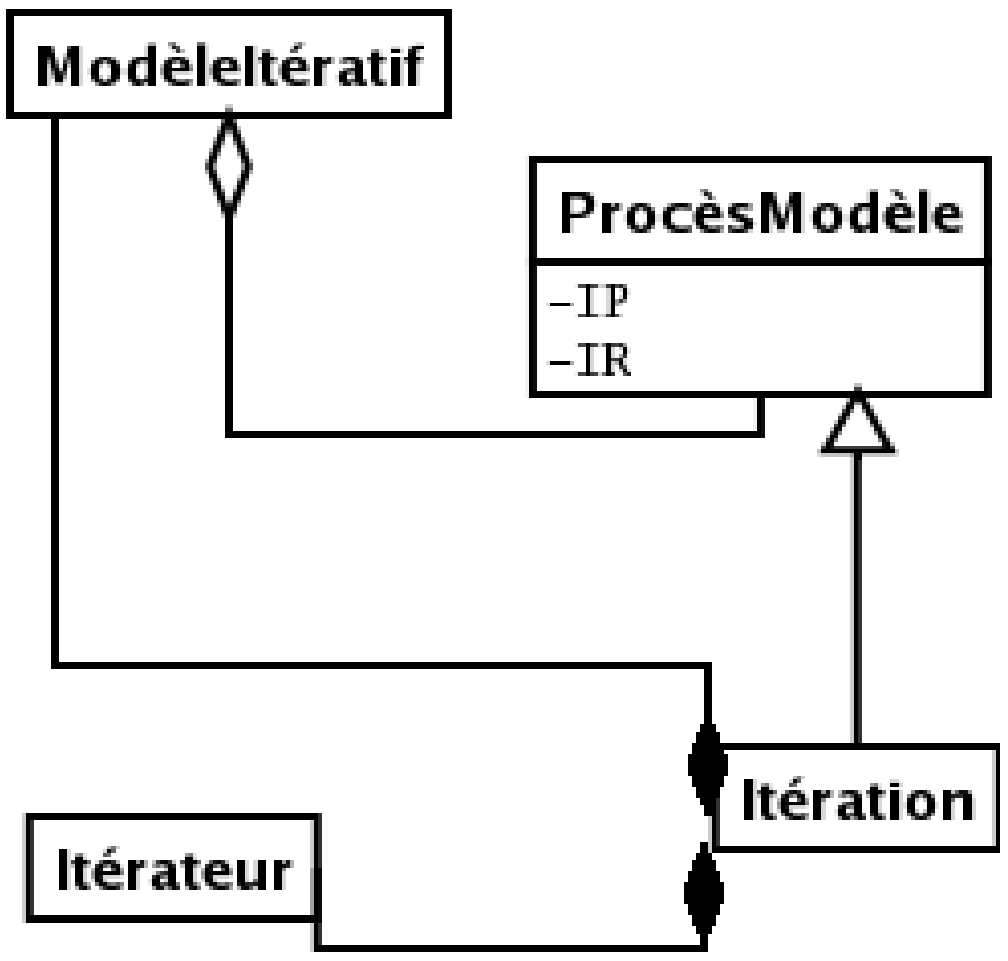

Ainsi, nous pouvons rendre compte de l'exemple suivant :

(19) Tous les dimanches, ils se baignaient deux fois.

Toutefois, lorsqu'une Itération sert de ProcèsModèle à une autre itération, ses itérés s'inscrivent dans une temporalité modèle. Ce n'est que l'itération de plus haut niveau qui s'inscrit dans la temporalité conventionnelle.

# 4. Mise en œuvre de la temporalité modèle : liens entre procès modèles

La temporalité modèle permet de s'affranchir des difficultés de manipulation et de représentation posées par les itérations. Tout revient finalement à considérer, dans cette temporalité, les procès comme s'ils n'étaient pas itérés. Nous retrouvons, en quelque sorte, un équivalent de la temporalité conventionnelle.

## 4.1. Procès modèles concomitants

Analysons l'exemple suivant :

(20) Depuis qu'il était marié, chaque dimanche, il nettoyait sa voiture quand sa belle-mère arrivait.

Il s'agit d'une itération dont l'itérateur est calendaire régulier, basé sur  la répétition des dimanches, induisant donc une période de longueur une semaine.
Par ailleurs, le cadre de cet itérateur à un début donné par le mariage du sujet, et une fin non spécifiée.
Enfin, le modèle itératif, induit par « il nettoyait sa voiture quand sa belle-mère arrivait » est composé de deux procès modèles. Nous devons travailler dans la temporalité modèle pour étudier comment ces deux procès s'agencent mutuellement. Ici, cette temporalité (dont les propriétés dépendent de l'itérateur) a une longueur d'une semaine, et englobe donc, nécessairement, un dimanche modèle. Dans cette temporalité modèle, nous pouvons appliquer les relations classiques de la théorie SdT :

- « au moment où sa belle-mère arrivait » introduit un aoristique, induisant une coïncidence de l'intervalle de procès avec l'intervalle de référence.
- « il lavait sa voiture » introduit un inaccompli (d'autres interprétations comme l'inchoatif seraient sans doute possibles, nous en choisissons une), induisant un recouvrement de l'intervalle de référence par l'intervalle de procès : on focalise sur une partie interne du procès.

Afin de saturer l'intervalle de référence relatif à « laver », celui-ci vient coïncider avec celui de « arriver », conformément à la théorie SdT. Nous obtenons donc, dans la temporalité modèle, le schéma suivant :

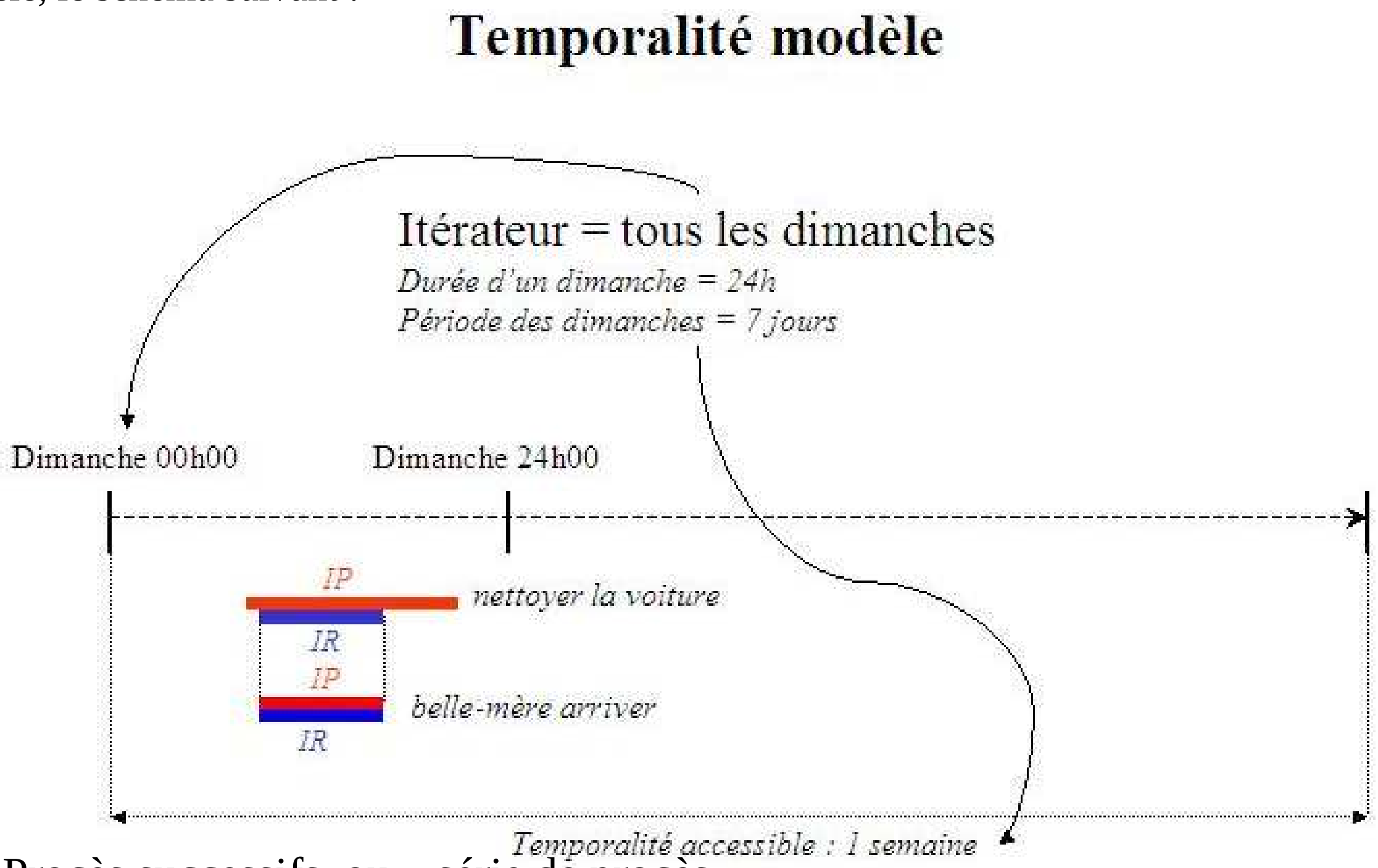

## 4.2. Procès successifs, ou « série de procès »

Nous pouvons, de façon similaire, rendre compte des « séries itératives de procès différents » évoqués p.10 du chapitre précédent. Il s'agit en fait d'un cas particulier de modèle itératif qui comporte plusieurs procès modèles se succédant temporellement.
En reprenant l'exemple (15a) du chapitre précédent, nous construisons un modèle itératif constitué de 4 procès modèles, correspondant respectivement à *descendre*, *conduire*, se *diriger* et *entrer*. Pour chacun d'entre eux, l'intervalle de procès et l'intervalle de référence coïncident (aoristiques). Par ailleurs, il y a successions des différents intervalles de procès (et donc de référence).

## 4.3. Cas général

Dans les deux parties précédentes (4.1 et 4.2), nous avons vu deux cas particulier, respectivement la concomitance et la succession. En fait, dans le cas général, on construit simplement les différentes relations comme on le fait dans la temporalité conventionnelle, ce qui peut mener à des structures plus ou moins complexes :

(21) A huit heures, le jeu débutait. A cette heure là, la maison était généralement assez calme depuis un long moment. Les parties s’enchaînaient alors. La soirée se terminait vers 22 heures, par un apéritif.

Nous avons en (21) un exemple mêlant les deux cas précédent, avec un succession de *débuter*, *enchaîner* et *terminer*, tandis qu’il y a concomitance, et plus précisément recouvrement, de *être calme* sur *débuter*.

# 5. Quelques constructions et analyses

(22) Souvent, après avoir fini nos devoirs, nous regardions un film.

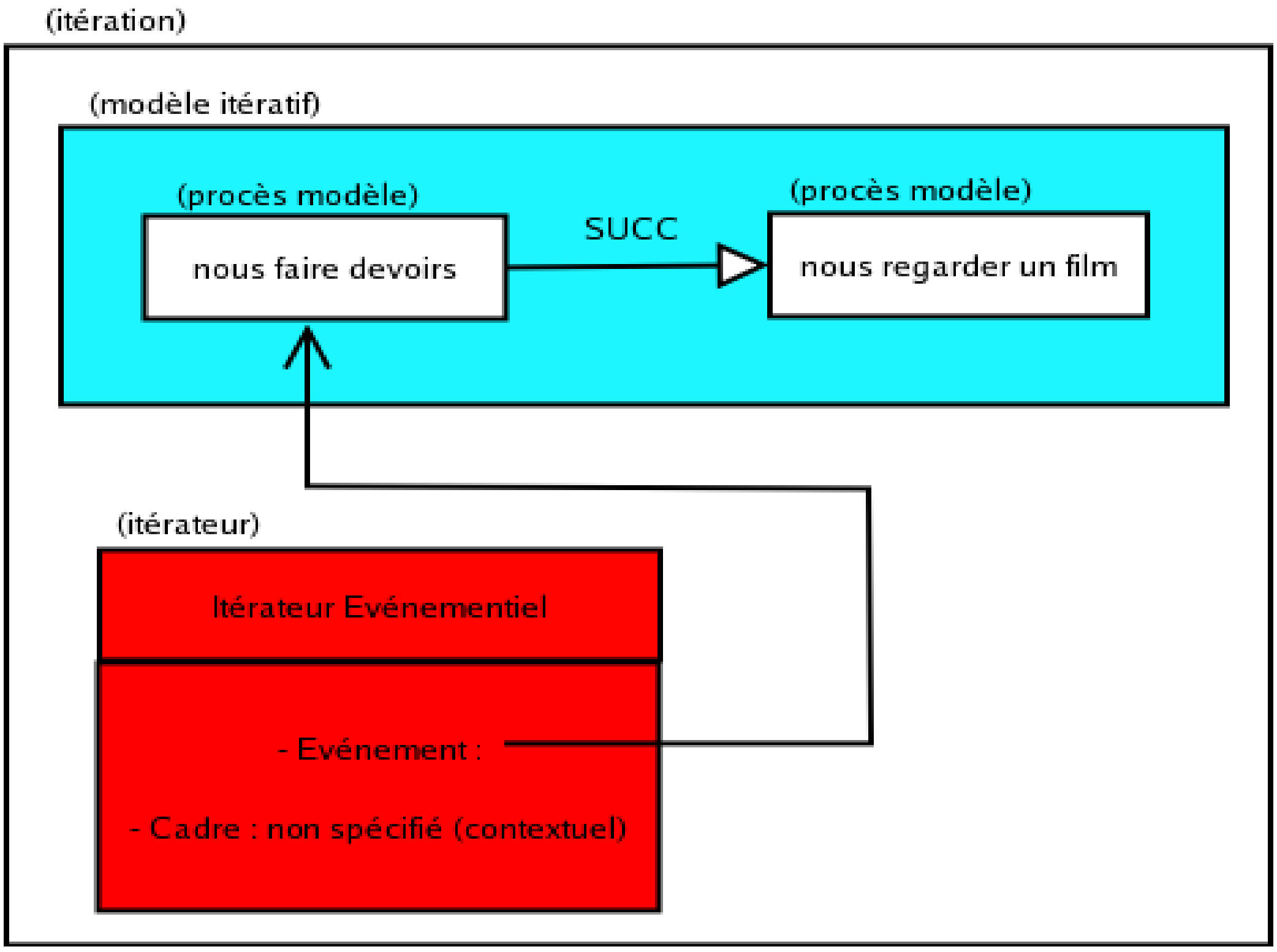

Dans cet exemple, l'itérateur retenu est une instance de la classe ItérateurEvénementiel. Il a donc un attibut de type “événement”, qui pointe vers un certain procès modèle déclencheur. La particularité d'une telle construction est que le même procès modèle sert à la fois à construire le modèle itératif, c'est-à-dire ce qui va être itéré, mais aussi l'itérateur, c'est-à-dire ce qui va servir à itérer. Dit autrement, cela signifie ici que le procès modèle “faire les devoirs” va déterminer tous les itérés, i.e. à chaque fois que l'événement correspondant se produit dans le temps, un itéré doit lui correspondre (c'est-à-dire l'aglomérat “faire devoirs” puis “regarder film”), et aussi que ce procès modèle fait partie intégrante de l'itéré.

L'exemple choisi débute par l'adverbe “souvent”, dont il est intéressant ici d'étudier l'entrelacement avec l'itération. On peut en considérer deux sens possibles, paraphrasés resp. par (23) et (24) :

(23) Il arrivait souvent que nous fassions nos devoirs, lesquels étaient suivis du séance de cinéma.

(24) Nos devoirs étaient souvent suivis d'une séance de cinéma.

Ces deux cas sembles représentables dans le modèle. Pour ce qui est de (23), l'itérateur événementiel pourra être assorti d'une donnée fréquentielle sur l'apparition de l'événement. Pour ce qui est de (24), l'entrelacement est plus intime, puisque “souvent” vient porter sur la flèche “SUCC”. Cette dernière devrait donc, à l'avenir, être réifiée (considérée comme un objet ayant ses propres attributs), afin de contenir une information fréquentielle. En première approche, il s'agirait de coder par ex. que cette SUCCession est avérée dans 80% des cas pour “souvent”, 20% pour “parfois”, etc.

Continuons notre exemple (22) par :

(25) Mais parfois, on préférait un poker.

On pourrait traiter ce cas sans trop déroger aux contructions précédentes en créant un ProcesModeleAlternatif, dont le rôle serait de coder de façon formelle l'alternative entre plusieurs procès modèles possibles.

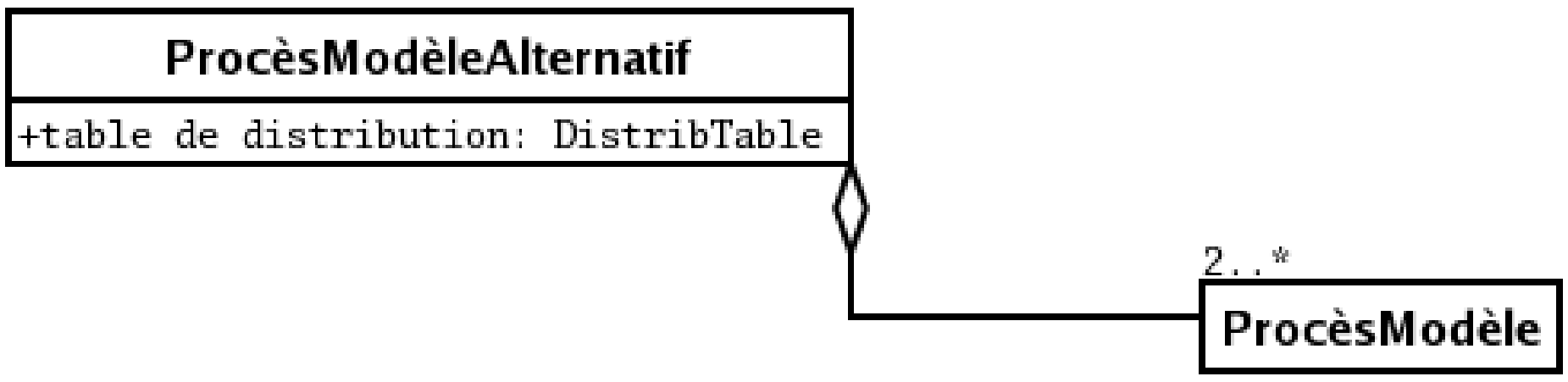

Ici, nous aurions une instance de ProcèsModèleAlternatif qui possèderait 2 procès modèles, resp. le procès modèle “voir un film” et le procès modèle “jouer au poker”. Il possèderait aussi une table de distribution indiquant par exemple la fréquence relative de l'alternative (ici par ex. 80% / 20%, vu que “parfois” induit une fréquence moindre du procès modèle “poker”).

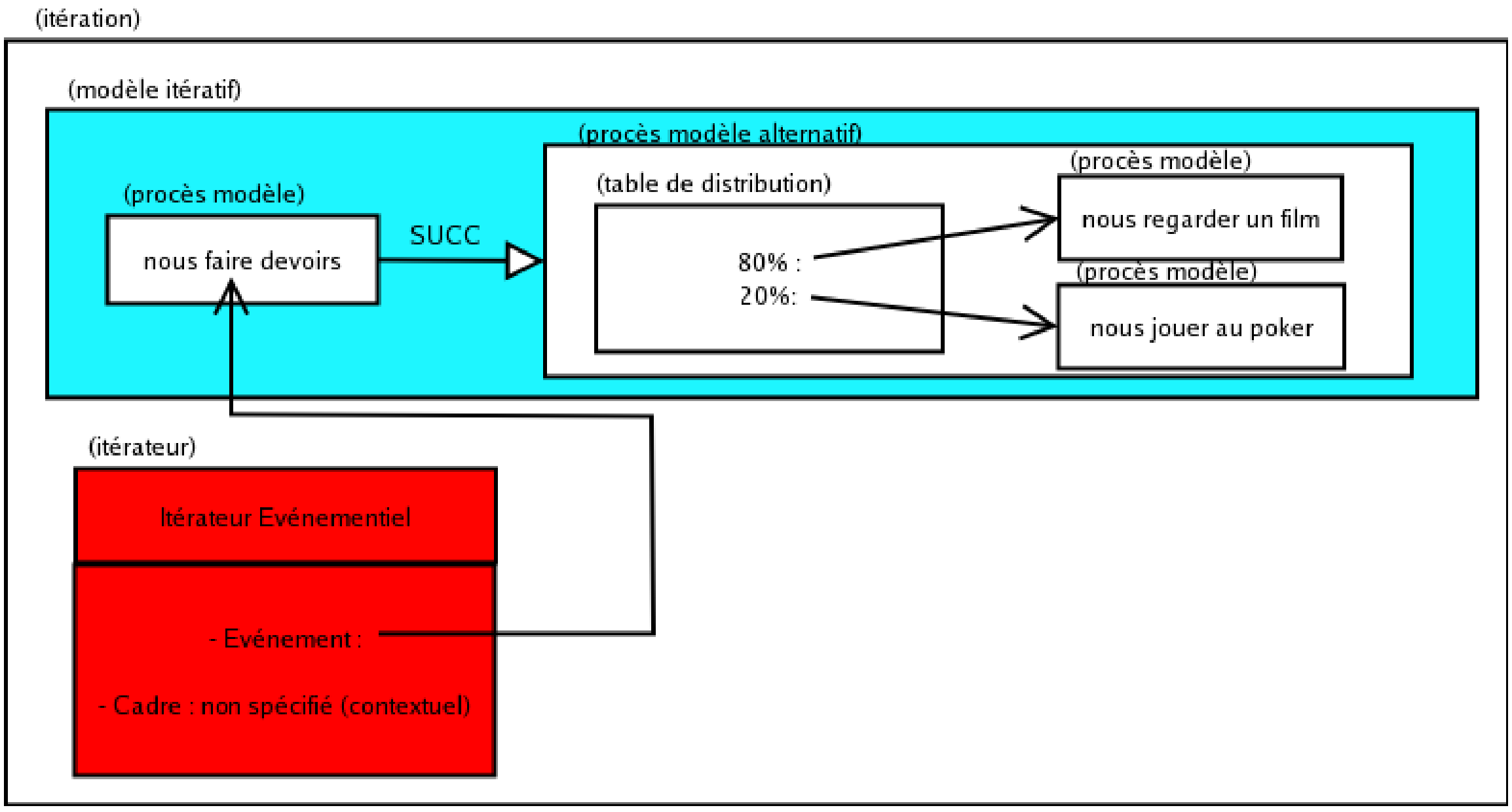

Remarque : ce formalisme est en fait bien plus général que les seules itérations. Il devrait pouvoir rendre compte de (26) qui n'en est pas une :

(26) Demain, nous irons à la mer ou à la campagne.

Passons à un exemple faisant intervenir un itérateur hybride, qui n'est pas encore défini dans le modèle : il s'agit d'assortir l'itérateur événementiel d'une contrainte supplémentaire, pour traiter un exemple comme (27).

(27) Quand elle venait le lundi, nous regardions un film

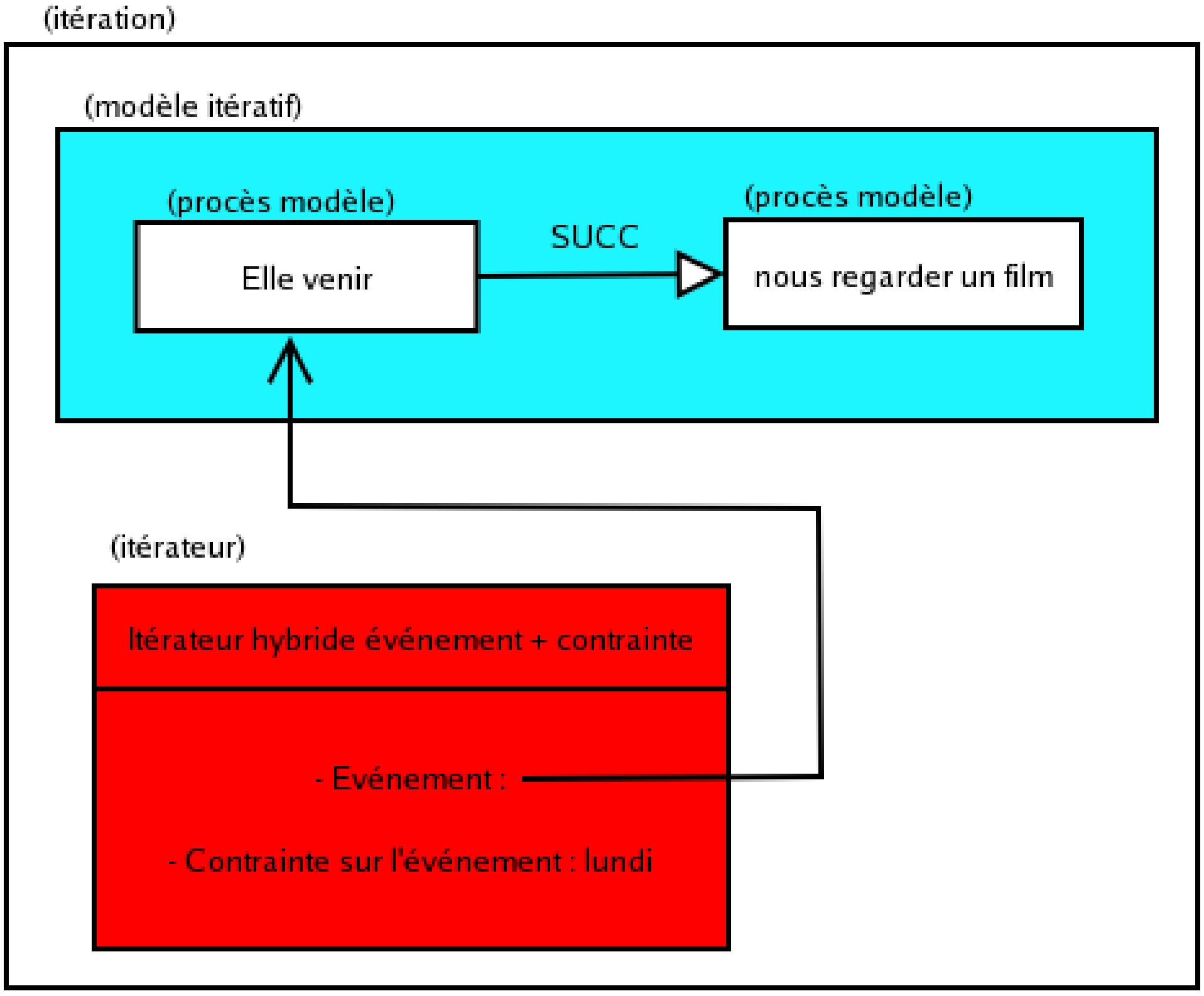

Enfin, Il faudra travailler aussi sur des itérations impliquant des états, comme dans l'exemple suivant :

(26) Elle vient toujours les mains vides.

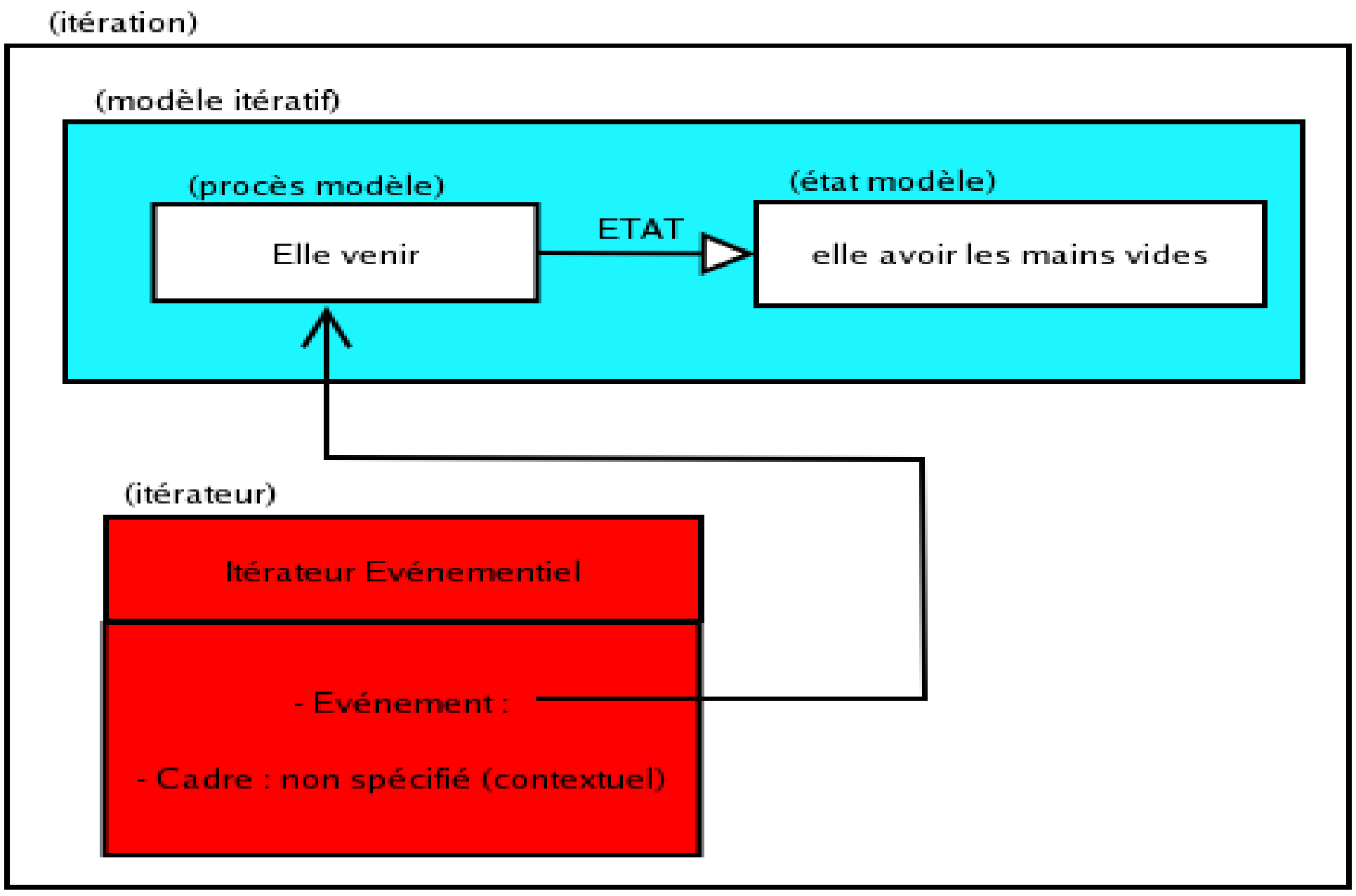

Par ailleurs, il sera nécessaire de disposer de mécansimes de rattachement d'itérations, lorsqu'il s'avère qu'une itération est une élaboration d'une autre (au sens des relations discursives “à la SDRT”), ou vient en compléter un autre.
Considérons par exemple la paire d'exemples suivante :

(29) Dès qu'ils arrivaient, nous faisions une partie
(30) Dès qu'il repartaient, on rangeait les cartes.

(note : l'emploi de “repartir” plutôt que “partir” semble être une marque du rattachement logique, et donc temporel, qui existe entre les deux phrases).

Imaginons qu'un traitement automatique nous produise une représentation par phrase, reportées resp. à gauche et à droite dans la figure suivante :

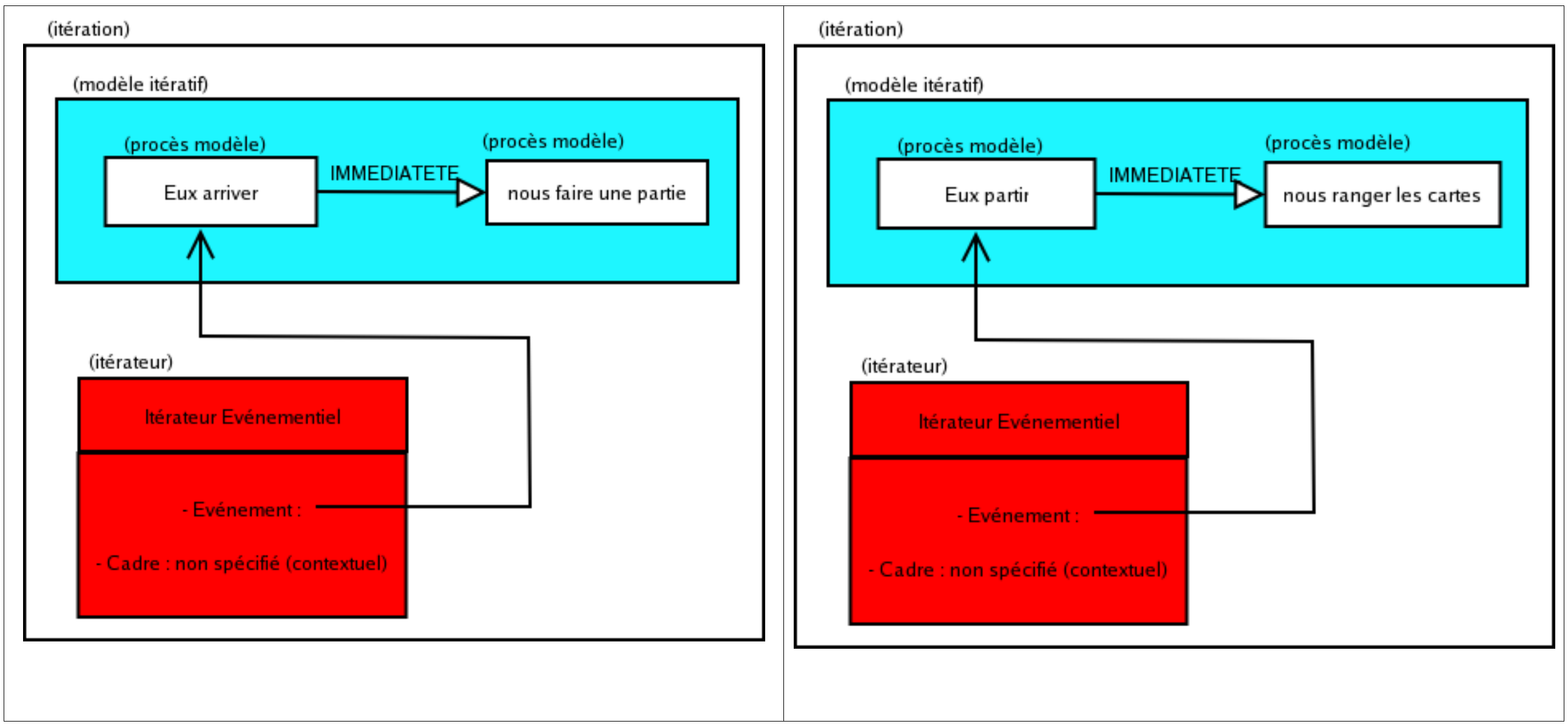

Il faudrait être capable par la suite d'indiquer qu'il s'agit d'une seule et même itération, et parvenir à la représentation suivante :

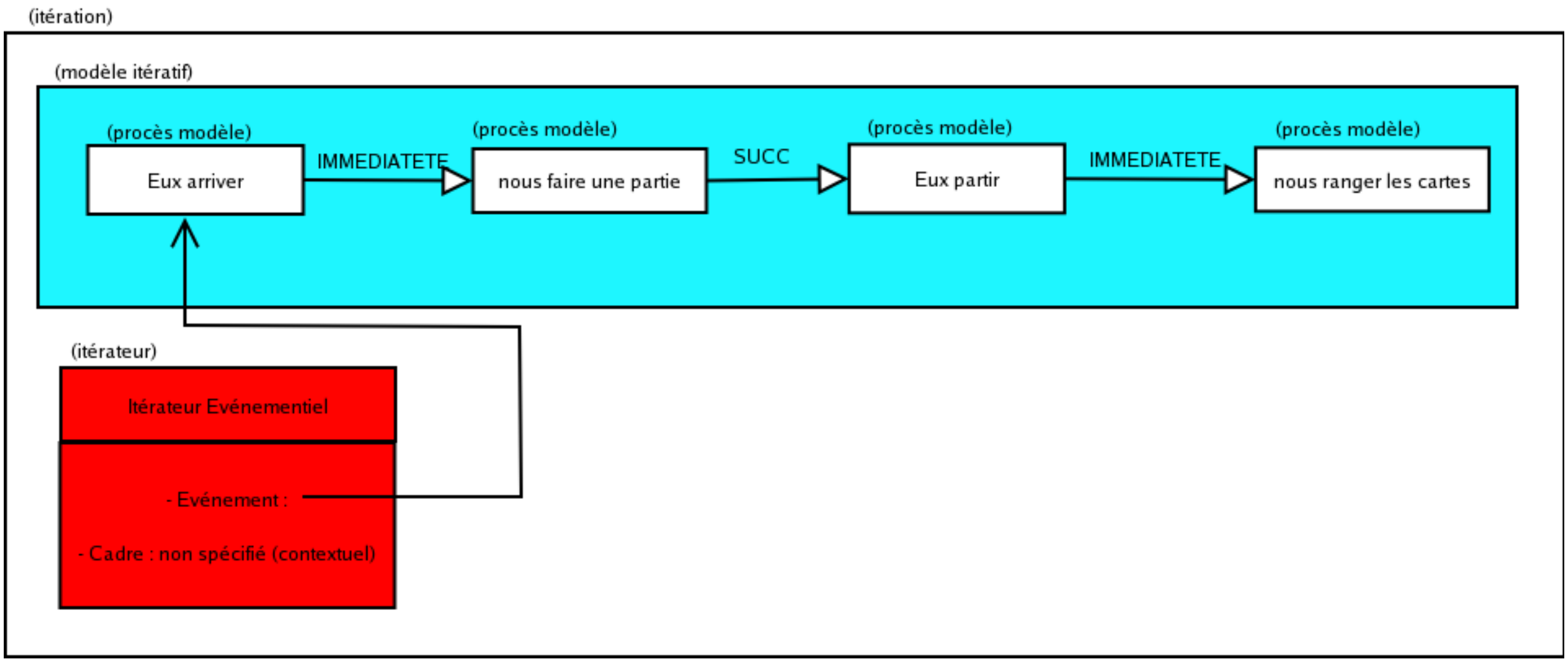

# 6. Itération et aspect

En ce qui concerne la temporalité modèle, nous avons vu que l'aspect était finalement traité de façon classique, vu que chaque procès se voit attribuer un intervalle de procès et un intervalle de référence. Cet outillage permet en particulier d'agencer les différents procès modèles les uns par rapport aux autres (cf. concomitance, succession, etc.).
Il nous reste cependant à voir l'aspect de l'itération elle-même, qui s'inscrit dans la temporalité conventionnelle.
Nous avons déjà associé à une itération un intervalle de procès, que nous avons appelé **cadre** (celui-ci est porté par l'itérateur). Il reste donc à lui attribuer un intervalle de référence, comme cela a déjà été présenté dans le chapitre précédent. Pour autant, nous n'avons pas pu, faute de temps, étudier les différentes relations possibles entre cet intervalle de référence et les autres intervalles. Nous considérerons donc, pour le moment, que toute relation est possible.

En guise d'illustration, reprenons l'exemple (9) du chapitre de Laurent Gosselin, reporté ici en exemple (31) :

(31) Depuis deux mois, il mangeait en 10 minutes.

Il s'agit là, comme l'a présenté Laurent Gosselin, d'une itération vue globalement sous un aspect inaccompli, c'est-à-dire que nous nous focalisons sur une partie interne de l'ensemble de la période où le sujet du verbe avait pour habitude de manger en 10 minutes. En des termes plus formels, l'intervalle de procès de l'itération (encore appelé cadre dans cette partie) recouvre l'intervalle de référence (ce qui est montré).
Les itérés sont quant à eux présentés de façon aoristique, c'est-à-dire que le procès modèle et l'intervalle de référence modèle coïncident.

Il nous semble par contre, pour le moment, qu'il n'est pas possible de spécifier davantage la position de l'intervalle de référence de l'itération par rapport aux autre intervalles (rappel : il est dans notre exemple en relation de recouvrement par l'interalle de procès de l'itération). En effet, comparons les deux exemple (32) et (33) comme suites à donner à (31) :
(32) Mais cette fois là, il prit son temps et mangea goulûment
(33) Mais ce jour là, il prit son temps et mangea goulûment
On constate en (32) que ce qui est montré de l'itération issue de (31) coïncide exactement avec un itéré (celui qui correspond à « cette fois là »), tandis qu'en (33), ce qui est montré de l'itération englode plusieurs itérés, vraisemblablement 3, correspondant aux trois repas d'une certain journée (« ce jour là »).
Nous pouvons ajouter que ce calcul se fait a posteriori (lors de l'interprétation de (32) ou (33)), et non directement lors de l'interprétation de (31).

# 7. Discussion et perspectives

## 7.1 Itérateurs et intervalles circonstanciels : point de vue « entensionnel » versus point de vue « modèle »

Il y a très clairement un lien entre : 1) la notion d'itérateur ici décrite et 2) la description des intervalles circonstanciels (pour une itération) en termes de série, décrite dans le texte de Gérard Becher et Patrice Enjalbert.
Une première idée, qui a le mérite de la simplicité, serait de dire que la série introduite par un circonstanciel (plus généralement spécifiée par un adverbial) sert d'ancrage temporel pour la répétition du procès modèle. L'itérateur en quelque sorte se réduirait à l'intervalle non convexe du circonstanciel (de l'adverbial). C'est d'ailleurs le cas en ce qui concerne l'une des classes d'itérateurs, la classe ItérateurParIntervalle (et donc, par héritage, sa classe fille CalendaireRégulier).
Toutefois, la discussion a fait apparaître une différence de point de vue entre les deux approches. Dans le présent texte, nous insistons sur un mécanisme cognitif de focalisation sur un procès « modèle », avec sa propre temporalité, détachée en quelque sorte de l'axe temporel « conventionnel ». Alors que le modèle proposé par G. Becher et P. Enjalbert, est guidé par un point de vue clairement « extensionnel ».

Dans le présent modèle, nous avons posé qu'un itérateur était un objet permettant d'itérer un modèle itératif dans le temps. Il s'agit d'un mécanisme cognitif de focalisation sur un proçès « modèle », avec sa propre temporalité détachée en quelque sorte de l'axe temporel

« conventionnel ». Nous considérons par ailleurs que le mécanisme associé est varié, et que la donnée d'un intervalle non convexe, une série, n'en est qu'un cas particulier.
D'un autre côté, le texte de Gérard Becher et Patrice Enjalbert présente la création d'intervalles circonstanciels (pour une itération) en termes de série comme moteur de l'itération, selon un point de vue clairement « extensionnel ».

Le lien entre ces deux point de vue est une question intéressante. A la fois d'un point de vue théorique : quelle est la meilleure modélisation cognitive ? les deux points de vue cohabitent-ils (nous serions enclins à le penser) ? Et du point de vue des traitements automatiques, pour lequel une forme de synthèse est tout à fait nécessaire.Voici trois points consituant l'amorce d'un débat.

A. Les itérateurs **par intervalles**

Nous commençons par le cas qui pose le moins de differences de points de vue, puisque le travail réalisé par G. Becher et P. Enjalbert (qui fournit une « série ») peut être utilisé pour « nourrir » un « itérateur par intervalles ». Loin d'un opposition, il nous semble que les deux approches peuvent se compléter.
Plus précisément, une instance particulière d'itérateur par intervalles pourra se voir fournir la « série » produite par l'approche extensionnelle, et de proposer un itéré par intervalle convexe de la série. Dans l'hypothèse d'un implantation réelle de ces deux approches, l'objet « itérateur par intervalles » ne créerait pas (de façon extensionnelle) l'ensemble des itérés, potentiellement infini (ex : « tous les matins du monde »), mais interrogerait le module qui a produit la série lorsque nécessaire (par exemple pour savoir si tel jour est concerné, ou pour connaître l'intervalle relatif au n-ième itéré, etc.). En parallèle, cet objet instancierait des itérés lorsque ceux-ci sont explicitement mentionnés linguistiquement (par exemple, dans « Tous les dimanches, ils vont à la pêche. Dimanche dernier, ils ont eu du bar », il est nécessaire d'une part de disposer de l'operateur « série » capable de restituer, potentiellement, tous les itérés, mais il est aussi nécessaire, d'autre part, de mémoriser les particularités propres à l'itéré auquel le texte réfère par « dimanche dernier »).

B. Les itérateurs **numéraires**

Comme nous venons de la voir, dans « tous les lundis » (calendaire régulier), c'est bien intervalle non convexe qui est le moteur de l'itérateur, celui-ci créant un itéré par convexe (donc ici, un par lundi).
Au contraire, dans « trois fois » (numéraire), nous pensons que l'itérateur n'a besoin d'aucun intervalle non convexe pour opérer le processus d'itération. Il s'agit juste de créer exactement 3 itérés. Certes, ces itérés étant ancrés dans le temps, et étant par construction disjoints deux à deux temporellement, il résulte de leurs intervalles associés un intervalle non convexe. C'est en cela que les différents points de vue convergent. Mais cet intervalle résultant est obtenu en bout de chaîne, et n'est donc pas, de notre point de vue, l'initiateur de l'itération.
(33) De toute sa vie, il n'a vu que trois fois son père. C'était en 1987, en 1989, et en 1995.
(34) En 1987, en 1989 et en 1995, il a vu son père.
Dans l'exemple (33), il y a de notre point de vue création de trois itérés (disjoints temporellement) dans la première phrase. Puis, la deuxième phrase vient localiser (avec un grain assez gros) chacun des itérés sur l'axe temporel (chacune des années évoquées devant recouvrir l'un des itérés). Par contre dans (34), c'est un intervalle non convexe (une « série ») qui est à l'origine de l'itération (et le fait qu'il y a trois itérés n'est pas marqué linguistiquement, il est inféré en comptant le nombre de convexes composant le non convexe), et un itéré est placé dans chacun des composants convexes.
Il est intéressant de constater que ces deux exemples sont sémantiquement proches (création de 3 itérés localisés dans le temps), mais mettent en jeu des mécanismes différents.

C. Les itérateurs **événementiels**

Le cas des itérateurs événementiels nous semble plus délicat.
(35) Quand je me promène, je rencontre Jean
Dans la partie 3.5. du chapitre précédent, il est proposé qu'une extension de la théorie SdT aux itérations permette de créer une série comme intervalle circonstanciel associé à la subordonnée.
Dans la présente partie, nous « évacuons » la question des intervalles circonstanciels vis-à-vis de l'itération, en faisant du procès même de la subordonnée le moteur de l'itération : pour chaque occurrence (événement) du procès décrit dans la subordonnée (je se promener), il y a création d'un itéré. De plus, le procès de la subordonnée fait partie intégrante de l'itération : ce qui est répété est un modèle itératif où le narrateur se promène et rencontre Jean au cours de sa promenade. C'est plus flagrant encore dans l'exemple (38) où le procès décrit dans la principale semble être en relation de méronomie avec celui de la surbordonée, ce dont ne rendrait pas compte une approche basée sur des intervalles qui présenterait ceci comme une coïncidence (au sens « un hasard ») temporelle : le fait que le sujet emprunte le boulevard Malsherbes n'est pas à rattacher (uniquement) à des intervalles temporels, mais doit être présenté comme faisant partie intégrante de « il venir ». C'est ce dont le modèle itératif des itérateurs événementiels essaie de rendre compte en intégrant le procès relatif à la subordonnée, et en permettant ainsi à ce dernier d'être en relation intime avec les autre procès modèles du modèle.
Il reste que cette question est encore très ouverte, et il nous semble même que l'on puisse établir un certain continuum entre des itérations plutôt induites par l'intervalle non convexe (série) issu de la subordonnée, comme en (36), et des itération clairement articulée autour du procès de la subornée, comme en (38).
(36) Quand j'ai le temps / Quand il fait beau, je me promène.
(37) Quand il vient, il est heureux.
(38) Quand il vient, il passe par le boulevard Malsherbes
Il serait intéressant de comparer plus avant les différences et similitudes sémantiques qu'il y a entre (37), construit avec « quand », et (39), construit avec « à chaque fois que ».
(39) A chaque fois qu'il vient, il est heureux.

Finalement, les deux points de vue se rejoignent quant au résultat produit (on retrouve toujours, en fin de compte, une série), la distinction se faisant sur la façon d'appréhender la construction de l'itération.

## 7.2. Perspectives

Comme nous avons pu le voir en 7.1, une perspective intéressante est d'approfondir les rapprochements possibles entre les différents points de vue. Tout d'abord d'un point de vue théorique, notamment dans une perspective de modélisation cognitive de l'itération. Les deux approches pourraient correspondre à deux processus cognitifs parallèles. Puis d'un point de vue pratique, puisque nous avons vu que dans certains cas les deux approches pourraient d'ores et déjà collaborer.
Notons pour finir que la notion de procès modèle pose en soi d'intéressantes questions théoriques. Il semble que les mécanismes mis en œuvre aient une certaine similarité avec la notion de monde possible du Eco de *Lector in fabula* ou avec les espaces mentaux de Fauconnier. Et un lien également avec les modalités, dans la mesure où il s’agit bien d’un procès « virtuel ».

# Sémantique des Compléments Circonstanciels Temporels Itératifs (CTI) : Bases algébriques

Gérard Becher, Patrice Enjalbert

## Présentation

On s'intéresse dans ce rapport aux compléments circonstanciels temporels indiquant une itération (CTI). Quelques premiers exemples, sur lesquels nous reviendrons, sont les suivants :

4. *Tous les lundis*
5. *Tous les lundis des mois de mars*
6. *Les deuxièmes lundis de mars*
7. *La plupart des lundis*
8. *Les lundis de mars, de 8h à 10h.*
9. *Quand je me promène [je rencontre Jean]*
10. *Souvent les lundis d'été*
11. *Un lundi sur 2*
12. *Souvent quand je me promène [je rencontre Jean]*

Le but à terme est :

- de donner *une sémantique formelle* à ces expressions ;
- de *la calculer à partir du texte* ;
- d'effectuer des *calculs/raisonnements* sur les représentations produites.

La présente étude constitue la première étape de ce programme. D'une part nous introduisons un modèle algébrique dans le cadre duquel peuvent être décrites des représentations sémantiques des CTI (section 2). D'autre part, nous décrivons les principales constructions linguistiques constituant les CTI, de manière à esquisser le principe d'une sémantique compositionnelle effective (section 3). Cette première étude est donc destinée à constituer la base d'une grammaire décrivant de manière relativement complète les CTI et leur sémantique. Ce point et d'autres prolongements seront abordés en conclusion (section 4).

Donnons ici les idées générales du modèle, à partir des exemples (1) à (8) ci-dessus. On s'accordera à considérer que chacune de ces expressions indique une succession d'intervalles temporels. Le concept de base du modèle est alors la *série temporelle* (ou *série* tout court) : suite d'intervalles, sans chevauchement, dans un rapport de consécution temporelle. D'une certaine manière, on pourra reconnaître la notion d'intervalle non convexe (Ligozat 1991), mais avec une structure *d'énumération en séquence*, permettant de parler « d'intervalle suivant », de « $n^{ème}$ intervalle » etc.  (un peu à la manière de (Cukiermann & Delgrande, 1996)..

En première approximation, nous dirons que la sémantique devra faire correspondre à un CTI une *série* de ce type. (Une vision un peu plus précise sera présentée en son temps dans la section 3). La sémantique visée a donc la forme d'une fonction dont la source est l'ensemble des expressions linguistiques de type CTI, et le but l'espace algébrique des séries, que nous allons construire. Soit, en notant [[C]] la sémantique formelle de C :

Ensemble des CTI ⟶ Espace des séries temporelles

$$C \longmapsto [[C]]$$

En particulier un nom calendaire, tel que *jour*, *mois*, *année*, *heure*, mais aussi *lundi*, *mardi… janvier*, *février… printemps*, *été…* dénotera dans le modèle une série particulière (constante du modèle), notée ici en petites majuscules : JOUR, MOIS, LUNDI, JANVIER, ETE… Ainsi : [[*lundi*]] = LUNDI, [[*mars*]] = MARS, etc. Nous devons ensuite prévoir un certain nombre *d'opérations algébriques sur les séries*, permettant d'interpréter les diverses constructions linguistiques intervenant dans les CTI. Voici, de manière informelle, quelques-unes de ces constructions.

Considérons d'abord l'expression (2) : *Tous les lundis des mois de mars*. Sa sémantique fait intervenir les deux séries prédéfinies LUNDI et MARS. (2) elle-même est interprétée comme la sous-série de la première constituée des éléments inclus dans un des éléments de la seconde. Nous appelons cette opération la *restriction* d'une série S1 par rapport à une autre S2, notée S1/S2. Ici donc nous aurons donc :

[[*Tous les lundis des mois de mars*]] = LUNDI / MARS

Une variante de la restriction, notée $S1/_nS2$, dans laquelle on « compte » les éléments de la S1 pour garder le $n^{ème}$ de chaque composante de S2, permet de récupérer (3) : *Les deuxièmes lundi de mars*, dont la sémantique est donc : LUNDI $/_2$ MARS (cf figure 1).

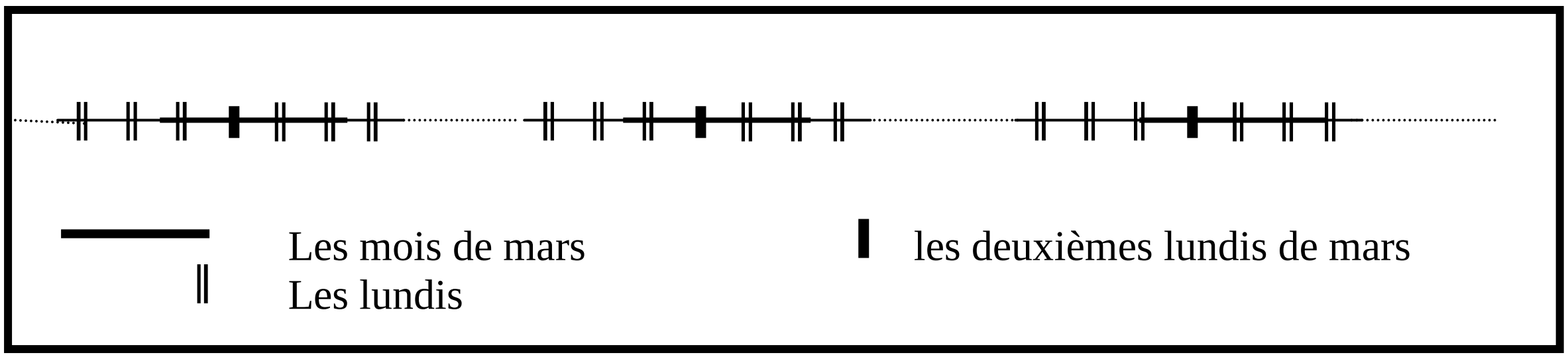

**Figure 1**

Considérons maintenant de plus près la structure syntaxique des expressions, en observant l'exemple (4) *La plupart des lundis*. Nous avons le nom *lundis* précédé d'un déterminant exprimant une quantification *La plupart des*. Sa sémantique doit donc être une série extraite de la série des lundis, composée d'une majorité d'intervalles de cette série. Clairement, il existe de très nombreuses séries possédant cette propriété. Parmi toutes, l'expression (4), insérée dans un texte, incite à en introduire une dans l'univers du discours. Notons ε une « fonction de choix » qui, à un ensemble E associe un élément de E. Nous aurons donc (semi-formellement ici ) :

[[*La plupart des lundis*]] =

ε {S tq 1) S est extraite de LUNDI et

2) est constitué d'une « forte majorité » des intervalles constituant LUNDI }

où l'expression « forte majorité » est évidemment qualitative et demande des considérants d'ordre pragmatiques pour être interprétée.

Nous disposons donc d'une notion générale de *quantification dans une série* (pour en extraire une sous-série), applicable dans différents contextes : déterminants (*les*, *un*, *tous les*, *certains*, *la plupart de…*) ; mais aussi adverbes fréquentatifs (*souvent*, *parfois*, *régulièrement…*) lorsqu'ils portent comme dans l'exemple (5) (*Souvent, les lundis d'été*), sur une expression dénotant déjà une série ; ou encore certaines formes de quantification explicite comme dans (6) : *Un lundi sur deux.*
Considérons pour finir les exemples (7) et (8) : (*Souvent*) *quand je me promène je rencontre Jean.* La question est de savoir comment intégrer des subordonnées circonstancielles dans notre modèle sémantique, de manière homogène au traitement des compléments de type calendaire. Intuitivement, l'idée est de considérer une subordonnée temporelle telle que *quand je me promène* comme spécifiant une série, dont les composantes sont les intervalles de validité de la propriété *'je me promène'* (notons la $S_{je\ me\ promène}$). Nous avons donc bien le même type d'objet sémantique que dans le cas des expressions calendaires. Un bon test est constitué par la sémantique de l'adverbe *souvent* dans (8) *Souvent quand je me promène je rencontre Jean* qui fonctionne exactement comme dans (5) *Souvent les lundis d'été* : il opère sur $S_{je\ me\ promène}$ pour créer une série extraite composée d'un ensemble « très majoritaire » d'éléments de cette série. Évidemment il s'agit ici de l'idée intuitive et une intégration précise dans le modèle SdT doit être faite : elle sera également esquissée en section 3.

# Modèle algébrique

Nous définissons ici formellement la notion de série ainsi qu'un langage formel mettant en œuvre ces séries et permettant de capturer les expressions de la langue faisant référence à de telles séries.
Nous définirons également une notion d'intervalles dont les bornes sont des intervalles et que l'on rencontre communément dans des expressions telles que "de janvier à mars" par exemple.

## Présupposés

Le modèle algébrique est basé sur un modèle linéaire du temps : nous considérons un axe temporel structuré par une relation d'ordre total notée ≤. Les objets primitifs de notre théorie sont les intervalles convexes. Un intervalle convexe I est la donnée d'un couple ordonné de points, notés beg(I) et end(I). Si beg(I) = end(I), nous avons affaire à un intervalle convexe dégénéré (un point).
À partir de ces objets primitifs, nous construisons de manière classique la notion d'intervalle généralisé.

**Définition : intervalle (généralisé)**

> *Un intervalle généralisé consiste en la réunion d'un nombre fini d'intervalles convexes et/ou de points. Pour tout intervalle I, on définit le terme* beg(I) *(resp.* end(I)*) comme étant la plus petite (resp. la plus grande) valeur de* beg(J) *(resp.* end(J)*) pour tous les intervalles J composant I.*

**Relations sur les intervalles**

L'ensemble des intervalles constitue une structure partiellement ordonnée par la relation d'ordre ≤ induite par l'ordre sur les points temporels et définie par : $\forall$ I, J I ≤ J ssi end(I) ≤ beg(J).

Un point temporel x appartient à un intervalle convexe I ($x \in I$) si et seulement si beg(I) ≤ x ≤ end(I). De même un point temporel x appartient à un intervalle généralisé I si et seulement si x appartient à l'un des intervalles composant I ou est égale à l'un des points composant I.

L'ensemble des intervalles est également muni de la relation d'ordre partiel correspondant à l'inclusion (au sens large) des intervalles, définie par : $\forall$ I, J on a $I \subseteq J$ ssi $\forall x\ (x \in I) \Rightarrow (x \in J)$.

**Définition : enveloppe convexe d'un intervalle**

*Si I est un intervalle généralisé, on définit* convexify(I) *comme étant le plus petit intervalle convexe contenant I.*

*On généralise la définition de manière évidente au cas de l'enveloppe convexe d'un nombre fini quelconque d'intervalles.*

## La notion de série

*Motivations*

Les circonstanciels itératifs, objets de notre étude, sont très largement basés sur des noms calendaires, c'est-à-dire de noms intervenant dans le système calendaire (du français), qu'ils soient génériques (… , *jours*, *mois*, *année*, …) ou spécifiques (*lundi*, *mardi*…, *janvier*, *février*…). Une question importante est alors de donner une sémantique à ces noms. Nous nous placerons dans le cadre d'une sémantique extensionnelle, associant à un substantif l'ensemble des individus (réels ou objets du discours) tombant sous cette dénomination. Ainsi par exemple, au terme *mois* sera associé l'ensemble des mois s'inscrivant dans le cadre du discours étudié. Toutefois, par rapport au cas général, cet ensemble a les spécificités suivantes :

- c'est un ensemble d'intervalles convexes structuré par une relation de succession préservant l'ordre sur les intervalles
- si l'on admet que l'interprétation se fait toujours dans un cadre clos (le cadre du discours), cet ensemble est toujours fini et possède donc un premier élément.

Dans la suite, nous désignerons un tel ensemble par le terme de "*série*" : une série est donc un intervalle non convexe particulier muni d'une structure additionnelle permettant d'énumérer ses composants en respectant l'ordre sous-jacent.

*Définitions*

**Série**

*Une série S est un ensemble fini d'intervalles convexes ou de points tel qu'il existe une bijection croissante b de S vers une section commençante de N [1,2,…,n], c'est-à-dire une bijection b vérifiant* $\forall I \in S$, $\forall J \in S$, $I \leq J$ *ssi* $b(I) \leq b(J)$.

**Successeur**

*Pour tout intervalle I dans S, on note succ(I,S), ou plus simplement succ(I) s'il n'y a aucune d'ambiguïté, l'unique intervalle ou point (s'il existe) qui est l'image réciproque par b du nombre b(I)+1..*

**Ordre d'un élément**

*On appelle premier élément (et on note fst(S)) l'intervalle* $b^{-1}(1)$.
*On notera* $S_n$ *le* $n^{ème}$ *élément de S (donc* $b^{-1}(n)$*) et ordre(I,S) = n si* $I = S_n$.

**Remarque :**

Une série dégénérée est limitée à un seul intervalle.
Les intervalles constitutifs d'une série sont nécessairement disjoints. En effet, en supposant qu'il existe deux intervalles distincts non disjoints I et J d'une série S, nous n'aurons alors ni $I \leq J$, ni $J \leq I$. Or b(I) étant nécessairement différent de b(j) puisque b est une bijection, nous aurons soit $b(I) < b(J)$, soit $b(J) < b(I)$, d'où la contradiction. Il est également aisé de vérifier que la bijection b est nécessairement unique.
La notion de série est assez proche de la notion d'intervalle généralisé, à ceci près que nous regardons l'un et l'autre à des niveaux différents : un intervalle est un ensemble de points temporels tandis qu'une série est un ensemble d'intervalles convexes muni d'une structure adéquate.
Le lien entre ces deux points de vue est réalisé par la définition suivante :

**Extension d'une série**

*L'extension d'une série S est l'intervalle (probablement non convexe) associé à S, c'est-à-dire composé des éléments de la série S : Ext(S) =* $\cup_{I \in S} I$

**Enveloppe convexe d'une série**

*L'enveloppe convexe d'une série S, notée* convexify(S), *est l'enveloppe convexe de l'extension de S (le plus petit intervalle convexe contenant S).*

Il ne faut pas confondre l'extension d'une série avec son enveloppe convexe : la première est en général contenue strictement dans la seconde.

## Relations et opérations sur les séries

*Relations entre les séries :*
Nous distinguons deux relations de type "hiérarchisation" entre les séries. La première, que nous désignerons par le terme *d'inclusion* est l'extension naturelle de la notion d'inclusion des intervalles : une série est incluse dans une autre si et seulement si tout intervalle de la première est inclus dans un intervalle de la seconde.
L'autre relation est plus forte que l'inclusion ; elle exprime le fait qu'une série S1 peut être constituée par tout ou partie des intervalles appartenant à une autre série S2 : dans ce cas, nous dirons que S1 est *extraite* de S2.

**Inclusion de séries**

*Si S1 et S2 sont deux séries, alors S1 est incluse dans S2 (*$S1 \subseteq S2$*) si et seulement si* $\forall I \in S1\ \exists J \in S2\ I \subseteq J$.

**Exemples :**

Si S1 est une série de jours, S2 pourrait être une série de mois, de semaines ou d'années englobant ces jours.

**Série extraite d'une série**

*Si S1 et S2 sont deux séries, alors S1 est extraite de S2 (*$S1 < S2$*) si et seulement si* $\forall I \in S1\ I \in S2$. *Nous dirons encore que S1 est une sous-série de S2.*

**Exemple :**

Si S2 est la série des mois de cette année, la sous-série S1 pourrait être la série des mois de janvier, février et mars.
Remarquons que la relation d'inclusion contient strictement la relation d'extraction dans ce sens que, si S1 est extraite de S2, alors S1 est nécessairement incluse dans S2, la réciproque étant fausse.

*Ratio d'une série par rapport à une sur-série*
Soit S1 et S2 deux séries telles que $S1 \subseteq S2$. Le *ratio* de S1 par rapport à S2 est une fonction qui, à chaque intervalle de S2 associera le nombre d'intervalles de S1 inclus dans cet intervalle. Notons ||E|| le cardinal d'un ensemble E.

**Définition**

*Soient S1 et S2 deux séries telles que* $S1 \subseteq S2$. *On appelle ratio(S1,S2) la fonction définie sur S2 par :* $\forall J \in S2$, $ratio(S1,S2)(J) = \|\{ I \in S1, I \subseteq J \}\|$

Ainsi le ratio de la série des jours par rapport à la série des semaines est une fonction constante : à chaque semaine, elle associe l'entier 7. Par contre le ratio de la série des jours par rapport à une série de mois ou d'années n'est pas une fonction constante.

*Composante d'un intervalle dans une sur-série*
Considérons le cas de deux séries S1 et S2 vérifiant S1 ⊆ S2 (ou *a fortiori* S1 < S2) : pour tout intervalle I de S1, il existe donc un intervalle J de S2 contenant I. Puisque les éléments de S2 sont ordonnés, cet intervalle J est nécessairement unique. Il en découle la définition suivante.

**Définition**

> *Soient S1 et S2 deux séries telles que S1 ⊆ S2. Soit I un intervalle de S1. On appelle composante de I dans S2, et on note* compos(I, S2), *l'unique intervalle J de S2 vérifiant I ⊆ J.*

À partir d'une série quelconque, on peut créer une nouvelle série grâce à trois opérateurs : la restriction, l'extraction et l'agglomération. Auparavant, nous introduisons la notion technique de complémentaire d'une série.

*Complémentaire d'une série*
Le *complémentaire* d'une série est défini par rapport à une référence qui la contient. Cette référence peut être soit une série, soit un simple intervalle. L'idée consiste à considérer la série formée de "tous sauf les éléments de la série initiale". On vérifie aisément que le résultat obtenu a toutes les propriétés d'une série.

**Définition**

> *Soient S1 et S2 deux séries telles que S1 ⊆ S2. On appelle complémentaire de S1 par rapport à S2, et on note* complément(S1, S2) *la série constituée par les "morceaux" de S2 sans intersection avec les éléments de S1.*
>
> *Soit S1 une série et I un intervalle tel que S1 ⊆ I (dans le sens où tout élément de S1 est inclus dans I) . Le complémentaire de S1 par rapport à I se définit comme le complémentaire de S1 par rapport à la série dégénérée constituée de l'unique intervalle I.*
>
> *Formellement : complément(S1, S2) est une série S vérifiant : (i) S ⊆ S2, (ii) ∀ I ∈ S ∀ J ∈ S2 I ∩ J = ø., (iii) S est maximale relativement à l'inclusion (⊆) pour ces des propriétés*

Ainsi, le complément de la série des lundis dans la série des mois de mars est une série dont les éléments sont des conglomérats de jours : la plupart contiendront 6 jours (du mardi au dimanche), sauf ceux situés en début ou en fin de mois qui pourront en avoir moins. Bien entendu, aucun de ces conglomérats ne contient de lundi.
Un cas particulier est celui où on l'on considère le complémentaire d'une série dans l'intervalle constituée par son enveloppe convexe : on obtient alors la série formée par les "gaps" de la série initiale.

**Définition**

> *Soient S une série quelconque.*
> *On appelle complémentaire de S, et on note Gap(S) la série définie par :*
> *Gap(S) = complément(S, Convexify(S))*

*Restriction d'une série*

**Restriction par rapport à un intervalle : restrict(S,J)**

> *Si S est une série et J un intervalle quelconque (convexe ou non), alors la série Restrict(S,J), notée aussi S/J, est la série constituée par les éléments de S inclus dans l'intervalle J :*
> *Restrict(S,J) = S/J = { I∈S, I ⊆ J}*

**Exemple et commentaire :**

Si S est la série des mois de 2005 et J est l'intervalle correspondant au premier trimestre 2005, alors la restriction de S par rapport à J est la série des mois de janvier, février et mars 2005.
Il faut noter que cette définition pourrait être assouplie en prenant en compte dans la restriction tous les éléments de S qui ont une intersection non vide avec J.
Ainsi, dans l'exemple suivant :

*Ce mois-ci, j'irai à la piscine le mardi de chaque semaine*

on va commencer par calculer la restriction de la série des semaines à l'intervalle constitué par le mois courant. Selon les termes de la définition ci-dessus, la dernière semaine du mois risque fort de ne pas être retenue car incomplète… et pourtant elle pourra très bien contenir un mardi que l'on a fort envie d'accepter dans l'ensemble des mardis où l'on ira à la piscine.
Pour permettre cette interprétation plus souple, nous proposons une variante de l'opérateur Restrict, que nous appellerons RestrictSouple, définie par :

**Définition : restriction souple**

*Si S est une série et J un intervalle quelconque (convexe ou non), alors la série RestrictSouple(S,J) est la série constituée par l'intersection avec J des éléments I de S qui ont une intersection non vide avec J :* $RestrictSouple(S,J) = \{ I \cap J \ \ I \in S, \text{ et } I \cap J \neq \emptyset \}$

Il reste que l'emploi de l'une ou de l'autre définition est difficile à trancher du fait d'une ambiguïté inhérente à la langue dans la plupart des cas. Ainsi, lorsqu'on parle des mois d'été, on aura tendance à y inclure ou non le mois de juin selon qu'il ait fait beau ou non en juin, ou selon des critères culturels ou personnels. Nous ne trancherons pas ce débat et nous contenterons de proposer les outils nécessaires à la représentation de l'une ou l'autre des deux acceptions.
En tout état de cause, le problème est sans doute encore plus complexe et il est des cas où même la restriction assouplie ne convient pas : en matière de numérotation des semaines dans l'année par exemple, la norme ISO-8601 prévoit que l'on numérote les semaines dans l'année qui contient le plus grand nombre de ses jours ! Un morceau de semaine de deux jours ne devrait donc pas être retenu dans la restriction semaines/années alors qu'il le sera manifestement dans notre définition.

**Restriction par rapport à une série : restrict(S1,S2)**

*Si S1 et S2 sont deux séries, alors la série Restrict(S1,S2), notée aussi S1/S2, est la série constituée par la restriction de S1 par rapport à l'extension de S2 :*
*Restrict(S1,S2) = S1/S2 = Restrict(S1, ext(S2))*

**Exemples et commentaire :**

Si S1 est la série des jours de 2005 et S2 est la série des mois pairs de 2005, alors la restriction de S1 à S2 est la série des jours contenus dans les mois pairs de 2005.
De même, la restriction *lundi/mars* fournira la série des lundis des mois de mars.
On retrouve évidemment le même type de questions que ci-dessus dès lors que l'on calcule la restriction d'une série par rapport à une autre pour laquelle les éléments ne sont pas alignés (semaines/mois par exemple).

**$Restrict_n(S1,S2)$ :**

*Si S1 et S2 sont deux séries et n un nombre entier, alors la série* $Restrict_n(S1,S2)$*, notée aussi* $S1/_nS2$*, est la série constituée par chaque nième élément de la série S1 restreinte au composant de S2 contenant cet élément :*
$Restrict_n(S1,S2) = S1/_nS2 = \{ I \in S1/S2, ordre(I,S1/compos(I,S2)) = n\}$

**Exemples et commentaire :**

Si S1 est la série des lundis et S2 la série des mois de mars (l'une et l'autre circonscrite par un cadre donné), alors lundi/$_2$mars représente la série des deuxième lundis de mars.
En effet, pour chaque élément I de S1 (pour chaque lundi donc), compos(I,S2) est le mois contenant ce lundi et la série S1/compos(I,S2) est la série des lundis du mois contenant le lundi I. Dire que l'ordre de I dans cette série doit être égal à deux signifie donc bien que I doit être un deuxième lundi du mois.
La définition ci-dessus se généralise de manière évidente à un ensemble E d'entiers.

**Restrict$_E$(S1,S2) :**

*Si S1 et S2 sont deux séries et E un ensemble de nombres entiers, alors Restrict$_E$(S1,S2) = { I∈ S1/$_n$S2 pour n ∈ E }.*

**Restrict$_C$(S) :**

*Si S est une série et C une contrainte quelconque, alors Restrict$_C$(S)est la série des éléments de S qui satisfont la contrainte C*

*Restrict$_C$(S) = {I∈S tq I |= C}*

En pratique, la contrainte est le résultat soit d'une qualification des éléments de la série (à l'aide d'un adjectif par exemple : *les lundis pluvieux*), soit d'une contrainte posée par une proposition subordonnée se rattachant aux éléments de la série : *les lundis où je vais à la piscine*.

*Agglomération*
L'opération d'agglomération consiste à créer une nouvelle série à partir d'une série S existante en fusionnant (agglomérant) des composants consécutifs de S. Bien que non limitée à ce cas, l'opération est particulièrement significative lorsque la série donnée est formée d'intervalles se jouxtant les uns les autres. Une telle série sera dite contiguë :
**Définition : série contiguë**
Une série S est contiguë si pour tout intervalle I la composant, à l'exception du dernier : end(I) = beg(succ(I)).
Observons que les noms calendaires génériques (jour, mois, année…) définissent des séries de ce type.
Dans l'opération d'agglomération, les composants fusionnés sont en relation entre eux dans une certaine relation d'équivalence qui a la propriété particulière de ne créer que des classes d'équivalence convexes, dans ce sens que tous les composants de S qui font partie de la même classe d'équivalence (qui vont donc être fusionnés par l'opération d'agglomération) sont consécutifs. Notons encore que si la série initiale n'est pas contiguë (par exemple les lundis) l'agglomération va fusionner des intervalles de cette série avec des intervalles intermédiaires).

**Définition : relation d'équivalence compatible avec une série**

*Soit S une série et ~ une relation d'équivalence sur les éléments de S.*

*Nous dirons que ~ est compatible avec S si et seulement si pour tout $S_i$ et $S_j$ ∈ S, si $S_i \sim S_j$ alors pour tout indice k compris entre i et j, on a $S_i \sim S_k \sim S_j$.*

On vérifie aisément qu'une telle relation d'équivalence permet de définir la relation de succession au niveau de l'ensemble quotient : pour cela, on montre d'abord que toute classe ne peut contenir qu'un unique élément sans successeur dans cette classe ainsi qu'un unique élément sans prédécesseur dans cette classe. Dans les deux cas, la preuve est symétrique et se fait par l'absurde : soient $S_n$ et $S_p$ appartenant à la même classe d'équivalence (donc $S_n \sim S_p$) et tels que ni $S_{n+1}$, ni $S_{p+1}$ ne soient dans cette classe. Supposons que p soit plus grand que n. Dans ce cas, par définition, puisque n+1 est compris entre n et p, $S_{n+1}$ devrait être équivalent à $S_n$.

Notons alors Max(C) (respectivement Min(C)) l'unique élément sans successeur (respectivement sans prédécesseur) de la classe d'équivalence C. Nous dirons que la classe C1 est le successeur de la classe C2 si et seulement si Min(C1) est le successeur de Max(C2). Clairement, la relation de succession ainsi définie sur les classes de l'ensemble quotient permet d'établir une bijection entre l'ensemble quotient et l'intervalle [1..k] (où k est le nombre de classes de cet ensemble). Si on considère alors les intervalles constitués par l'enveloppe convexe de chaque classe, il est aisé de se rendre compte que cette bijection est une bijection croissante de cet ensemble d'intervalles sur [1..k]. Autrement dit, les enveloppes convexes des classes d'équivalence d'une série S par une relation d'équivalence ~ compatible avec S constituent une série que nous nommerons la série quotient de S par ~.
Observons que dans le cas d'une série contiguë, l'enveloppe convexe de chaque classe ne contient que des éléments de la série sous-jacente S, tandis que dans le cas contraire, les intervalles entre les éléments de S sont intégrés dans les éléments de l'agglomération.

**Définition : série quotient**

*Soit S une série et ~ une relation d'équivalence sur les composants de S compatible avec S. La série quotient de S par ~ est la série notée $S/_{\sim}$ constituée par les enveloppes convexes des classes d'équivalence de ~ dans S.*

Cette opération permet de fabriquer à partir d'une série donnée des séries de grain supérieur : ainsi on peut à partir de la série des jours créer la série des semaines comme étant le résultat du quotient de la série des jours par une relation d'équivalence convenable.
On note qu'on a toujours $S \subseteq S/_{\sim}$ pour n'importe quelle relation ~ compatible avec S.
Une extension naturelle de cette opération consiste à regrouper les composants d'une série par paquets de taille constante. Ainsi, soit n un entier strictement positif, on peut considérer la relation d'équivalence ~ définie sur les composants d'une série S par : $S_i \sim S_j$ si et seulement si les quotients entiers de i-1 et de j-1 par n sont identiques. Les classes d'équivalence d'une telle relation sont constituées par des ensembles de la forme $\{ S_{kn+1}, S_{kn+2}, \ldots S_{(k+1)n} \}$ pour un certain entier $k \geq 0$. On vérifie sans peine que cette relation est bien compatible avec n'importe quelle série S. On peut alors poser :

**Définition : agglomération**

*Soit S une série et n un entier strictement positif, On note Agglo(S,n) la série définie par : Agglo(S, n) = $S/_{\sim}$ où $S_i \sim S_j$ si et seulement si quotient(i-1,n) = quotient( j-1,n).*

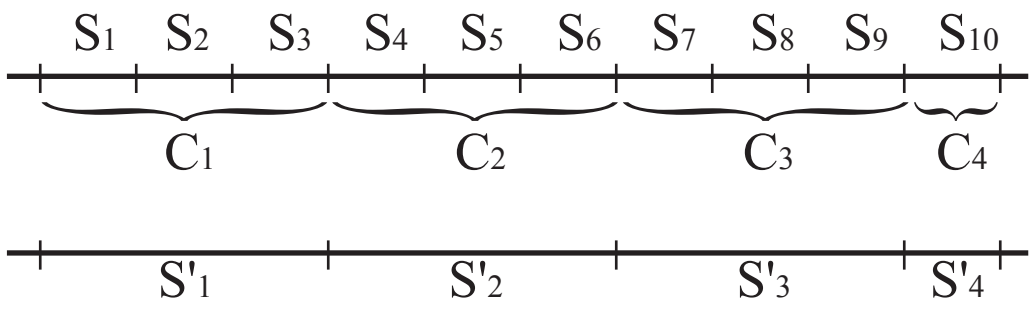

*S' = Agglo(S,3) où S est contiguë*

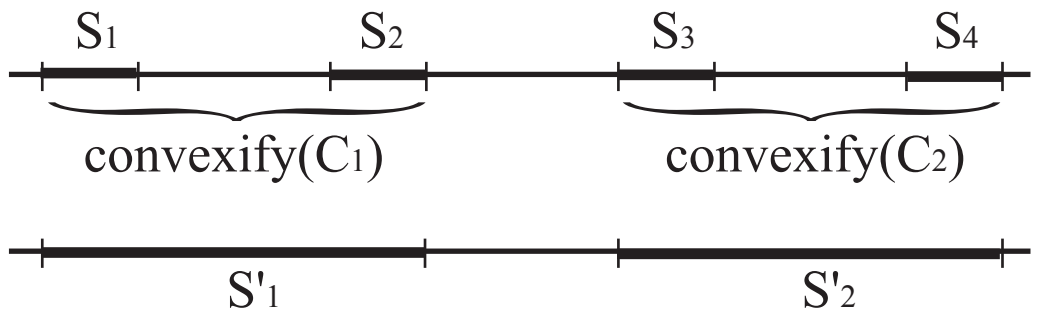

*S' = Agglo(S,2) où S est non contiguë*

*Extraction de sous-série*

L'opération d'extraction consiste à extraire d'une série S une sous-série constituée d'un certain nombre d'intervalles de la série. Nous en distinguons plusieurs variantes : les n premiers, les n derniers, n sur p.

**Définition**

*Soit S une série et soit n et p deux entiers strictement positifs. On désigne par :*

- *Extract(S,n) la série constituée par les n premiers intervalles de S :*
  *Extract(S,n) = $\{S_1, S_2, \ldots, S_n\}$*

- *Extract(S,-n) la série constituée par les n derniers intervalles de S :*
  *Extract(S,n) =* $\{S_{p-n+1}, S_{p-n+2}, \ldots, S_{p-1}, S_p\}$ *avec* $p = ||S||$
- *Extract(S,n,p) la série constituée la répétition du motif "n intervalles retenus, p-n intervalles écartés" autant de fois qu'il le faut pour épuiser les intervalles de S :*
  *Extract(S,n,p) =* $Restrict_E(S, Agglo(S,p))$ *avec* $E=\{1, 2, \ldots, n\}$

## Sémantique des principales constructions

Nous montrons dans cette section comment les principales constructions intervenant dans les CTI peuvent être définies sémantiquement dans le modèle. Chacune de ces constructions sera présentée de manière indépendante et selon un nombre limité de formes syntaxiques. Une description tant soit peu exhaustive de la forme générale des CTI (et de leur sémantique), imbriquant l'ensemble des constructions ici répertoriées, apparaît comme une tâche relativement complexe nécessitant une étude spécifique (cf. section 4 , « Perspectives »). Nous nous attacherons ici à montrer que les opérations définies dans la section 2 fournissent les « ingrédients » d'une description sémantique complète.
Nous considérerons d'abord la base lexicale, les « noms calendaires », puis différents types de constructions nominales qui, du point de vue sémantique, se traduisent par une quantification, ainsi que les adverbes fréquentatifs, sémantiquement similaires. Nous poursuivrons avec la question des heures, des intervalles définis (de… à …) et conclurons avec les subodonnées. L'ensemble des types de CTI mentionnés en introduction sera ainsi couvert.

### *Lexique calendaire*

Comme annoncé en introduction, nous considérons qu'un *nom calendaire*, tel que *jour*, *mois*, *année*, *heure*, mais aussi *lundi, mardi… janvier, février… printemps, été…* dénote dans le modèle une série particulière (notée en petites majuscules : JOUR, MOIS, LUNDI, JANVIER, ETE… Ainsi : [[*lundi*]] = LUNDI, [[*mars*]] = MARS, etc.).

La question qui se pose alors est de déterminer les relations existant entre ces séries, et qui à la vérité, les caractérisent comme constituant un système calendaire. Ces relations sont bien connues et nous ne ferons ici que les évoquer. Pour une application particulière, constituer la base de connaissance adéquate constitue une opération de routine.

- *Hiérarchie « sorte de »*

Un lundi est une sorte de jour, mars une sorte de mois, été une sorte de saison. Formellement la relation correspondante est la relation de suite extraite : LUNDI < JOUR, MARS < MOIS, etc.

- *Inclusions,*

LUNDI $\subseteq$ JOUR $\subseteq$ MOIS $\subseteq$ AN…
De plus les jours constituent une partition des mois ou des semaines, les séries génériques (JOUR, MOIS, AN…) sont contiguës, etc.

- *Mesure*

Le nombre de jours est fixe dans la semaine et « presque » dans le mois etc.
Il conviendrait d'ajouter des relations métriques : tous les jours ont une durée égale, etc. il semble toutefois que la définition adéquate d'une métrique en ce sens intervient peu dans les aspects proprement sémantiques.

*Formes de quantification*

Les noms calendaires n'apparaissent jamais seuls dans les CTI[17] : ils sont toujours soit précédés d'un déterminant (*tous les lundis*, *chaque semaine*, *certains étés*, etc…), soit utilisés dans le cadre d'une quantification explicite (*trois lundis sur quatre*, *deux jours par semaine*, etc…).

Notre souci ici consiste à calculer, de manière compositionnelle, la sémantique du CTI à partir de celle du nom calendaire contenu, autrement dit de définir de quelle façon le déterminant par exemple agit sur l'interprétation du nom calendaire pour fabriquer celle de l'expression complète.

Dans la tradition montagovienne, la sémantique du nom est un sous-ensemble du domaine de l'interprétation et celle du groupe nominal une famille de tels sous-ensembles du domaine. Nous ne nous écarterons pas sensiblement de ce point de vue. Nous venons en effet de définir la sémantique nominale des noms calendaires comme étant une série, laquelle, en tant qu'ensemble d'intervalles, peut être considérée comme un sous-ensemble du domaine de l'interprétation. De même nous allons poser qu'un CTI complet sera interprété comme un ensemble de séries autrement dit une famille de sous-ensembles du domaine. Ce sera le rôle du déterminant — ou de la construction syntaxique particulière – que de passer de la série dénotée par le nom à l'ensemble de série dénoté par le CTI.

Pour illustrer ce point de vue par un exemple, considérons une expression comme "*certains lundis*". Le nom calendaire *lundi* sera interprété par une série bien définie s'intégrant dans le cadre du discours. Le déterminant *certains* va ensuite agir sur cette série pour en extraire une *famille* de sous-séries, car il n'existe manifestement pas qu'une unique façon de définir la sous-série correspondant à *certains lundis* à partir de la série complète des lundis.

Cela étant posé, nous considérons que le déterminant joue aussi un rôle d'introducteur d'une nouvelle entité dans l'univers du discours, cette entité étant dans tous les cas une série choisie parmi les séries candidates, c'est-à-dire celles faisant partie de la famille de séries résultant de l'interprétation du CTI. Plus précisément, si nous considérons que les séries correspondant aux noms calendaires font partie intégrante (par défaut) de cet univers du discours, nous considérons également que chaque CTI va *de facto* introduire dans cet univers[18] un nouvel élément – une série – construite (calculée) à partir des séries existantes à l'aide des opérations algébriques de la section précédente.

C'est la raison pour laquelle nous allons considérer la sémantique d'un CTI comme étant le résultat d'une fonction de choix ε agissant sur l'ensemble de séries constituant l'interprétation *stricto sensu* du CTI pour en extraire un élément particulier. L'interprétation de cette fonction ε sera repoussée jusqu'au moment du calcul effectif d'une interprétation globale du discours. Cette façon de voir nous semble présenter le double avantage d'une part de concilier la théorie montagovienne avec une certaine effectivité du calcul de la représentation et d'autre part d'uniformiser le traitement dans le cas d'expressions imbriquées. En effet, lorsque nous parlons par exemple des *matins des jeudi de certains mois d'été*, il faudrait systématiquement redéfinir au niveau des ensembles de séries les opérateurs définis sur les séries elle-même dans la mesure où l'interprétation de chaque composante de l'expression est en soi une famille de séries. L'utilisation de la fonction de choix ε permet alors d'uniformiser le mode de représentation en considérant que ces opérateurs s'appliquent à l'élément élu par la fonction de choix parmi les éléments de la famille dénotée par la sous-expression.

[17] La seule circonstance où un nom calendaire peut apparaître de manière isolée semble être l'usage déictique de certains noms : *lundi, j'irai à la piscine*. Il va de soi que lundi ne serait pas interprété comme une série dans ce cas pour la simple raison qu'il ne s'agit pas ici d'un circonstanciel itératif. En cela, il sort du champ de notre étude.

[18] Il se peut bien entendu qu'une expression fasse strictement référence à une entité pré-existante dans l'univers du discours : ce serait le cas notamment d'expressions anaphoriques comme "*cette année-là*" ou "*pendant cette période*". Mais ces expressions relèvent plutôt d'une résolution de l'anaphore (calcul de la référence) et ne rentrent pas à proprement parler dans le cadre de notre étude. Par ailleurs, des CTI anaphoriques tels que "*les jours suivants*" (que nous ne traiterons pas davantage ici) , même s'ils demandent également un calcul de la référence, sont aussi introducteurs d'une nouvelle entité, à ceci près que cette entité sera dans une certaine relation avec des entités existantes.

Dans les paragraphes suivants, nous allons considérer quelques expressions ou types d'expression spécifiques (parmi les plus courantes) et définir explicitement leur interprétation. Nous classerons les expressions retenues en trois catégories : celles qui sont introduites par un déterminant (*les*, *un*, *tous*, *chaque*, *certains*, etc.), celles qui comportent explicitement une quantification – un nombre d'occurrences – (*deux lundis sur trois*, *trois fois par jour*) et enfin celles constituées d'un fréquentatif agissant sur une expression de la première catégorie (*souvent le lundi* par exemple).[19] Nous compléterons cet examen par l'examen de la question spécifique des heures, et par un aperçu sur les subordonnées temporelles.

*Détermination*

Les expressions de cette catégorie sont décrites par la grammaire suivante :

CTI → DET NCSPEC
NCSPEC → NC NCSUITE
NCSUITE → vide[20] | DE CTI
DE → *de* | *en* | vide

Il s'agit donc d'un CTI commençant par un déterminant suivi d'un nom calendaire spécifié (NCSPEC), lequel est constitué soit d'un nom calendaire uniquement (NC), soit d'un nom calendaire suivi d'un CTI complément du nom calendaire et relié à lui par une préposition engendrée par le symbole non-terminal DE et qui est en général la préposition *de*, laquelle pourrait parfois être remplacée par *en* (*chaque lundi en été)* ou éventuellement être élidée (*à huit heures le lundi*).

Comme nous l'avons dit plus haut, nous ne prétendons pas faire preuve ici d'une quelconque exhaustivité, la grammaire complète de ces expressions nécessitant des développements spécifiques, ne serait-ce que du fait de la possibilité d'une éventuelle inversion des termes (*le lundi à huit heures*) ou d'appositions avec ou sans inversion (*chaque semaine, le lundi*), etc…

Dans tous les cas, le déterminant va agir sur le nom calendaire spécifié comme un quantificateur généralisé, son interprétation consistant à construire – comme introduit plus haut – l'ensemble des sous-séries acceptables pour ce quantificateur, ensemble spécifié en intension par une contrainte qui devra être satisfaite par chacun de ses éléments. Ainsi par exemple, le déterminant *quelques* appliqué à un nom calendaire spécifié va produire un ensemble de séries dont chaque élément répondra à la double contrainte : d'une part être une sous-série de l'interprétation du nom calendaire spécifié et d'autre part avoir un nombre d'occurrences faible.

D'une manière générale, l'interprétation du déterminant est donc une fonction qui prend en argument une série et renvoie un tel ensemble.

[[DET]] = f : S → { S' vérifiant S' < S et *contrainte sur S'* }

Une fois cet ensemble déterminé, l'interprétation du CTI consistera à lui appliquer la fonction de choix ε pour en extraire un élément arbitraire.

---

[19] Nous retrouvons à peu près (à la différence de point de vue près) les trois « sources de l'itération » énoncées par L. Gosselin dans son propre rapport, respectivement : « les circonstancielles de localisation temporelle », les « circonstants aspectuels itératifs » et les « circonstants aspectuels fréquentatifs »

[20] Nous faisons évidemment référence à la production vide, notée traditionnellement ε. Nous écrivons ici *vide* en toutes lettres pour ne pas risquer de confusion avec la fonction de choix ε précédemment définie (bien que le contexte n'autorise pas vraiment une telle confusion).

Autrement dit, on aura :

[[DET NCSPEC]] = ε ([[DET]] ([[NCSPEC]]))
[[NCSPEC]] = [[NC]] / [[NCSUITE]]
[[vide]] = S0 (le cadre de référence)
[[DE CTI]] = [[CTI]]

le cadre de référence est un intervalle ou une série donnée par le contexte (cas sans doute le plus probable : période dans laquelle le discours est situé, « cadre du discours » du 2.2.1). Mais on peut sans doute trouver des situation où ce cadre de référence est anaphorique, comme dans : « Un année sur deux, je passe mes vacances à la mer. Les lundis je vais pêcher, les soirs je joue au bridge… ». Le cadre de référence de « les lundis », et « le soir » est donné par la première proposition)..

Dans ce paragraphe, nous définissons donc les fonctions correspondant aux déterminants les plus courants. Nous désignerons ces fonctions par un nom prototypique écrit en petites capitales, étant entendu que chacune de ces fonctions recouvre plusieurs déterminants dans la langue. Ainsi, la fonction notée CERTAINS correspondra aux déterminants *certains, quelques, des,* etc…

**Les déterminants "les", "tous les", "chaque", …**

Des exemples de CTI construits avec ces déterminants sont : *les lundis, tous les lundis de mars, chaque année.*

Ces déterminants indiquent que l'intégralité de la série correspondant à l'expression qui suit fait partie de la sémantique de l'expression. Ils seront donc interprétés par la fonction LES définie par :

$\text{LES}\ S = \{\ S' \ \ \text{tq}\ \ S' < S \ \ \text{et}\ \ S' = S\ \}$

**Les déterminants "un", "un certain", …**

Des exemples de CTI construits avec ces déterminants sont : *un lundi, un certain lundi de mars*. On rencontre également la forme moins usitée (désuète ?) "*certain jour*".

Ces déterminants consistent à ne retenir qu'un unique intervalle de la série parente, sans contrainte de position sur cet intervalle. Ils seront donc interprétés par la fonction UN définie par :

$\text{UN}\ S = \{\ S' \ \ \text{tq}\ \ S' < S \ \ \text{et}\ \ \|S'\| = 1\ \}$

**Les déterminants "la plupart des", "presque tous les" , …**

Des exemples de CTI construits avec ces déterminants sont : *la plupart des mois d'été, presque tous les jours*.

Ces déterminants consistent à extraire de la série parente une sous-série constituée d'un nombre d'intervalles légèrement plus faible que celui de la série parente. Si la série parente a une cardinalité finie, ceci peut s'exprimer simplement sous la forme d'un ordre de grandeur du rapport entre la cardinalité de S et celle de sa série parente.

$\text{PLUPART}\ S = \{\ S' \ \ \text{tq}\ \ S' < S \ \ \text{et}\ \ \|S'\|/\|S\| \ > \mathit{seuil}\}$ où *seuil* = 0.66 par exemple

Si la cardinalité de la série parente est infinie[21], la définition ci-dessus peut être reformulée en considérant la sous-série extraite constituée d'un nombre suffisamment grand d'éléments de la série parente :

$\text{PLUPART}\ S = \{\ S' \ \ \text{tq}\ \ S' < S \ \ \text{et}\ \ \exists\ N,\ n>N \Rightarrow \|\text{Extract}(S',n)\|/\|[[\text{Extract}(S,n)]]\| \ > \mathit{seuil}\}$

[21] Cas exclu pour cet article (cf. 2.2.1) mais qui pourrait certainement être considéré.

### Les déterminants "certains", "quelques" , …

Des exemples de CTI construits avec ces déterminants sont : *certains jours*, *certains lundi de mars*. Les exemples avec le déterminant "*quelques*" semblent plus difficiles à trouver, ou recevoir une interprétation en termes de durée (*quelques jours*).
Ces déterminants consistent à extraire de la série parente une sous-série constituée d'un nombre d'intervalles sensiblement plus faible que celui de la série parente. La définition est calquée sur celle de PLUPART en remplaçant le terme "> *seuil* " par "< *seuil*" avec une valeur de seuil faible :

CERTAINS S = { S' tq S' < S et ||S'||/||[[S]]|| < *seuil*} où *seuil* = 0.33 par exemple

CERTAINS S = { S' tq S' < S et ∃ N, n>N => ||Extract(S',n)||/||[[Extract(S,n)]]|| < *seuil*}

*Quantification explicite*

La quantification sur une série n'est pas nécessairement produite par un déterminant. Il existe en effet des expressions quantifiant par elles-mêmes. C'est le cas notamment des CTI de la forme "*n X par Y*" telles que *trois jours par mois*. Comme pour tous les CTI, la sémantique de ces expressions sera une famille de séries (obtenue à partir des interprétations de X et de Y) et à laquelle nous appliquerons la fonction de choix ε.
Pour rester cohérent avec ce qui précède, nous définirons pour chaque expression une fonction FEXP (propre à cette expression) qui rendra comme valeur l'ensemble des séries acceptables pour la fonction de choix. Cependant, alors que dans le cas des déterminants cette fonction s'appliquait à un unique argument (de type Série), nous aurons affaire ici avec des fonctions à plusieurs arguments. Nous distinguerons essentiellement trois signatures pour ces fonctions :

FEXP : (n, S1, S2) → { S' tq *contrainte 1 sur S'* }
FEXP : (n, S) → { S' tq *contrainte 2 sur S'* }
FEXP : (n, p, S) → { S' tq *contrainte 3 sur S'* }

où S, S1 et S2 sont des séries et n et p des entiers naturels

### Expressions de la forme n X par Y

Rentrent dans cette catégories les expressions "*trois jours par semaine*", "*trois lundis par mois*", etc… Plus généralement nous nous intéressons ici à toute expression reconnue par la grammaire :

CTI → n NC1 *par* NC2

où NC1 et NC2 sont des noms calendaires vérifiant [[NC1]] $\subseteq$ [[NC2]].
Du point de vue de la sémantique, nous aurons :

[[n NC1 *par* NC2]] = ε (PAR(n, [[NC1]], [[NC2]]))

où la fonction PAR (n, S1, S2) renvoie un ensemble de séries dont chaque élément est une série extraite de S1 dont le ratio par rapport à S2 est une fonction constante qui à un intervalle J quelconque de S2 associe l'entier *n* :

PAR(n, S1, S2) = { S' avec S' < S1 et ratio(S',S2) = λJ · n }

### Expressions de la forme n fois par Y

C'est une variante de la forme précédente qui fait l'économie de la première série : *trois fois par jour*, par exemple. La règle de la grammaire génératrice est :

CTI → n *fois par* NC

Comme précédemment, la sémantique est un ensemble de séries résultant de l'application de la fonction FOISPAR aux deux arguments n et [[NC]] :

[[n *fois par* NC]] = ε (FOISPAR(n, [[NC]]))

où FOISPAR est définie ainsi :

FOISPAR (n, S) = { S' avec S' $\subseteq$ S et ratio(S',S) = $\lambda$J · n }

### Expressions de la forme n X sur p

Nous considérons ici des expressions telles que *"deux mois sur douze"* ou *"deux jours sur trois"* par exemple. Pour l'interprétation de ces formes, on se ramènera au cas *n X par Y* (étudié plus haut) en considérant que la série Y est le résultat de l'agglomération des composants de X par paquets de p intervalles.
Nous aurons donc :

CTI → n NC *sur* p

[[n NC *sur* p]] = ε (SUR(n, p, [[NC]]))

où SUR est la fonction définie par :

SUR (n, p, S) = { S' avec S' < S et ratio(S', Agglo(S,p)) = $\lambda$J · n }

### Tous les n X

Les CTI de la forme *tous les n X* (par exemple *"tous les cinq jours"*, *"tous les deux ans"*, … ) dénotent un ensemble de séries extraites de [[X]] par l'opération Extract([[X]], 1, n).
On a donc :

CTI → *tous les* n NC

[[*tous les* n NC]] = ε (TOUSLESN(n, [[NC]]))

où TOUSLESN est la fonction définie par :

TOUSLESN (n, S) = { S' avec S' = Extract(S,1,n) }

Rappelons que Extract(S,1,n) = $\text{Restrict}_1$(S, Agglo(S,n)) = $S/_1$Agglo(S,n))

**Forme alternative (plus souple) :**

TOUSLESN (n, S) = { S' avec S' $\subseteq$ Agglo(S,n) et ratio(S',Agglo(S,n)) = $\lambda$J · 1 }

### Problèmes de distributivité

De la manière dont nous l'avons défini plus haut, une expression introduite par le déterminant *un* ("*un lundi de mars*" par exemple) s'interprète naturellement comme le résultat de la fonction UN appliquée à la sémantique du nom calendaire spécifié qui suit. Autrement dit, nous considérons ici (et cela semble raisonnable) qu'il s'agit de la série dégénérée composée d'un unique intervalle (un lundi) choisi parmi tous les lundis disponibles durant les mois de mars.
Par contre, les mêmes règles appliquées à l'expression "*un lundi de chaque mois de mars"* (engendrée pourtant par la même grammaire) vont conduire à une interprétation erronée, puisque identique à la précédente.
Ceci est simplement une preuve de plus, s'il en fallait, que la grammaire des CTI est relativement complexe, avec des règles souffrant des exceptions, et que en l'occurrence une expression telle que "*un lundi de chaque mois de mars"* serait à interpréter comme "*un lundi par mois de mars*". Evidemment, nous avons fourni ici les outils algébriques pour représenter l'une et l'autre acception. Il reste qu'il faudrait – et c'est sans doute autant un travail de linguiste que d'informaticien – construire une grammaire exhaustive et rigoureuse des CTI. Nous faisons simplement ici le pari que les outils fournis permettent dans tous les cas une telle description.

*Fréquentatifs*

Une manière courante d'exprimer la répétition utilise des adverbes et locutions adverbiales de type fréquentatif : *souvent, fréquemment, parfois, rarement, régulièrement, à intervalles réguliers…*

Il y a ici plusieurs cas de figure, selon la place de l'adverbe dans la phrase. Nous considérons qu'un premier correspond à une situation où l'adverbe porte sur un CTI pour le préciser, comme dans :

13. *Fréquemment le lundi…*
14. *Parfois quand je me promène…*

Sémantiquement, l'adverbe va opérer une extraction « quantifiée » sur la série dénotée par le CTI. Ainsi dans (13) *fréquemment* extrait un ensemble largement majoritaire de la série [[le lundi]]. Dans (14) *parfois* extrait un ensemble au contraire réduit de [[quand je me promène]]. Clairement, les opérations liées à ces adverbes seront du même type, et pour beaucoup les mêmes que certaines opérations liées aux déterminants. Cette configuration ne pose donc pas de problème nouveau.

Un second cas de figure est celui dans lequel l'adverbe est directement rattaché au verbe, comme dans :

15. *En été je me promène souvent/rarement/parfois*.
16. *Paul change souvent de voiture*.

Ici, selon L. Gosselin (cf. Contribution de ce dernier au rapport TCAN) les adverbes déclenchent une itération — et, plus précisément, la construction d'une série d'intervalles de procès dont ils précisent la fréquence. Grâce à la section 2, nous avons clairement l'appareillage formel adéquat pour exprimer cette modification. Le problème est ici de nature pragmatique, pour estimer en quelque sorte « l'unité de base » de la répétition, évidemment différente en (12 ) et en (13).

Enfin un autre cas est où l'adverbe présuppose une série implicite, comme dans :

17. *Paul achète souvent une Peugeot.*

Ici, nous avons la série implicite des occurrences d'un achat de voiture par Paul. Et c'estpar rapport à cette série qu'opère *souvent*. On pourrait paraphraser (14) par :

18. *Souvent, quand Paul achète une voiture, c'est une Peugeot.*

Nous sommes donc conceptuellement dans le premier cas envisagé, mais avec une information à rétablir.

Au total, si le premier cas de figure ne pose pas de problème nouveau et peut être intégré simplement dans un modèle des CTI, les deux autres nécessitent une étude linguistique complémentaire.

### *Heures*

Il s'agit de décrire la sémantique d'expressions telles que :

19. *Chaque jour à 8h*

dans laquelle 8h apparaît comme un instant itéré. Nous considérerons donc, de manière cohérente avec ce qui précède, que *8h* désigne une série, la série des instants repérés chaque jour de cette manière. Pour la formalisation, nous disposons de la série des heures, et l'opération de sélection indicée nous permet de repérer la $8^{ème}$ heure de chaque jour : $heure/_8jour$. Mais clairement *8h* n'est pas *la huitième heure* de chaque jour, mais un instant qui marque le début de cet intervalle. Selon les conventions adoptées en 2.1 c'est un « point » sur l'axe du temps.

Nous définissons alors un nouvel opérateur qui retourne le début de chaque intervalle constituant une série.

beg(S) = {beg(J) pour $J \in S$}.

Nous avons alors :

[[8h]] = beg($heure/_8jour$)

L’opération s’étend de manière triviale à d’autres granularités inférieures au jour (minute, seconde…).

### *Les intervalles définis*

Il s’agit dans ce paragraphe de donner une sémantique à des expressions telles que "de lundi à vendredi" ou "de juin à septembre" par exemple. Ces expressions sont du type « de A à B » où A et B sont des noms de série. Nous leur associerons un intervalle défini qui sera une série dont les éléments résultent de la convexification d'un élément de la première série et de l'élément de la seconde qui lui est immédiatement postérieur.

**Définition**

> *Soit S1 et S2 deux séries quelconques. La série Intdef(S1,S2) est définie par :*
> *Intdef(A,B) = {convexify(A’,B’) où* $A' \in A$*,* $B' \in B$*, B’ = fst{I tq* $I \in B$ *et A’<I }*

Comme toutes les séries, les intervalles définis peuvent être restreints à une autre série ou à un intervalle donné. Ainsi une expression comme "l*es jours de semaine de mars*" pourra être interprétée par Intdef(lundi, vendredi)/mars.
Notons qu'en l'occurrence, il conviendrait plutôt d'adopter ici la forme souple de la restriction : [[jours de semaine de mars]] = RestrictSouple(Intdef(lundi, vendredi), mars).
Notons enfin que les bornes d'un intervalle définies peuvent être des séries quelconques, même si dans pratique, elles auront le plus souvent sensiblement la même granularité. On emploiera rarement une expression comme "*de lundi à septembre*" par exemple, encore que la définition ci-dessus donnerait un sens à une telle expression. Un exemple moins arbitraire serait : *"du second lundi à la troisième semaine de mars"* qui se traduirait par : Intdef(lundi/$_2$mars, semaine/$_3$mars).

### *Subordonnées*

Que faire maintenant des CTI subordonnées ? Rappelons ici que dans le modèle SdT, dans une phrase avec subordonnée temporelle telle que *:*

20. *Quand je me promène je rencontre Jean*

Un intervalle circonstanciel est associé à la subordonnée (en tant que telle), qui est relié aux intervalles IR/IP du procès de la subordonnée d’une part (*je me promène*) et aux mêmes de la principale (*je rencontre Jean*). L’idée est alors simplement d’étendre ce mécanisme aux itérations. Nous associerons donc une *série* à la subordonnée (*Quand je me promène*), cette série étant définie par ses relations avec l’IR/IP du procès subordonné (*je me promène*), lui-même défini comme *série*. D’une certaine manière nous n’avons donc pas ici à décrire de calcul spécifique, la question étant renvoyé à une extension de la théorie pour les subordonnées.

## Discussion, perspectives

Nous avons dans ce rapport introduit un modèle formel, de nature algébrique, et les principales opérations permettant de définir une sémantique formelle des CTI. Cette étude constitue la première étape du programme présenté en introduction. Voici pour conclure les suites maintenant planifiées.

### 6) *Définition d’une grammaire sémantique des CTI*

Il s’agit de définir une grammaire constituant une bonne couverture des CTI, dans la diversité de leur formes syntaxiques, en étudiant en particulier l’imbrication des différentes constructions étudiées indépendamment en section 2. De plus cette grammaire devra permettre de calculer la sémantique associée, formulée en termes de série associée à une expression linguistique.
Cette grammaire sera implémentée, évaluée, et mise au point dans la cadre de la plate-forme LinguaStream[22] (Bilhaut, Widlöcher, 2005) Rappelons que diverses grammaires spatiales et temporelles (périodes historiques non itérées) ont été élaborées dans ce cadre (notamment pour le projet GéoSem (Bilhaut et al. 2003) et que la plate-forme permet de manière particulièrement aisée une expérimentation sur corpus. De plus, dans le cadre du présent projet, le logiciel NLPT dévelopé par C. Person (Person, 2004), qui implémente la théorie SdT a été porté sous LinguaStream : ce qui permettra donc une intégration du calcul spécifique sur les CTI.
Une première esquisse (limité à l'analyse syntaxique) a déjà été réalisée, qui pourra être complétée à partir de la présente étude et de travaux linguistiques descriptifs, tels que (Condamines, 1992).

*7) Calcul et raisonnement sur les représentations produites*

L'étape suivant naturellement le calcul sémantique lui-même est constitué par l'élaboration de méthodes de raisonnement et de calcul sur les représentations produites. L'opération type consiste à programmer un test de compatibilité entre deux séries provenant de deux expressions distinctes.
Deux types de problèmes se posent. Le premier est la non-canonicité des représentations calculées : plusieurs expressions linguistiques et plusieurs expressions formelles peuvent dénoter la même série (ou le même ensemble de séries). Ceci rend a priori délicat un calcul direct sur les expressions sémantiques produites. Or nous savons par ailleurs que toute structure temporelle périodique (ou ultimement périodique, c'est-à-dire périodique à partir d'un certain moment) peut être codée dans l'arithmétique de Presburger[23] (Egidi,Terenziani, 2004) Ceci garantit la possibilité « de principe » de transformation de nos représentations en formules arithmétiques canoniques. La comparaison de deux expressions linguistiques se ramène ainsi *in fine* à un test arithmétique décidable. Toutefois, une procédure effective, « adaptée » à nos représentations est évidemment souhaitable et devra donc être mise au point.

Le second problème est lié au caractère « flou » de certaines expressions de quantification : *la plupart de, souvent, parfois*… Des calculs peuvent être menés en principe : si je dis d'une part que *la plupart des lundi d'été je vais voir ma grand-mère* et de l'autre que, certaine année, *je suis parti deux mois à Tombouctou* (ville remarquable mais dans laquelle mon aïeule ne dispose d'aucune villégiature), il y a clairement une contradiction. Toutefois, il s'agit là d'un problème complexe, et les objectifs devraient être clairement précisés. Cet aspect est donc à traiter dans un second temps.

*8) Intégration de cette sémantique dans une sémantique de la phrase et du texte*

Finalement, les CTI n'ont véritablement d'intérêt[24] que dans le cadre d'une analyse sémantique de la phrase et du texte. Voici quelques problèmes à traiter :
- le rôle de *déclencheur d'itération* des CTI, c'est-à-dire leur fonctionnement comme indice conduisant à interpréter comme itératif le procès qu'il situe temporellement ;

[22] http://www.linguastream.org
[23] Rappelons qu'il s'agit d'un fragment de l'arithmétique entière limité à l'addition et aux relations d'égalité et d'ordre. Rappelons également que ce fragment est décidable.
[24] Sauf éventuellement pour certaines tâches de recherche d'information.

- *la relation entre un CTI et son procès.* Deux aspects principaux peuvent être retenus : d'une part la relation entre l'intervalle circonstanciel et les intervalles associés au procès itéré, comme esquissé en section 3 ; et de l'autre, dans la constitution même du procès itéré, la manière dont le *procès modèle* (au sens du texte de Yann Mathet) s'ancre temporellement dans cet intervalle circonstanciel pour créer le procès itéré[25].
- *les ambiguïtés d'interprétation itératif/non itératif* ; par exemple dans *les lundis de mars*, *mars* peut renvoyer à la série des mois de mars (l'expression se lit alors : « les lundis de tous les mois de mars ») ou au mois de mars « de référence » (par exemple celui de l'années courante).
- l'ancrage référentiel des CTI sur une période donnée par la contexte d'énonciation ou, de manière anaphorique, le co-texte précédent l'expression : ce que nous avons appelé le « cadre de référence» en 3.2.1.

Ces différents points doivent faire l'objet d'études linguistiques, la formalisation ne semblant pas poser de problème par elle-même.

## Bibliographie

[25] Voir le texte de Yann Mathet pour une discussion de ce second point

# Circonstants temporels et réseaux de contraintes

Estelle Fiévé, Gérard Ligozat

## Introduction générale

Le travail du groupe d'Orsay s'est principalement située sur deux plans :

1) Marqueurs circonstanciels temporels. Notre travail s'est concentré sur l'utilisation de circonstants temporels, notamment des marqueurs liés à des notion de répétition, avec application à l'analyse de quelques textes du corpus initialement choisi. Nous avons en particulier étudié le conte "Les souliers du bal usés". Il s'agissait de déterminer :

> a) Les effets de sens induits par la présence des marqueurs. Que se passe-t-il si ces marqueurs sont retirés du texte ? Quelles permutations peut-on faire à signification constante ?
>
> b) Principes de représentation des marqueurs de fréquence et de répétition. Parmi les choix possibles, deux se présentent comme aptes à donner lieu à une implémentation informatique. Le premier consiste à considérer des opérateurs engendrant des séries d'événements en quantité indéfinie à partir d'un événement élémentaire. Son avantage est d'amener à une modification minimale de la représentation des événements individuels. L'inconvénient est la difficulté de mettre en relation ces séries avec des événements particuliers dès lors qu'ils servent à constituer un cadre dans lequel va venir s'insérer un déroulement particulier.
> Le second consiste à traiter sur un plan différent les événements ou chaînes d'événements typiques, d'une part, et les occurrences particulières de l'une de ces séries particulières.

Nous verrons que, dans le cadre particulier de représentation que nous proposons, fondé sur l'utilisation de réseaux de contraintes, c'est le deuxième type de solution que nous avons retenu.

2) Représentation de la structure temporelle : construction de chronogrammes à l'aide de réseaux de contraintes.

Tout en reprenant et étendant des idées développées dans des travaux antérieurs de Bestougeff et Ligozat, Galiotou et Ligozat, nous sommes essentiellement partis de l'analyse du temps et de l'aspect dans la langue telle qu'elle a été proposée par Laurent Gosselin (LG).

Le type de contrainte décrite par la théorie, de même que d'autres propriétés analogues des représentations, nous amènent à considérer comme devant faire partie du langage de base de la représentation des relations entre intervalles qui ne sont pas des relations de base dans le style d'Allen, mais plutôt des relations disjonctives particulières. C'est du reste le chemin choisi par Gosselin, qui utilise des relations élémentaires descriptibles en termes de contraintes sur les bornes des intervalles mis en jeu (techniquement, il s'agit de relations dites convexes dans le domaine du raisonnement qualitatif).

Du fait que tout énoncé élémentaire met en jeu plusieurs intervalles, il peut paraître naturel de considérer que cet ensemble d'intervalles constitue l'unité de base pour l'ancrage temporel des énoncés, et de lui accorder donc un statut d'unité élémentaire. C'est le principe utilisé par Bestougeff et Ligozat, Gruselle et Dormont, Galioutou et Ligozat, qui les conduit à fonder la construction de sites temporels (structures analogues aux chronogrammes du projet décrit ici) sur les intervalles généralisés de Bestougeff et Ligozat. La désignation des intervalles ayant pour chaque événement élémentaire l'une des fonctions de référence, etc. est alors faite par une fonction qui, pour chaque représentation d'énoncé élémentaire (appelée TPS par Bestougeff et Ligozat), et pour chacune de ces fonctions, indique les bornes de l'intervalle généralisé correspondantes. Les relations mises en jeu sont alors des relations entre intervalles généralisés. Le travail de Gruselle et Dormont constituait une première tentative de mise en oeuvre de ce type d'approche.

Une deuxième solution consiste au contraire à rester au niveau sous-événementiel, sans chercher nécessairement à remonter au niveau des événements, et à n'utiliser que des relations entre intervalles ordinaires. Les relations mises en oeuvre sont alors des relations d'Allen (disjonctives, comme cela a été vu plus haut).

Les deux choix en question ont été simultanément explorés au cours de notre travail.`

3) Problème de construction incrémentale. De manière simplifiée, le processus de compréhension / interprétation d'un texte consiste, partant d'une structure déjà construite, à intégrer pour chaque nouvel énoncé à interpréter la représentation de ce dernier dans le contexte de la représentation courante.

Un cas particulier simple est étudié par Estelle Fiévé dans sa thèse : d'une part, on se limite à deux énoncés élémentaires. Une fois construite la représentation du premier, on doit déterminer la façon dont la représentation du second (qui est également conditionnée par le cadre formé par celle du premier) va s'intégrer pour former une représentation globale.
D'autre part, les énoncés mis en jeu ne comportent pas de circonstants temporels, que ce soit en tant que circonstants modifiant chaque énoncé qu'en tant que connecteurs.

# I Le choix de la représentation

## I.1 .Le modèle de L. Gosselin

Le système de représentations aspectuo-temporelle proposé par L. Gosselin associe à tout procès quatre types d’intervalles :

9) Intervalle de procès, noté [B1,B2].
10) Intervalle d’énonciation, noté [O1,O2].
11) Intervalle de référence, noté [I,II].
12) Intervalle circonstanciel, noté [ct1,ct2].

Un exemple tiré de Gosselin (1996) illustre le fonctionnement du modèle. A l’énoncé :

*Luc avait terminé son travail depuis deux heures*

on associe une représentation du type suivant :

```
                          { ------2 heures------}
_________|_______________|_______________________|______|__________________|_______|______
         B1              B2                      I      II                 O1      O2
  Luc terminer son travail
```

Dans cette représentation, l’intervalle [B1,B2] représente sur l’axe du temps la période où « Luc termine son travail ». L’intervalle d’énonciation [O1,O2] correspond au laps de temps où est prononcé l’énoncé. Le fait qu’un temps du passé (le plus-que-parfait) soit employé est lié au fait que cet intervalle est postérieur à [B1,B2]. Mais le plus-que-parfait ici correspond à une particularité supplémentaire : le point de vue temporel adopté par le locuteur se réfère à une période [I,II] qui est située elle aussi dans le passé (elle précède [O1,O2]), mais par rapport à laquelle le procès qui avait lieu en [B1,B2] était déjà, à cette époque, du passé : [B1,B2] précède [I,II].

Ces considérations illustrent déjà clairement le type de résultat auquel on s’attend, à savoir, être capable d’associer à chaque énoncé élémentaire (terme à préciser) une configuration particulière des quatre (ou trois) intervalles, c’est-à-dire en particulier d’être capable d’indiquer dans quel ordre apparaissent leurs bornes, et éventuellement de décrire des contraintes sur ces bornes (par exemple, L. G. utilise trois relations de base, où la relation de précédence se décompose en une précédence « immédiate » (les bornes sont dites infiniment proches) et une précédence qui ne l’est pas.

Le cas où figure ce que nous appellerons ici un circonstant temporel est illustré par l’exemple (toujours tiré de LG.)de la représentation de l’énoncé :

*Samedi passé, Luc a été à la pêche*

dont la représentation proposée est la suivante :

```
___|___|__________|_____|_________|_____|____
   ct1   B1         B2     ct2        O1     O2
         I          II
```

*Luc aller à la pêche*

Ici, on a également un temps du passé, mais l'intervalle de référence coïncide avec l'intervalle du procès. La nouveauté est la présence d'un intervalle circonstanciel, correspondant à « samedi dernier », qui englobe l'intervalle de procès : c'est durant la journée de « samedi dernier » que Luc est allé à la pêche.

Nous n'avons fait jusqu'ici allusion qu'aux relations « internes » entre les éléments de ces représentations : comment sont disposés, pour chaque énoncé, les différents intervalles qui lui sont associés ? Mais le but étant de représenter non pas des énoncés isolés, mais des textes complets, il s'agit également de déterminer comment se disposent sur l'axe du temps les représentations des divers énoncés que contient un texte. En particulier, le problème est vu en termes incrémentaux : si l'on dispose d'une représentation d'une partie commençante du texte, puis que l'on ajoute un nouvel énoncé à représenter, quelles sont les lois qui permettront d'intégrer une représentation de ce nouvel énoncé à la représentation courante ? (et cela étant entendu que le contexte que représente cette représentation, courante peut influer sur la structure interne à donner à l'énoncé).

## I.2 Chronogrammes ou réseaux de contraintes ?

De manière simplifiée, l'idée des chronogrammes consiste à réaliser effectivement, pour un texte donné, une figure qui intègre l'ensemble des figures pour chaque énoncé, en indiquant explicitement (de manière graphique donc) les relations mises en jeu entre l'ensemble des différentes bornes. C'est à cette tâche que s'est consacrée une partie des participants au projet.

Outre la difficulté fondamentale que représente  la description des lois d'interprétation, le programme de construction de chronogrammes se heurte à une difficulté bien reconnue du reste par l'ensemble des chercheurs, et décrite par L.G. à propos de la relation entre intervalle de procès et intervalle d'énonciation, mais qui vaut également pour les relations internes entre intervalles associés à un procès et a fortiori celles entre intervalles liés à des procès différents : ces relations ne sont pas en général entièrement déterminées. L'exemple donné par L.G. est celui d'emplois de l'imparfait comme dans  :

*(A ce moment là,) Luc écrivait un roman*

Dans ce cas, rien n'indique que la borne finale du procès (B2) soit antérieure au moment de l'énonciation ; et donc, la figure :

```
___|___|_____|_____|_________|_____|____
     B1  I      II      B2        O1     O2
```

n'illustre que l'une des interprétations possibles de cet énoncé.

La solution choisie par L.G. consiste à conserver la représentation iconique tout en affirmant qu'elle n'est qu'un support intuitif pour une représentation symbolique qui, elle, est rigoureuse : « Plus précisément, on remplace, dès que possible, par souci de lisibilité, les bornes dont la position n'est pas totalement déterminée […] par des bornes fixées arbitrairement à l'une de leurs positions possibles. Si bien que l'on propose, pour une structure donnée, une seule et non pas l 'ensemble de ses réalisations possibles, tel qu'il est calculable à partir de la formule symbolique. »

Notre proposition ici est de conserver l'indétermination en changeant toutefois la nature des « formules symboliques », pour adopter un formalisme qui, à nos yeux, présente également un caractère intuitif, mais qui permet une représentation naturelle de l'indétermination : le formalisme des réseaux de contraintes.

## I. 3 Réseaux de contraintes temporelles

Un réseau de contraintes (temporelles) est par définition un graphe orienté (fini) dont les sommets représentent des entités temporelles (dans les exemples les plus classiques, des instants ou des intervalles), et dont les arcs sont étiquetés par des relations binaires entre ces entités. Pour fixer les idées, décrivons quelques uns des types les mieux connus : si les entités sont des instants, on peut par exemple utiliser des réseaux étiquetés par des relations « qualitatives » entre instants telles que « < » (précède), « = » (simultané), et « > » (suit), ainsi que par des relations dites disjonctives telles que « ≤ » (précède ou égale), « ≥ » (suit ou coïncide avec), « ≠ » (ne coïncide pas avec », et enfin « ? » (relation indéterminée). La puissance (limitée) du formalisme vient de la représentation de connaissances indéterminées que permet l'utilisation d'étiquettes disjonctives.

Un formalisme apparenté est celui des STN (simple temporal networks) où les relations entre instants sont cette fois qualitatives : une étiquette sur l'arc <x,y> est un intervalle de la droite réelle, ce qui représente la contrainte suivante : la distance y-x est un nombre dans l'intervalle en question. On augment également l'expressivité de ce type de réseau en considérant des étiquettes qui sont des réunions d'intervalles.

Un troisième exemple classique est celui des réseaux d'intervalles utilisant les relations d'Allen. Ici, les sommets du graphe représentent des « intervalles », c'est-à-dire des couples de réels <u,v>, avec u < v, et les étiquettes sont des sous-ensembles de l'ensemble des treize relations de base d'Allen {p,m,o,s,d,pi,mi,si,oi,di,eq}. Les sous-ensembles sont interprétés de manière disjonctive, ce qui signifie par exemple que l'étiquette {p,m} sur l'arc <X,Y> signifie que l'intervalle X=<X1,X2> précède (p) ou rencontre (m, pour *meets*) l'intervalle Y=<Y1,Y2>, ou encore, en termes de bornes, que X2 <= Y1.

Nous laissons de côté pour l'instant l'aspect raisonnement des formalismes à base de réseaux de contraintes, en mentionnant pour l'instant trois notions importantes. Le premier est celui d'inverse d'une relation : si un arc <X,Y> est étiquetée par une certaine relation R, la relation entre Y et X est ce que l'on appelle la transposée de R, notée Rt. Par exemple, pour deux instants, si X < Y (X précède Y), alors Y > X (Y suit X) :les deux relations « < » et « > » sont transposées l'une de l'autre. Si X ≠ Y (X et Y ne coïncident pas), alors Y ≠ X : la relation « ≠ » est sa propre transposée.

La deuxième notion est celle de composition : connaissant une contrainte R sur <XY> et une contrainte S sur <YZ>, que peut-on dire sur la contrainte entre X et Z ? Cette contrainte est appelée composition des deux relations R et S, et notée R o S.

## I. 4 Réseaux et raisonnement

Nous avons vu que le principe de construction d'une représentation tel qu'il est décrit par L.G. amène à résoudre des contraintes issues de l'interprétation des marqueurs linguistiques, ainsi que de principes de construction textuelle.
D'autre part, la représentation sous forme de réseaux de contraintes permet, en utilisant les deux opérations de transposition et de ciomposition, de déduire des relations implicites qui ne sont pas explicitement données lors de la construction. C'est ici que cette représentation manifeste son avantage sur la représentation iconique, car le même type de raisonnement, en présence de représentatifons picturales, exigerait l'explicitation de chacun des modèles possibles lorsque plusieurs cas de figures sont possibles.

Dans ce contexte, toutefois, où on ne désire pas une représentation explicite de toutes les situations possibles, il ne reste pas évident que la représentation obtenue soit cohérente, c'est-à-dire qu'elle corresponde à un modèle au moins. Du point de vue théorique, les travaux ont montré que pour la plupart des formalismes cette vérification de la cohérence a un coût algorithmique exponentiel en fonction de la taille des données (représentée ici par le nombre de sommets du réseau).

Il est important à cet égard de pouvoir disposer de méthodes effectives pour pouvoir vérifier cette cohérence, ou plus généralement, por l'imposer durant la construction incrémentale des représentations.

## I. 5 Aspects algorithmiques

Les algorithmes utilisés pour le test de cohérence d'un réseau de contraintes temporelles sont des variantes de la méthode dite de la chemin-consistence : pour chaque triangle (i, j, k) du réseau, on impose que la relation Rij qui étiquète l'arc <ij> soit contenue dans la composée de celles étiquetant les arcs <ik> et <kj>, en prenant l'intersection :

$$Rij := Rij \cap Rik \circ Rkj$$

Si l'une des intersections est vide, il est clair que le réseau de contraintes considéré ner peut pas être cohérent, et on arrête les opérations. Sinon, le processus devient stationnaire après un nombre fini de changements, puisque les relations d'étiquetage sont des ensembles finis de relations de base.

Comme on l'a signalé, on ne peut pas dans ce cas conclure à la cohérence si les relations sont quelconques. Lorsque elles appartiennet à une des classes dites traitables, la cohérence est assurée.

Toutes les classes de relations convexes que nous sommes amenés à considérer dans ce rapport sont des classes traitables.

La mise en oeuvre des algorithmes de chemin-cohérence utilise en général une file d'attente qui contient au départ tous les triangles du réseau, et les maintient en attente aussi longtemps que la relation sur l'un de leurs côtés est amenée à être modifiée.

## II. Représentation par des réseaux d’intervalles

Nous avons vu plus haut que le principe de départ du modèle de L.G. consiste à associer à chaque énoncé simple un ensemble d'intervalles.

Du point de vue qui est le nôtre, cela peut être vu de deux manières différentes : soit en termes de réseaux de contraintes d’intervalles, soit en termes de réseaux de contraintes d’intervalles généralisés. Nous examinons successivement les bases de ces deux choix.

La méthode la plus simple consiste à traiter les intervalles qu’utilise le modèle comme devant être représentés par des sommets d’un réseau de contraintes dont les relations deux à deux sont à déterminer.

Ce choix est parfaitement cohérent avec le point de vue de L. G . pour qui un modèle adéquat du fonctionnement aspectuo-temporel dans la langue doit être ce qu'il appelle un modèle de sémantique *instructionnelle* : les marqueurs linguistiques correspondent à des instructions pour la construction d’éléments de représentation, ou encore, à des contraintes (L.G. parle aussi de conflits et de leur résolution).

Prenons l'exemple de l'énoncé vu plus haut :

*Luc avait terminé son travail*

Nous avons ici trois intervalles à considérer : l'intervalle de procès B, l'intervalle d'énonciation, et l'intervalle de référence. Le réseau représentatif aura donc trois sommets. Les contraintes sont liées au temps grammatical employé (le plus-que-parfait). Plus précisément, on doit déterminer trois relations linguistiques : l'aspect, le temps absolu, et le temps relatif :

1) Aspect (grammatical), c'est-à-dire relation entre [I,II] et [B1,B2] : ce peut être (d'après L.G., table page 28) l'aoristique ou l'accompli.

2) Temps absolu : il s'agit ici de la relation entre [I,II] et [O1,O2].

3) Temps relatif : la relation en question est ici entre deux intervalles de référence. Nous avons affaire ici à un énoncé isolé.

La résolution des contraintes attachées à ces trois relations permet d'arriver au réseau représenté dans la figure 1.

Dans ce cas, on n'utilise pas la possibilité de représentation indéterminée : chacune des étiquettes du réseau est une relation de base bien déterminée.
Par contre, dans l'exemple mentionné plus haut :

*(A ce moment là,) Luc écrivait un roman*

on obtiendra le réseau de la figure 2, dans lequel la relation entre [B1,B2] et [O1,O2] est disjonctive.

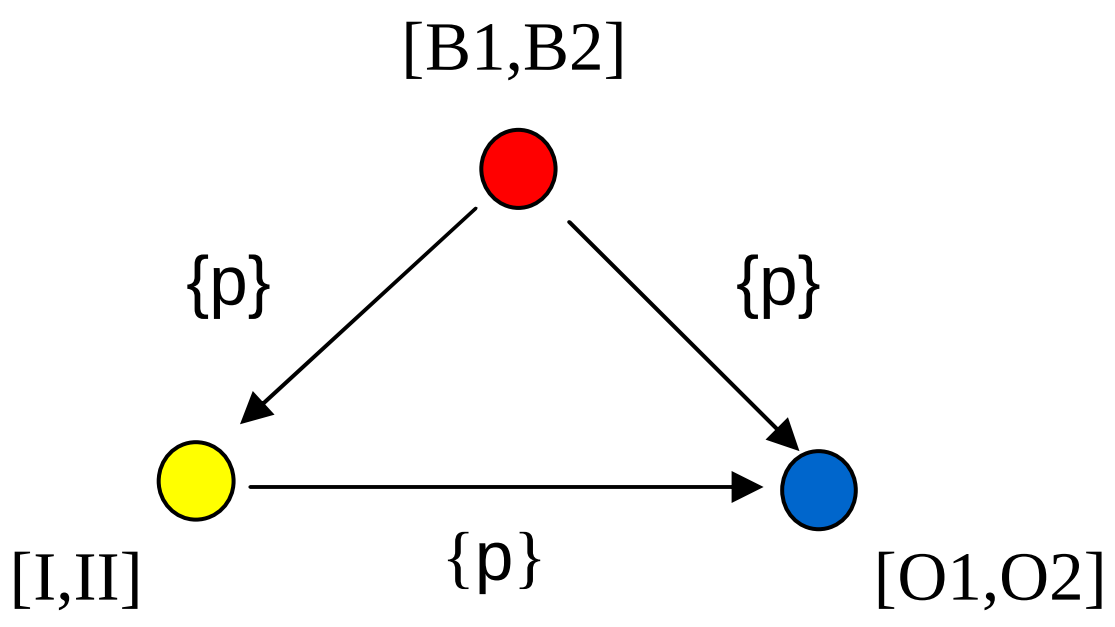

Figure 1 : Représentation de « *Luc avait terminé son travail* »

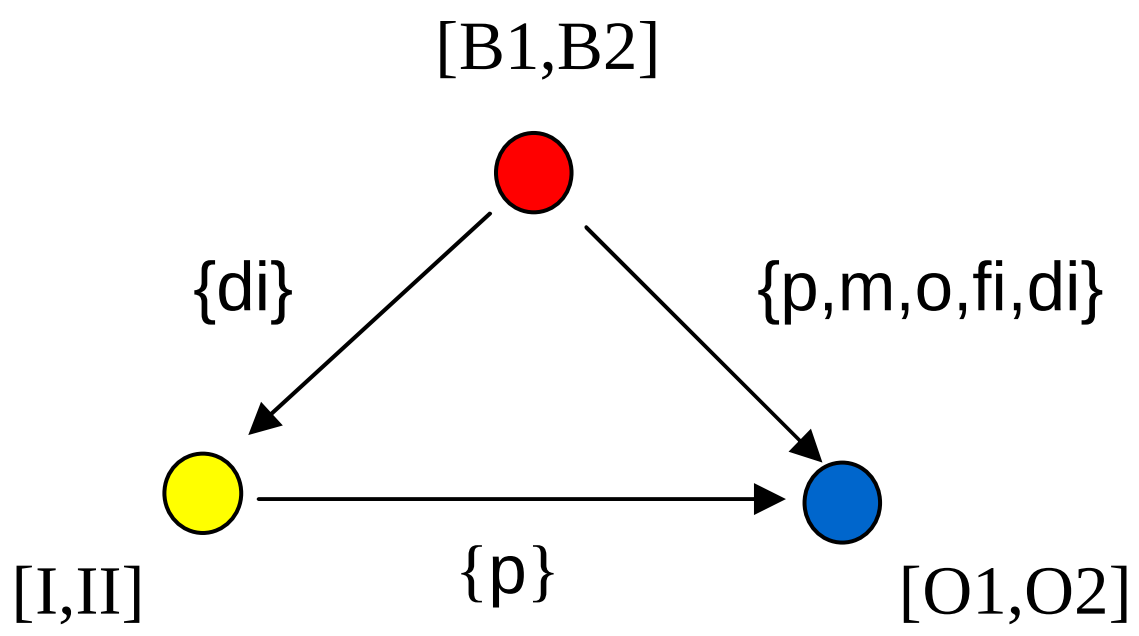

Figure 2 : Représentation de « *Luc écrivait un roman* »

## II. 1 Relations convexes dans l'algèbre d'Allen

Dans le système de L.G., les relations disjonctives sont limitées à un sous-ensemble réduit, à savoir les relations d'antériorité, de postériorité, de simultanéité, de recouvrement, de coïncidence, d'accessibilité, de succession et de précédence.

Nous verrons plus loin que la considération explicite de telles relations (impliquant en particulier que ces relations reçoivent un nom les caractérisant) s'intègre en fait dans un cadre plus général qui peut être utile dans d'autres domaines, tels que par exemple la description qualitative de périodes archéologiques.

Pour décrire les relations disjonctives utilisées par LG. (et également, dans le contexte archéologique, par T. Accary-Barbier et S. Calabretto), il est commode d'utiliser la représentation bien connue des relations d'Allen en termes de treillis, en reprenant le point de vue et les notations de GL.

Rappelons que l'ensemble des relations d'Allen possède une structure d'ordre que l'on obtient en considérant le codage des relations sous la forme de couples d'entiers (entre 0 et 4) comme proposé par G.L.

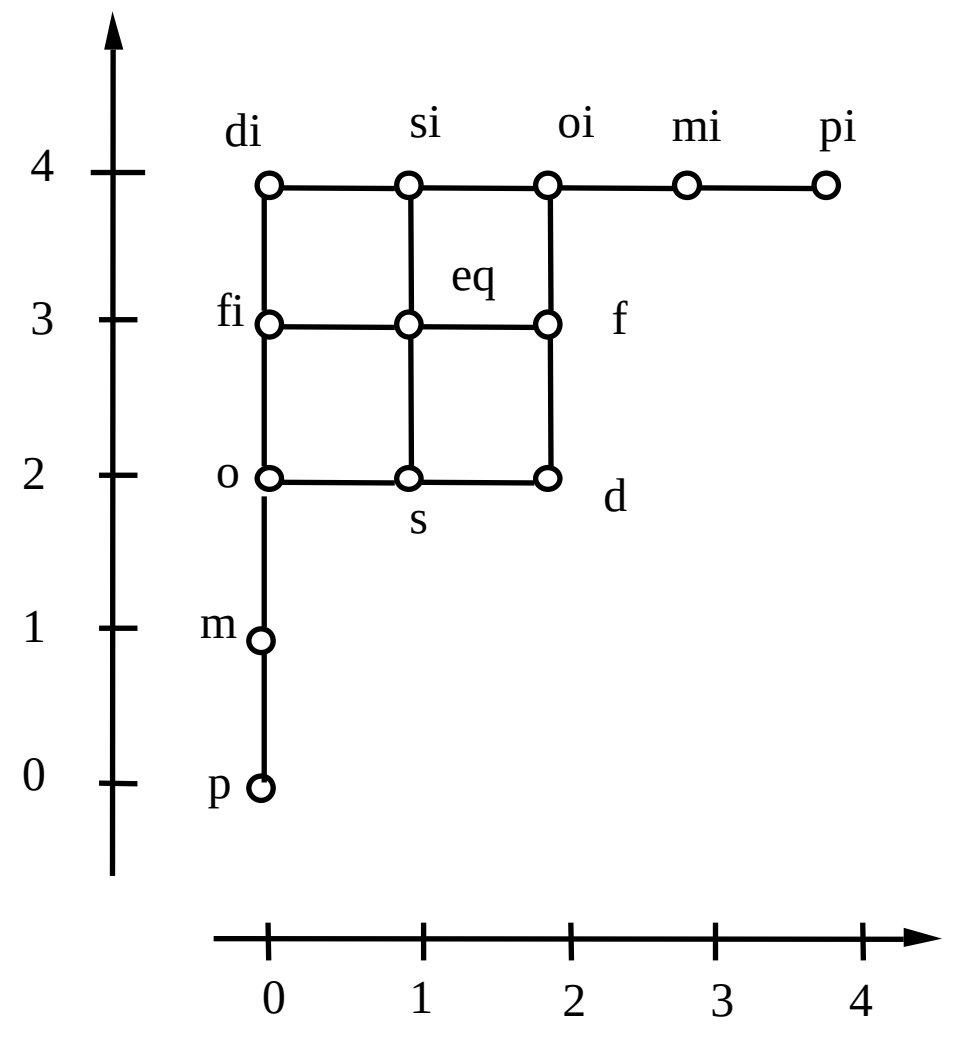

Figure 3 : Treillis des relations d'Allen

Le treillis obtenu est représenté dans la figure 3. Il permet de caractériser aisément les relations disjonctives qui correspondent à des contraintes sur les bornes :

- Les relations dites convexes correspondent aux disjonction de relation de base qui peuvent être exprimées en utilisant de manière conjonctive des contraintes sur les bornes n'utilisant pas la relation « ≠ ». Par exemple, la relation {p,m} entre X et Y est convexe : elle peut être définie par $X2 \leq Y1$.
- Les relations ponctualisables correspondent aux disjonction de relation de base qui peuvent être exprimées en utilisant de manière conjonctive des contraintes sur les bornes, sans restriction cette fois sur le type de contrainte utilisé. Par exemple, la relation {m,o,s, di,si} est ponctualisable, mais non-convexe. Elle peut être exprimée par les conditions suivantes sur les bornes : $X1 \leq Y1$, $X2 \geq Y1$, $X2 \neq Y2$.

G.L. a montré que ces classes de relation avaient une définition très simples en termes du réseau :

2. Les relations convexes correspondent à des intervalles du réseau, c'est à dire des ensembles de relations atomiques définis par une condition du type
{r | a <= l <= b}, où a et b sont deux relations de base telles que a <=b, et où la relation d'ordre est celle du treillis. On note [a,b]l'intervalle en question.

3. Les relations ponctualisables sont celles que l'on obtient lorsque on s'autorise, dans une relation convexe, à ôter des sous-ensembles de relations définis par une coordonnée impaire. Par exemple, la relation {m,o,s, di,si} mentionnée plus haut est obtenue à partir de l'intervalle [m,si] en enlevant l'ensemble des relations de deuxième coordonnée égale à 3, à savoir les relations fi et eq (on pourra se reporter à la figure 3).

L'étude des sous-classes des relations d'Allen a montré l'existence d'une classe plus large de relations, appelées relations pré-convexes ou ORD-Horn, et qui possède également une caractérisation très simple :

- Les relations pré-convexes sont celles que l'on obtient lorsque on s'autorise, dans une relation convexe, à ôter des relations autres que p, o, d, di, oi, pi.

Un exemple typique est la relation {o, s, fi} que l'on obtient en enlevant à la relation convexe [o,eq] la relation eq.

L'importance de cette classe tient à ce que, comme l'ont montré Nebel et Buerckert, elle constitue l'unique classe maximale traitable en temps polynomial. Une démonstration géométrique de ces propriétés a été donnée par G.L.

Les sept relations utilisées par L.G. sont toutes des relations convexes. Trois d'entre elles (antériorité, postériorité, coïncidence) sont des relations de base (à savoir, les relations p, pi, et eq d'Allen). Les quatre autres (simultanéité, accessibilité, succession, précédence) sont des disjonctions qui correspondent à des intervalles du treillis : [m,mi] pour la simultanéité, [fi,si] pour l' accessibilité, [d,pi] pour la succession, et [p,di] pour la précédence. La figure 4 en donne la représentation graphique en termes d'intervalles du treillis.

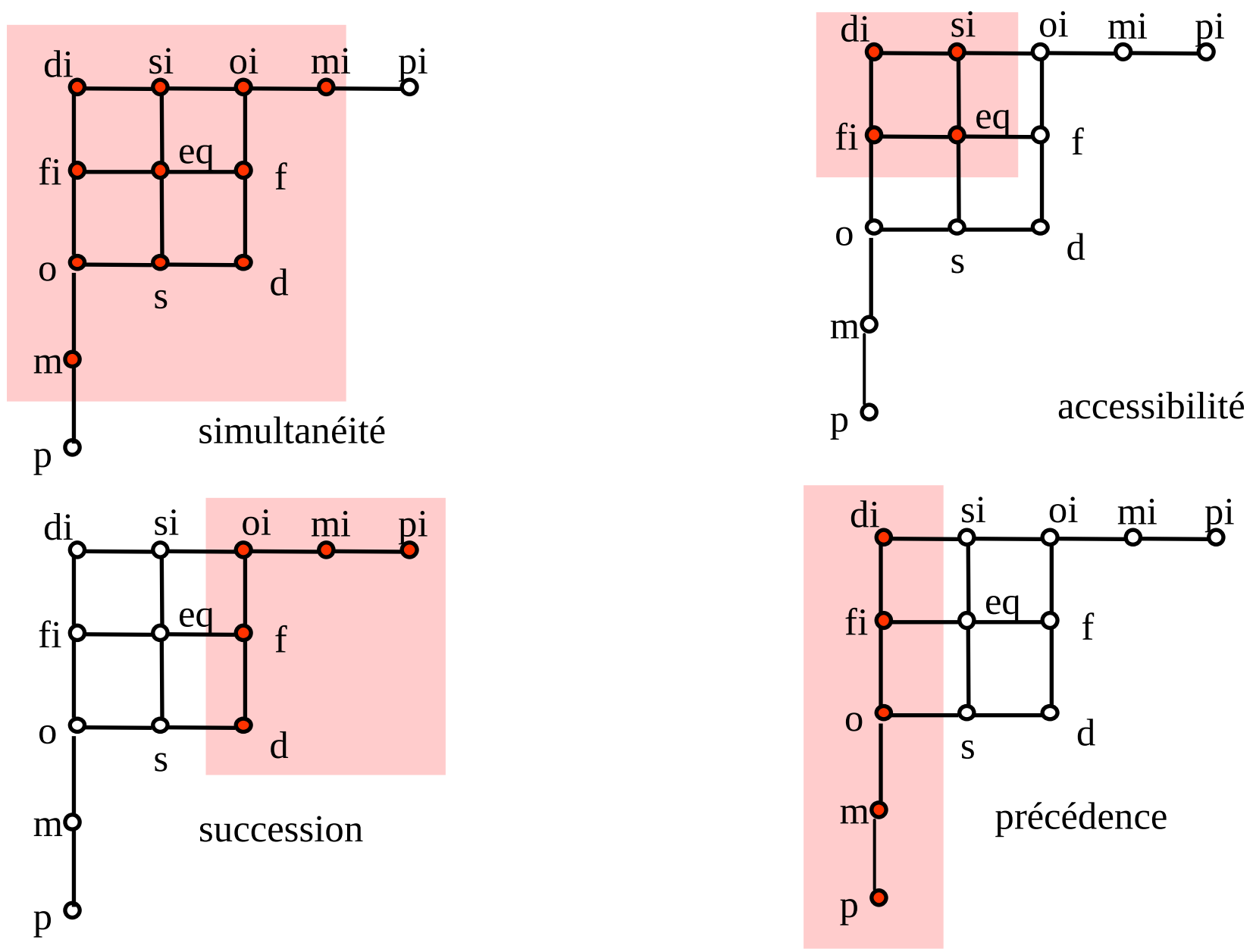

Figure 4 : Les relations simultanéité, accessibilité, succession et précédence

## II. 2 Algorithmique des relations convexes

Nous avons rappelé plus haut le principe des algorithmes de chemin-cohérence. De manière générale, ces derniers amènent à calculer des compositions de type Rik o Rkj. Pour cela, dans le cas général, on doit calculer r o s pour chaque r IN Rik et s IN Rkj, puis prendre leur réunion. On utilise en général une table de composition précompilée qui donne la composition des relations de base. Dans le cas général, donc, si Rik contient M relations de base et Rkj en contient N, on sera amené à consulter la table MN fois pour calculer Rik o Rkj.

Lorsque les relations Rik et Rkj sont des relations convexes, on sait grâce à G.L. que ce calcul peut être fait en effectuant seulement deux accès à la table. En effet, utilisant les notations liées à la structure de treillis, on a :

[a, b] o [c, d] = [inf( a o c), sup(b o d)]

pour tout couple [a, b], [c, d] de relations convexes. En d'autres termes :

1) La composition de deux relations convexes est elle-même convexe

2) La relation convexe obtenue est l'intervalle du treillis dont l'inf est l'inf de la composition des inf, et le sup est le sup de la composition des sup.

## II. 3 Des relations à part entière

Le traitement linguistique du temps amène donc naturellement à considérer comme des "citoyens à par entière" des relations d'Allen disjonctives particulières, qui sont du reste toutes des relations dites convexes (intervalles du treillis). En particulier, il s'avère commode de leur donner des noms et de les utiliser directement au même titre que les 13 relations de base.

Il est intéressant de noter que cette démarche n'est pas restreinte aux applications linguistiques. Nous la relions à deux séries de travaux :

1) Les travaux de C. Freksa (1996) qui considère ce qu'il appelle des "relations grossières" (*coarse* relations) entre "semi-intervalles. Les semi-intervalles sont des intervalles dont les bornes ne sont pas entièrement déterminées. Les relations grossières apparaissent alors comme définies en termes de contraintes sur les bornes : de ce fait, ce sont des relations convexes. Il n'est pas lieu ici de détailler les motivations de Freksa, qui sont en particulier d'ordre cognitif (pertinence cognitive de certaines relations disjonctives) pour l'"introduction des relations grossières. Par contre, il convient de remarquer que les remarques que nous avons faites sur les propriétés algorithmiques s'appliquent à ces relations.

La table 1 donne la liste des 17 relations considérées par Freksa.

2) Dans le cadre de l'élaboration d'un langage permettant aux l'archéologues de formuler de manière commode les relations entre périodes archéologiques, en T. Accary-Barbier et S. Calabretto (2005) proposent d'utiliser neuf relations disjonctives de base. Ces relations sont là encore des relations convexes, et les remarques faites précédemment s'appliquent également à elles. On peut remarquer du reste que des opérations "naturelles" (intersection, transposition, complémentaires) peuvent leur être appliquées, donnant lieu à des relations dérivées, ce qui au bout du compte porterait le nombre total des relations de ce type à 22 relations. Les travaux ultérieurs sur les applications à l'archéologie ou à d'autres domaines permettront d'évaluer la pertinence de ces relations dérivées.

Les relations considérées par T. Accary-Barbier et S. Calabretto sont également données dans la table 1.

| Relation | Gosselin Nom | Symbole | Freksa Nom | Symbole | Accary-Barbier et Calabretto |
|---|---|---|---|---|---|
| [d,pi] | succession | SUCC | younger | yo | |
| [m,mi] | simultanéité | SIMUL | | | common_period |
| [p,di] | précédence | PREC | older | ol | begin_before |
| [fi,si] | accessibilité | ACCESS | | | |
| [p,m] | | | precedes | pr | fuzzy_before |
| [s,f] | | | | | fuzzy_during |
| [s,si] | | | head to head | hh | common_begin |
| [fi,f] | | | tail to tail | tt | common_end |
| [s,mi] | | | | | begin_in |
| [m,f] | | | | | end_in |
| [p,f] | | | | | first_to_end |
| [mi,pi] | | | succeeds | sd | |
| [di,pi] | | | survives | sv | |
| [p,d] | | | is survived by | sb | |
| [o,pi] | | | born before death | bd | |
| [p,oi] | | | died after birth | db | |
| [o,oi] | | | contemporary | ct | |
| [p,o] | | | older and survived by | ob | |
| [oi,pi] | | | younger and survived by | yb | |
| [o,di] | | | older contemporary | oc | |
| [di,oi] | | | surviving contemporary | sc | |
| [o,d] | | | survived by contemporary | bc | |
| [d,oi] | | | youngercontemporary | yc | |

Table 1 : Les relations de Freksa, Gosselin, et Accary-Barbier et Calabretto

## II. 4 Réseaux d' intervalles généralisés

Nous avons dans ce qui précède présenté la solution qui consiste à traiter les réseaux de relation au niveau des entités que constituent les intervalles. Les arcs des réseaux obtenus sont alors étiquetés par des relations d'Allen.

On peut également considérer que l'énoncé constitue en réalité l'unité de bas de la matière linguistique, et qu'il n'y a donc pas lieu de le fragmenter, mais au contraire de le considérer comme un élément (complexe) dont on ne désire pas à priori regarder la structure interne. On pourra, dans un ordre d'idées analogue, appliqué au domaine de l'analyse syntaxique des langues, se reporter aux arguments de Joshi en faveur des grammaires de type TAG (tree adjoining grammars), dans lesquelles les unités de base sont des structures complexes (des arbres) et non des primitives plus simples.

On est amené alors à considérer qu'à un énoncé simple correspond un ensemble de bornes en nombre variable (entre deux et huit) selon qu'on sera en présence ou non d'un, deux, trois, ou quatre intervalles parmi [O1,O2], [B1,B2], [I,II], et [ct1,ct2].

Ce point de vue amène donc à considérer des réseaux de contraintes dont les sommets correspondent à des entités temporelles qui ne sont pas en général des intervalles "ordinaires" (deux bornes), mais des suites finies de bornes (entre deux et 8). Les divers "rôles" tels que celui d'intervalle de parole, de procès, de référence sont représentés par des fonctions qui spécifient à quelle borne correspond chacun des éléments de la suite. Si l'on conserve les notations de L.G., on aura donc des rôles O1, O2, B1, B2, I, II, et ct1, ct2.

### *II. 4. 1 Rappel sur les intervalles généralisés*

Les relations possibles entre deux énoncés peuvent être décrites en utilisant le formalisme des intervalles généralisés introduit par Ligozat et dont l'étude a été développée et étendue par la suite par Ligozat, Osmani, Condotta, et Balbiani.

Rappelons qu'un p-intervalle (pour un certain entier p) est une suite strictement croissante de p éléments d'un ordre total (par exemple des réels, représentant l'axe temporel).

Un q-intervalle y = (y1,...,yq) définit ainsi 2q+1 zones Zi adjacentes dans le temps, qu'on peut numéroter de 0 à 2q : Z0 = {y ! y < y0}, Z1 = {y ! y= y0}, Z3 = {y ! x0 < y < x1}, etc.

La relation qualitative entre un p-intervalle x =(x1, ..., xp) et le q-intervalle y peut alors être encodée en indiquant, pour chacun des points x1, ..., x1, la zone Zj dans laquelle il se trouve. On obtient ainsi une suite non-décroissante d'entiers entre 0 et 2q, qui en fait caractérise entièrement la situation.

On définit donc l'ensemble des (p,q)-relations (entre un p-intervalle et un q)-intervalle de manière abstraite comme l'ensemble des suites de p entiers entre 0 et 2q, qui sont non décroissantes, et dans lesquelles un entier impair n'apparaît qu'une fois au plus.

Ligozat a montré dans une suite de travaux que les intervalles généralisés partagent une grande partie des propriétés du formalisme d'Allen (qui correspond au cas où p=q=2, c'est-à-dire aux intervalles au sens d'Allen.

Le codage obtenu pour les treize relations de base est précisément celui mentionné à propos du treillis : il correspond donc à celui représenté dans la figure 3.

Dans ce contexte, les notions de relation convexe, et plus généralement de relation pré-convexe sont également pertinentes. Il faut cependant distinguer parmi les relations pré-convexes une sous classe, celle des relations fortement pré-convexes, pour laquelle on sait montrer que le problème de la cohérence est polynomial.

### *II. 4. 2 Chaînes typées et réseaux d’intervalles généralisés*

Les travaux de Bestougeff, Ligozat (1989) et Galiotou, Ligozat (2002a, 2002b) décrivent la structure temporelle des textes en termes de « sites temporels » dont les composantes de base sont des « chaînes typées », suites finies de bornes dites « typées », c'est-à-dire pouvant être ouvrantes, fermantes, ou de type indéterminé.

Par exemple, à l'énoncé:

*Paul programmait en Java*

correspondra la structure de la figure 5.

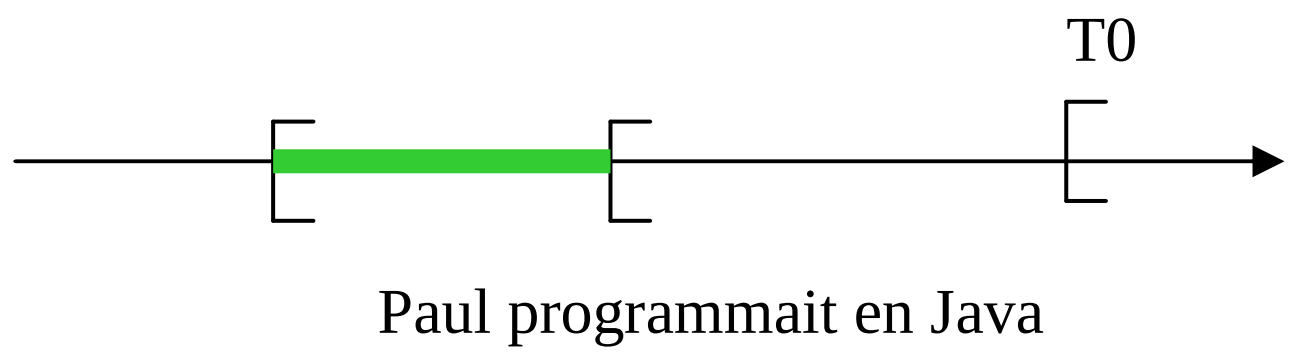

Figure 5 : Une chaîne typée

Les chaînes typées sont utilisées pour construire des structures appelées « sites temporels ». Par exemple, un site temporel contenant deux chaînes typées est associé à l'énoncé :
*Paul traversait la rue quand Alain vint à sa rencontre*

Dans ce site (figure 6), la deuxième chaîne typée est dans la (3,5)-relation dénotée (2,2,5) par rapport à la première : la première et la deuxième borne de l'intervalle du procès correspondant à « venir » est dans la zone 2, et T0 dans la zone 5.

En général, ces relations entre chaînes typées ne sont que partiellement déterminées, ce qui fait que l'on emploiera une représentation sous forme de réseau dont les étiquettes sont des relations disjonctives dans le treillis correspondant.

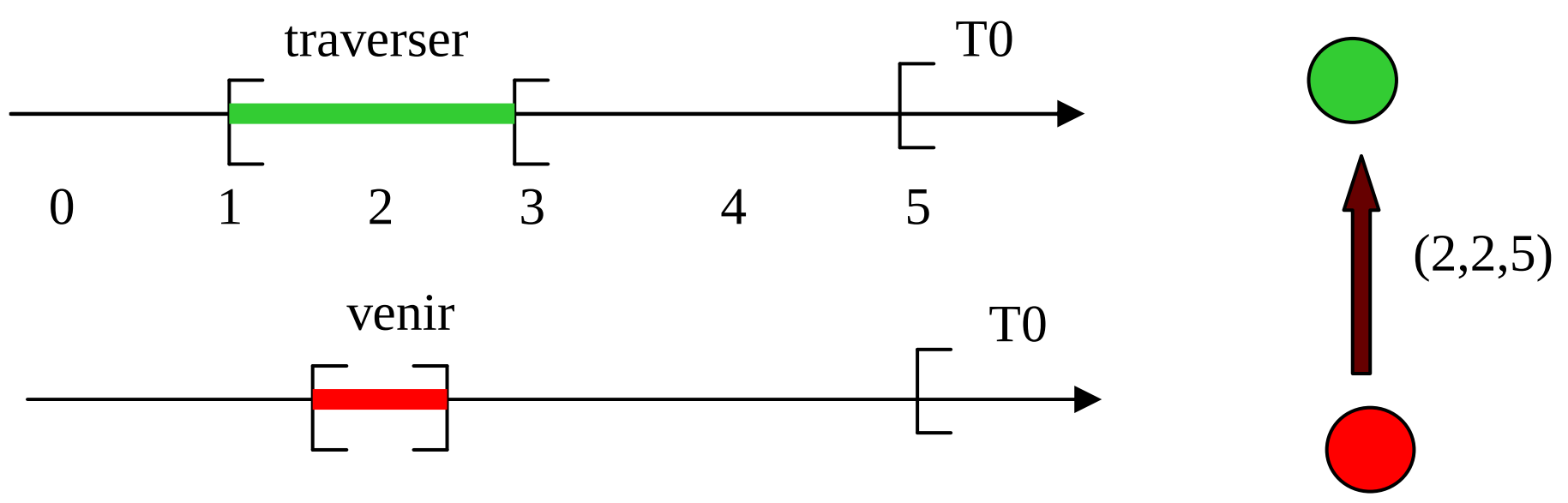

Figure 6 : Un site temporel contenant deux chaînes typées

Le travail de Dormont et Gruselle (1992) (D&G) se base sur ce formalisme pour représenter la structure temporelle de textes narratifs. La figure 7 montre un exemple du type de réseau obtenu, pour le texte suivant :

> *Hier, la délégation du MIDEM est arrivée. Le médiateur australien a accueilli personnellement le neveu du leader indépendantiste. Monsieur O* avait les traits tirés. La* *veille les représentants du RAPP avaient négocié séparément avec les activistes du BIBOP. Les négociations avaient été rudes mais avaient ensuite abouti. La délégation du MIDEM a exigé des explications.*

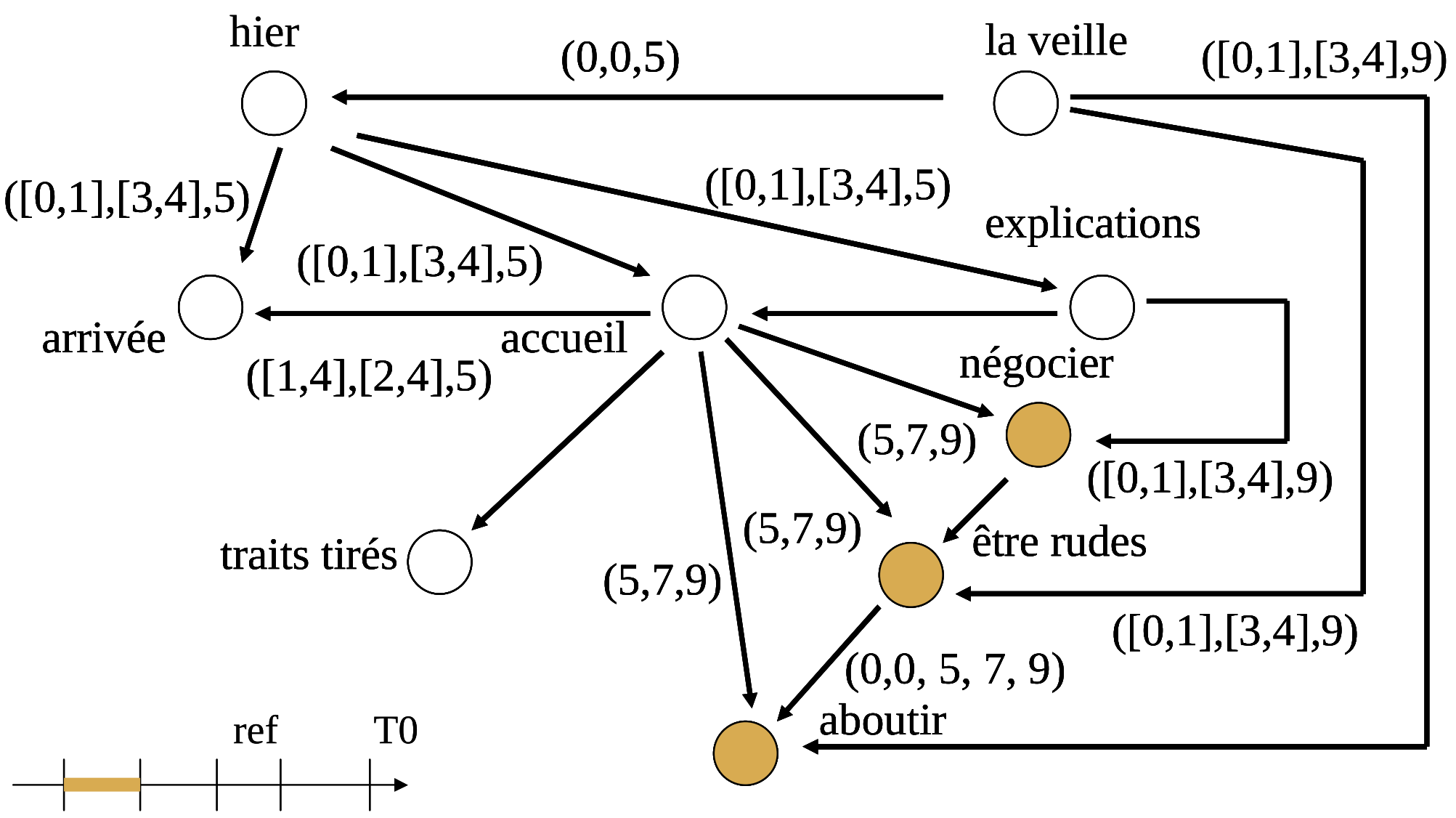

Figure 7 : Un exemple de réseau d'intervalles généralisés

*Commentaire sur les notations utilisées :*

1) Dans le formalisme utilisé ici, on utilise un intervalle de procès, un intervalle de référence, et un instant (plutôt qu'un intervalle) d'énonciation T0. On a donc des 5-intervalles pour représenter des énoncés tels que « les négociations sont rudes ». Par ailleurs, un circonstant temporel tel que « la veille » est représenté par un 3-intervalle. Une relation entre « la veille » et « les négociations sont rudes » est donc une (3,5)-relation, représentée par une suite non décroissante d'entiers entre 0 et 10.

2) Les relations convexes entre intervalles généralisés sont notées par leurs projections. Par exemple, la relation ([0,1], [3,4], 9) est la disjonction des relations (0,3,9), (0,4,9), (1,3,9), (1,4,9) entre le 3-intervalle correspondant à "la veille" et le 5-intervalle correspondant à « les négociations sont rudes ». Ce sont toutes les relations de base comprises (au sens de l'ordre du treillis) entre la relation (0,3,9) et la relation (1,4,9). Avec la notation introduite plus haut pour les intervalles du treillis, elle devrait se noter [(0,3,9), (1,4,9)]. La notation « par les projections » présente l'avantage d'être sensiblement plus compacte.

*Remarques : les intervalles généralisés dans le cadre de LG*

L'application des intervalles généralisés dans le cadre proposé par L.G. pose cependant un problème technique en raison de l'indétermination possible des bornes au niveau de l'énoncé, qui devrait être représentée par des fonctions partiellement définies. Par contre, le formalisme permet là encore de représenter sans difficulté l'indétermination des relations de bornes entre deux (représentations) d'énoncés.

Notons également que D et G choisissent de représenter les circonstants temporels comme des entités à part entière (et donc correspondant à des sommets du réseau), contrairement à ce qui est fait dans le formalisme de LG, où les circonstants sont directement liés à l'énoncé comme intervalles de circonstants. Sur ce point l'approche de D et G conduit à séparer ce qui est vu globalement chez LG.

Un travail approfondi sur la comparaison entre l'approche par intervalles et l'approche par intervalles généralisés reste à faire, et constitue une des perspectives de poursuite du projet.

## II. 5 Structures répétitives et réseaux de contraintes

Le formalisme des réseaux de contraintes (sous l'une des deux versions présentées précédemment) permet également d'envisager un traitement formel des structures répétitives, en partant des deux principes suivants :

1) Stuctures répétitives liées à l'interprétation itérative.
Typiquement, il s'agit de l'interprétation d'un énoncé comme répétition d'un événement de même nature, à la suite par exemple de la résolution d'une contrainte. On est donc en présence d'un sommet du réseau représentatif (l'événement itérartif) qui peut lui-même être analysé comme une séquence 'en nombre indéterminé) d'événements élémentaires de même type. On peut donc ici utiliser l'idée de réseaux hiérarchisés (le sommet en question correspond lui-même à un réseau, et les relations à un niveau de granularité inférieure s'intègrent dans la relation de ce réseau avec les autres éléments de granularité supérieure.

2) Répétitions de séquences d'événements
Il s'agit ici de la répétition de suites d'événements, qui forment une sorte de structure générique dans laquelle viendra se greffer une suite particulière constituant une sorte de variation sur ce modèle général. Il en résulte des possibilités d'inférence basées sur les attentes que suggère la structure générale. En voici un exemple, tiré du corpus :

*Le roi avait douze filles, plus belles les unes que les autres. Elles dormaient ensemble dans une vaste pièce, leurs lits étaient alignés côte à côte, et chaque soir, dès qu'elles étaient couchées, le roi refermait la porte et poussait le verrou. Or, le roi constatait tous les matins, après avoir ouvert la porte, que les princesses avaient des souliers usés par la danse. […].*
***Bientôt, un prince, voulant tenter sa chance, se présenta****. Il fut très bien accueilli, et le soir on l'accompagna dans la chambre contiguë à la chambre à coucher des filles royales. On lui prépara son lit et le prince n'avait plus qu'à surveiller les filles pour découvrir où elles allaient danser ; et pour qu'elles ne puissent rien faire en cachette, la porte de la chambre à coucher resta ouverte.*

Ici, la phrase « ***Bientôt, un prince, voulant tenter sa chance, se présenta*** » introduit une occurrence particulière de la suite (à l'imparfait) mise en place dans le premier paragraphe.

Dans ce cas, nous proposons l'utilisation de « réseau patrons » : un tel réseau patron est associé à la suite générique (dans l'exemple, au premier paragraphe). Lorsque un marqueur d'instanciation (du type « cette semaine là », « ce matin là » ; dans l'exemple, il s'agit de « bientôt »), le réseau à construire peut utiliser le réseau patron comme modèle. L'existence du réseau patron permet de disposer, pour cette occurrence particulière qui ne contient pas explicitement -- au niveau du texte – des termes de la séquence générale (dans l'exemple, il s'agit en particulier des actions du roi : il « referme la porte », « pousse le verrou », « constate « le matin venu, etc.). On pourra ainsi dans la séquence particulière, soit constater que le schéma général s'est reproduit, soit au contraire utiliser la connaissance de l'écart (qui sera cette fois explicitement marqué dans le texte).

## Références

# Annexe : un cadre général pour la classification des expressions temporelles dans la langue

La classification générale proposée ici a été élaborée dans le cadre du projet Ogre par un groupe de travail qu'avait organisé G. Bécher. Elle s’appuie sur un travail préalable d'E. Fiévé et de G. Ligozat, ainsi que des travaux effectués dans le cadre du projet GéoSem.

Il ne s’agit pas ici de se limiter aux intervalles circonstanciels (dans la terminologie de Laurent Gosselin), mais d’élargir l’examen à l’ensemble de tous les marqueurs portant une information temporelle.

## Catégories d’expressions temporelles

Les expressions temporelles dont il s’agit relèvent de ce que l’on appelle également des circonstants temporels. Elles sont regroupées ici non pas du point de vue du résultat, mais de celui de la fonction : on ne cherche pas un lien direct entre l'expression et sa référence telle qu'elle apparaît dans la représentation, mais plutôt quelle information l'expression apporte au processus de construction de cette représentation. Ce point de vue converge bien avec le holisme sous-jacent aux analyses de L. Gosselin.

Ce point de vue nous amène à regrouper les expressions temporelles en quatre catégories : celle des **sites**, celles des **marqueurs de positionnement**, celle des **descripteurs de temporalité interne**, et enfin celle des **sélecteurs** :

### 1) Les sites

Les sites sont des expressions qui ont pour fonction de créer un intervalle (convexe ou non) circonstanciel au sens de Laurent Gosselin. On aura comme exemple de sites introduisant un intervalle convexe « le mois prochain », « depuis 1999 », « dorénavant ») ; comme exemple d’une série d’intervalles temporels amenant un intervalle non convexe « tous les lundis »). Le terme *site*renvoie à la propriété qu'a l'intervalle circonstanciel ainsi créé d'acueillir l'intervalle de procès.
Nous distinguons dans les sites deux grandes sous-catégories : les sites **convexes** et les sites **non convexes**.

#### *a) Les sites convexes*

Dans ce cas, l'intervalle correspondant est un intervalle convexe. Selon les cas, on aura ou non des informations de localisation de cet intervalle. Si tel est le cas, nous parlerons d'intervalles situés ; dans le cas contraire, il s'agira simplement d’une durée (intervalles de temps non repérés et non situés, dans la terminologie proposée par les chercheurs de Toulouse).

##### *a.1 Les intervalles situés*

Les expressions temporelles correspondant à des intervalles situés sont de natures différentes selon que les bornes sont implicites ou explicites.

**Avec deux bornes implicites**

Il s'agit des cas où l'intervalle est nommé (le plus souvent par référence à un calendrier) : par exemple « le lendemain ». Bien qu'il ait deux bornes, ces bornes ne sont pas explicitées et sont nécessairement de même nature.
Cette catégorie se subdivise à son tour en

**Relatif**

Il s'agit de ce que le groupe de Toulouse appelle des repères non stables : pour situer l'intervalle, il faut se référer à un autre intervalle qui peut être soit le moment d'énonciation (déictiques), soit un autre intervalle circonstanciel ou intervalle de référence d'un procès (anaphoriques).
Exemples de relatif déictiques : demain, le mois prochain, l'année dernière
Exemples de relatifs anaphoriques : le lendemain, le mois précédent, l'année suivante.

**Absolu**

Les intervalles absolus sont ceux qui sont situés de manière absolue par référence à un calendrier
Exemples de sites absolus : en mars 2005, au début de 1990, au vingtième siècle, …

**Avec au moins une borne explicite**

Ces expressions parlent explicitement de leur bornes (ou au moins de l'une des deux) qu'elles introduisent ou relient par les prépositions «de», «à», «depuis», «jusqu'à», etc…
Dans ce cas, chacune des deux bornes peut avoir une nature différente : dans l'expression «depuis 1990» par exemple, la borne de gauche est absolue, l'autre déictique.
Exemples : de février à mars 90, à partir de

*a.2 Les durées*

Il s'agit d'expressions définissant un intervalle dont on connaît la durée sans pouvoir le positionner sur l'axe du temps : «pendant 8 jours».

Les prépositions «pendant» ou «durant» introduisent souvent ce genre d'expressions. Mais il est tout à fait possible d'envisager des expressions construite sur «pendant» qui soient en fait des intervalles par rapport à une référence déictique ou anaphorique : pendant ces huit derniers jours, pendant les deux mois précédents, …

### *b) Les sites non convexes*

Les sites non convexes se rattachent nécessairement à un cadre englobant dans lequel ils s'inscrivent. Ce cadre peut être implicite (le cadre du discours) ou explicite («tous les jeudis de 2005» par exemple).

## 2) Les marqueurs de positionnement

Ce sont des expressions temporelles qui ne créent pas d'intervalle circonstanciel, mais qui précisent la position d'un procès par rapport à un autre élément temporel, par exemple « puis », « après » + durée, « en même temps ». L'élément par rapport auquel le procès est positionné  peut être un autre procès, ou un intervalle circonstanciel déjà créé par ailleurs. Nous distinguons deux sous-catégories :

### *a) Marqueurs de séquentialité*

Ce sont tous les marqueurs qui indiquent que deux procès vont s'enchaîner en séquence moyennant éventuellement une interruption non nulle.
On a par exemple : après + durée, dans + durée, après, après que, avant, avant que, plus tard, ensuite, dès que, à peine , aussitôt, bientôt, etc…

### *b) Marqueurs de recouvrement*

Ce sont des marqueur de même type que les précédents, à ceci près qu'ils n'indiquent pas que les procès se suivent, mais au contraire qu'ils se recouvrent, partiellement ou non  : en même temps, simultanément, pendant que, au cours de, lorsque, quand, …

## 3) Les descripteurs de temporalité interne

Il s'agit ici d'expressions qui précisent le déroulement interne d'un procès, c'est-à-dire la manière dont le procès va se dérouler à l'intérieur de son intervalle de processus (terminologie de L. Gosselin). L'information concerne donc certains aspects qualitatifs de déroulement du processus et ce que F. Lévy appelle dans son texte la façon dont le processus occupe son intervalle de procès ; des exemples sont « peu à peu », « régulièrement », « au fur et à mesure », « trois fois de suite »). Ils peuvent s'appuyer aussi bien sur une vision statique du procès que préciser sa dynamique ; le terme de temporalité interne nous paraît convenir à tous les cas.
Les traits modifiés peuvent être :

- la fréquence : souvent, parfois,
- la cardinalité : trois fois,
- la vitesse : lentement, rapidement,
- l'intensité : peu à peu, progressivement,
- la contiguïté : trois fois de suite

Notons que le procès peut être une série sans que cela remette en cause le caractère «interne» de la temporalité. Ainsi, les expressions «peu à peu», «trois fois de suite», etc.. précisent bien la temporalité interne d'un unique processus qui est vu comme une série.

## 4) Les sélecteurs

Les sélecteurs opèrent sur un site (convexe ou non) et introduisent un nouveau site (qui lui-même peut être convexe ou non convexe). Le nouveau site est obtenu par dérivation à partir du site existant, les cas les plus typiques étant celui de la sélection d'une phase particulière d'un processus (« au début de »), et celui de l'extraction de l'un des termes d'une série (« la troisième fois », « ce jour là »). Du point de vue référentiel, ces expressions se comportent donc comme des opérateurs de sélection de site à partir d'un site donné ; du point de vue linguistique, elles ont éventuellement la propriété de focaliser l'interprétation sur une des structures sous-jacentes possibles – par exemple le découpage de l'année en jours plutôt qu'en semaines ou en mois.
Quelques exemples de sélecteurs sont :

- au début de, à la fin de, au milieu de , ... qui agissent sur des sites convexes
- la 3$^{ème}$ fois, cette fois-ci, qui agissent sur des sites non convexes

Le site renvoyé peut être non convexe comme dans le cas de «les trois premières fois» par exemple.

La classification proposée est résumée par le schéma ci-après.

# Schéma de classification des expressions temporelles

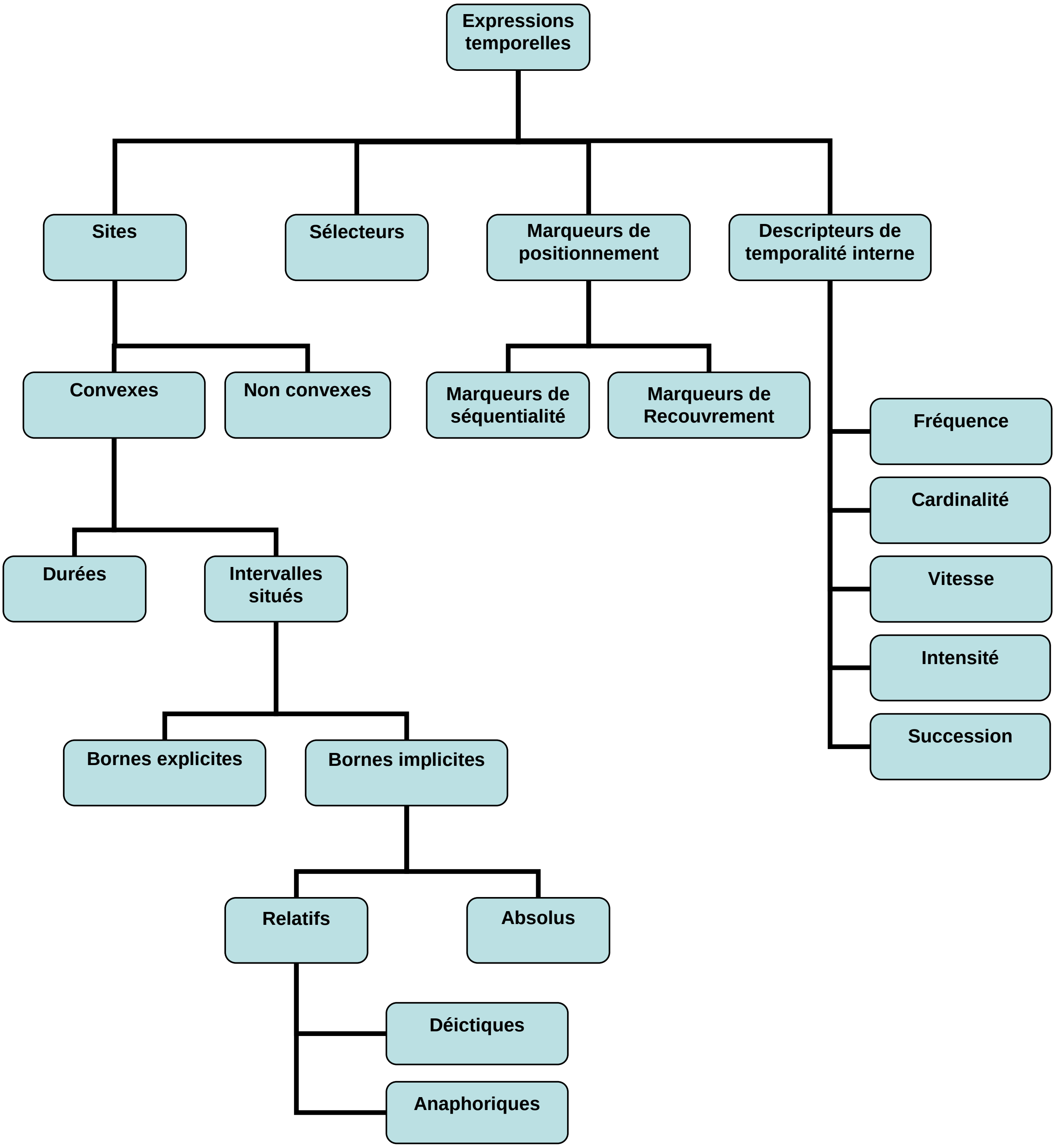